\documentclass{article}
\let\VPTaddcontentsline\addcontentsline
\renewcommand{\encodingdefault}{T1}
\usepackage{iclr2027_conference,times}
\let\addcontentsline\VPTaddcontentsline
\usepackage{amsmath,amsfonts,bm}

\def\eqref#1{equation~\ref{#1}}
\def\1{\bm{1}}

\DeclareMathAlphabet{\mathsfit}{\encodingdefault}{\sfdefault}{m}{sl}
\SetMathAlphabet{\mathsfit}{bold}{\encodingdefault}{\sfdefault}{bx}{n}

\usepackage{titletoc}
\usepackage{hyperref}
\usepackage{url}
\hypersetup{hidelinks,
  pdftitle={Beyond Endpoints: B\'ezier Paths for Vision Foundation Model Adaptation},
  pdfauthor={Chenzhi Zheng, Xi Xiao, Chih-Ting Liao, Qizhen Lan, Jingwen Lin, Tianyang Wang, Hao Xu, Xin Cao}}
\usepackage{booktabs}
\usepackage{colortbl}
\usepackage{tabularx}
\newcolumntype{L}[1]{>{\hsize=#1\hsize\linewidth=\hsize\raggedright\arraybackslash}X}
\newcolumntype{C}[1]{>{\hsize=#1\hsize\linewidth=\hsize\centering\arraybackslash}X}
\usepackage{graphicx}
\usepackage{wrapfig}
\usepackage{pifont}
\definecolor{methodshade}{rgb}{0.898,0.937,0.973}
\definecolor{panelshade}{rgb}{0.957,0.969,0.980}
\definecolor{promptorange}{RGB}{196,90,19}
\definecolor{frozenblue}{RGB}{29,104,160}
\definecolor{estimatecolor}{RGB}{190,45,63}
\definecolor{revblue}{RGB}{0,0,0}
\definecolor{projectpink}{RGB}{196,62,119}
\newcommand{\rev}[1]{#1}
\newcommand{\sd}[1]{\rev{{\tiny$\pm #1$}}}

\iclrfinalcopy
\author{%
\textbf{Chenzhi Zheng}\textsuperscript{\ding{171},}\thanks{Equal contribution.}\quad
\textbf{Xi Xiao}\textsuperscript{\ding{168},\textdagger,}\thanks{Corresponding: \href{mailto:xxiao@uab.edu}{\texttt{xxiao@uab.edu}}.}\quad
\textbf{Chih-Ting Liao}\textsuperscript{\ding{170}}\quad
\textbf{Qizhen Lan}\textsuperscript{\ding{168}}\\[0.4em]
\textbf{Jingwen Lin}\textsuperscript{\ding{171}}\quad
\textbf{Tianyang Wang}\textsuperscript{\ding{168}}\quad
\textbf{Hao Xu}\textsuperscript{\ding{169}}\quad
\textbf{Xin Cao}\textsuperscript{\ding{170}}\\[0.7em]
{\normalfont\small
\begin{tabular}[t]{@{}l@{\qquad}l@{}}
\textsuperscript{\ding{171}}University of Birmingham &
\textsuperscript{\ding{168}}University of Alabama at Birmingham\\
\textsuperscript{\ding{170}}University of New South Wales &
\textsuperscript{\ding{169}}Harvard University
\end{tabular}}%
}
\title{Beyond Endpoints: B\'ezier Paths\\for Vision Foundation Model Adaptation}

\newtheorem{theorem}{Theorem}[section]

\newtheorem{proposition}[theorem]{Proposition}
\newtheorem{corollary}[theorem]{Corollary}

\begin{document}
\begingroup
% Use author-note symbols only for the title; body footnotes keep their defaults.
\renewcommand{\fnsymbol}[1]{%
  \ifcase\value{#1}\or\ensuremath{\dagger}\or\mbox{\scriptsize\ding{41}}\else\arabic{#1}\fi}
\maketitle
\endgroup
% The conference style sets a publication header inside \maketitle.
\lhead{Preprint}
% Official University of Birmingham crest and wordmark on the first page only.
\rhead{%
  \ifnum\value{page}=1\relax
    \raisebox{-3pt}[0pt][0pt]{\includegraphics[height=27pt]{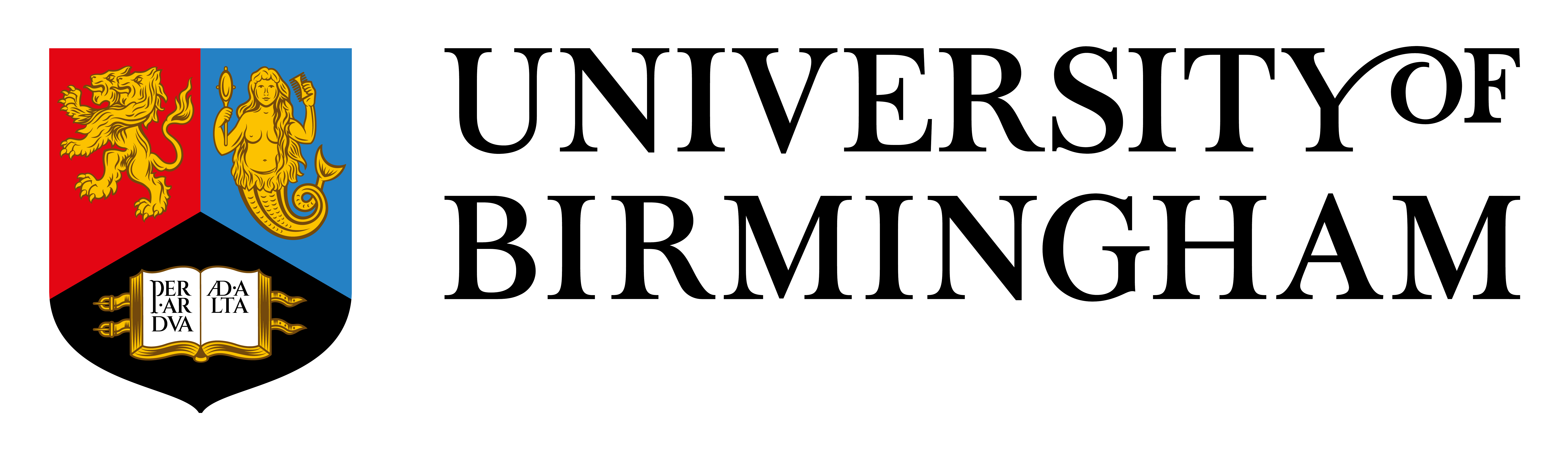}}%
  \fi
}
\begin{abstract}
Frozen vision foundation models can be adapted by learning a few prompt tokens shared across all images. When prompts are updated as data shift or tasks change, the distribution of image features moves from one endpoint to another. Prior theory asks which endpoints prompts can reach; we ask whether they can travel between them along the shortest path. Because shared prompts compete with each image's own content for attention, they often cannot. We quantify this restriction as the extra transport cost that shared attention imposes over the shortest path. In an exactly solvable attention model, the reachable features form a cone; unrolling it into a plane gives the \textbf{exact minimum action} for every pair of endpoints and explicit optimal prompts on short arcs. The analysis reveals a \textbf{sharp threshold in content normalization}, which measures how much attention each image's own tokens absorb: \rev{for same-sign endpoints, at most two distinct levels allow} prompts \rev{to} follow the shortest path, whereas three or more force a strictly positive extra cost unless the endpoints differ only by rescaling. This gap persists with more prompts, small content perturbations, and aligned multihead outputs. On ViT-B/16 with VTAB-1k, a quadratic B\'ezier prompt path that minimizes feature action reduces held-out transport action relative to linear and cross-entropy-trained paths between the same endpoints \rev{and improves intermediate and online accuracy under unlabeled distribution shift}, while cross-entropy-trained paths retain higher minimum accuracy \rev{when labeled calibration data are available}. \emph{Reachable endpoints do not guarantee a reachable shortest path}: how prompts move matters, not only where they end. Project page: \textcolor{projectpink}{\url{https://xixiaouab.github.io/Beyond-Endpoints/}}.

\end{abstract}

\begin{figure}[!ht]
  \centering
  \includegraphics[width=\linewidth]{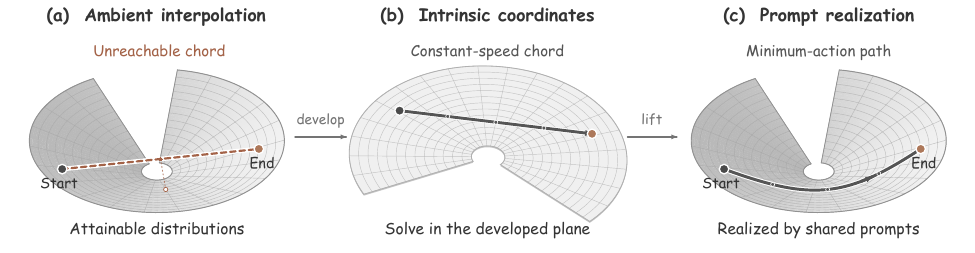}
  \caption{\textbf{Reachable endpoints do not guarantee a reachable shortest path.}
  Schematic; the surface is the family of feature distributions attainable by
  shared prompts. (a) Shared attention excludes the ambient shortcut. (b)
  Unrolling its cone geometry gives a straight path, (c) lifted to optimal
  prompts; its extra cost over (a) is the intrinsic gap. See
  Theorem~\ref{prop:action} and Corollary~\ref{prop:threshold}.}
  \label{fig:teaser}
\end{figure}

\section{Introduction}
\label{sec:introduction}
Vision foundation models are widely used across diverse tasks and domains,
driving demand for adaptation that leaves the pretrained transformer
untouched. One lightweight adaptation mechanism appends a few
learned tokens, called prompts, to the patch tokens of every image
\citep{jia2022visualprompttuning}. These
prompts are often updated as inputs shift or new tasks arrive, while the
model remains in use \citep{gao2022visual,lu2024nullspace}. Since the same
prompts serve all images, each update moves their feature distribution.
A short transport path is a natural reference for gradual adaptation
\citep{wang2022understanding,he2024gradual}. Yet frozen attention makes each
image respond differently to the same prompt update, so their features may
be unable to take the shortest path together. This raises a question:
\emph{what extra transport cost do shared prompts impose between reachable
endpoints?}

Existing work studies either distributional paths or what prompts can
express, but not their intersection. Geodesic interpolation and stochastic
interpolants construct paths without requiring shared-prompt realizability
\citep{he2024gradual,albergo2023stochasticinterpolants,ma2025conditionaldependentcoupling}.
Parameterized Wasserstein flows constrain paths to model families but do
not account for frozen attention \citep{jin2025parameterizedwasserstein}.
Prompt theory studies attainable outputs and capacity
\citep{petrov2023promptingprefixtuning,hu2025fundamentallimits}, while mode
connectivity studies low-loss bridges \citep{garipov2018losssurfaces}.
The missing quantity is the extra transport action required to stay
within the feature family of shared prompts.

To study the gap, we use a frozen head with different content normalizers
across images; each normalizer measures how much attention an image's own
tokens absorb. Prompt count leaves its attainable family unchanged: only
aggregate prompt mass and weighted value matter. With zero content value,
its positive branch is a \textbf{flat cone}. Unrolling the cone solves
every endpoint sign and angle; short positive arcs have explicit optimal
prompts (\mbox{Figure~\ref{fig:teaser}}; Appendix~\ref{app:action}). For
positive endpoints that are not rescalings, two normalizers give no gap,
whereas three or more give a gap at every angle
(Corollary~\ref{prop:threshold}). Intuitively, with two normalizers the
response direction stays in a plane and follows a great circle; a third
bends it out of that plane.

The explicit head shows that adding prompt tokens cannot close this gap,
and a small content-value perturbation preserves a positive gap
(Theorem~\ref{prop:stability}). We prove exact extensions for wider heads
and independently controlled, aligned, unmixed multihead outputs, and give
a projection test for more general
mixing; independent head sums can remove the gap
(Section~\ref{sec:architecture_extensions}). \rev{In continuous time, with a fixed
classifier and positive margin, excess paired action analytically bounds the
fraction of images whose prediction can change along a transition
(Appendix~\ref{app:application_bounds}), although our discretized ViT
evaluations do not establish a non-vacuous certified bound between grid nodes.}
Beyond the intrinsic gap,
action separates into the cost of a
longer route and the cost of uneven speed. Retiming removes the latter,
while the developed chord, the straight path in the unrolled cone,
removes both. We compare feature-action and
loss-based paths between fixed endpoints on VTAB-1k
\citep{zhai2019vtab}, tracking feature motion, intermediate accuracy\rev{, and
unlabeled test-time adaptation under distribution shift}
(Section~\ref{sec:planned_validation}). A B\'ezier path trained for feature
action lowers held-out transport action from \rev{4.50 to 3.20} in the shallow prompt
configuration \rev{(and from 6.30 to 4.20 in the deep prompt configuration)} relative to its
cross-entropy-trained counterpart\rev{, improves minimum shifted accuracy and
online adaptation when target labels are unavailable, and in a hybrid
objective ($\mathcal L_{\mathrm{CE}}+0.15\widehat{\mathcal A}_S$) retains
minimum accuracy within 0.1 percentage points of cross-entropy while
substantially reducing feature motion}.

Our contributions can be summarized in three parts. \ding{182}~We derive
the \textbf{exact minimum action} for a realizable shared-prompt head
across all endpoint signs and angles, and construct optimal prompts on
short arcs. \ding{183}~We prove a \textbf{sharp threshold at three
normalizers} and show when the gap persists with more prompts, small
content perturbations, and vector or multihead outputs. \ding{184}~We
separate \textbf{route and timing costs} and evaluate feature-action
B\'ezier paths between fixed prompts in frozen ViTs. Together, these
results extend prompt-tuning theory beyond endpoint reachability to the
geometry and cost of feasible feature paths.

\section{Related Work}
\label{sec:related_work}
\noindent\textbf{Visual prompt tuning and its theoretical limits.}
Prompt and prefix tuning adapt frozen models through learned context
\citep{li2021prefixtuning,lester2021prompttuning}; VPT brings this idea to
vision transformers \citep{jia2022visualprompttuning}. Later work improves
initialization \citep{wang2024selfprompt,park2025vipamin}, token statistics
\citep{zeng2024fourier,wang2025burstiness}, semantic guidance
\citep{ren2025davpt,xiao2026cascadedsemantics}, and block placement
\citep{yoo2023gatedprompt,shang2025provpt}. Other variants change
cross-attention \citep{huang2025cvpt}, input dependence
\citep{xiao2025viapt,le2026promptexperts}, cross-layer dynamics
\citep{wang2026promptagnostic}, or information allocation
\citep{li2026informationallocation}. Theory studies attention restrictions,
capacity, memorization, and non-interfering updates
\citep{petrov2023promptingprefixtuning,hu2025fundamentallimits,wang2025promptmemorization,lu2024nullspace}.
\rev{Recent adaptation methods address complementary questions: dual-head
optimization separates supervised and VLM-distillation signals
\citep{kang2026dualhead}, TPD stabilizes test-time prompt updates through
magnitude and direction control \citep{wang2026testtime}, and diffusion
meta-prompting generates prompts for cross-task generalization
\citep{sridhar2026diffusion}. Our focus is the constrained geometry and
transport cost of feature-distribution paths between fixed prompt endpoints.}

\noindent\textbf{Stochastic interpolants and constrained transport.}
Stochastic interpolants and flow matching learn fields for prescribed
paths \citep{albergo2023building,albergo2023stochasticinterpolants,lipman2023flowmatching}.
Couplings use minibatch transport, conditional dependence, or model-aligned
pairs \citep{tong2024minibatch,ma2025conditionaldependentcoupling,lin2026modelalignedcoupling};
model-aware schedules adjust timing \citep{jia2026modelawareschedules}.
Riemannian flow matching uses state-space geometry
\citep{chen2024geometries}, \rev{Wasserstein information geometry pulls back the
transport metric to parametric statistical manifolds
\citep{li2018natural,li2023wasserstein},} parameterized Wasserstein flows
use model-family metrics \citep{jin2025parameterizedwasserstein},
\rev{parametric density path optimization (PDPO) optimizes spline trajectories
in parameter space for pushforward density paths \citep{gutierrez2025pdpo},}
and PLOT aligns visual and textual features \citep{chen2023plot}.
\rev{Our analytical contribution is the exact constrained minimum and strict-gap
threshold for the shared-attention cone family:} we derive its cone metric,
minimum-action prompts, and sharp threshold for excess transport cost.

\section{Preliminaries}
\label{sec:setup}

With softmax attention \citep{vaswani2017attention}, a frozen scalar head
with a common query isolates how content normalizers shape each sample's
response to shared prompts.

\subsection{Shared attention and the attainable family}
\label{sec:attention_setup}

Let $N\geq1$ and $m\geq1$ count samples and prompt tokens.
Indices $i,j,\ell,b$ denote samples, prompts, content tokens, and endpoints,
with $b\in\{0,1\}$. Sample $i$ has $L_i$ fixed content tokens
$s_{i\ell}\in\mathbb R^2$, $\ell=0,\ldots,L_i-1$, with readout token $\ell=0$.
The shared prompt tokens $\textcolor{promptorange}{p_j}\in\mathbb R^2$ are unconstrained embeddings. Frozen
linear maps $\textcolor{frozenblue}{H},\textcolor{frozenblue}{K},\textcolor{frozenblue}{V}:\mathbb R^2\to\mathbb R$ produce queries, keys, and values.
\textcolor{promptorange}{Orange} marks prompt embeddings and coordinates;
\textcolor{frozenblue}{blue} marks frozen model maps and parameters.
Assume $\textcolor{frozenblue}{H}(s_{i0})=q$ for a common scalar query $q$, with the attention scaling
absorbed into $q$. Define the content and shared-prompt aggregates:
\begin{equation}
 \begin{aligned}
 Z_i&=\sum_{\ell=0}^{L_i-1}e^{q\textcolor{frozenblue}{K}(s_{i\ell})},&
 A_i&=\sum_{\ell=0}^{L_i-1}e^{q\textcolor{frozenblue}{K}(s_{i\ell})}\textcolor{frozenblue}{V}(s_{i\ell}),\\
 \kappa&=\sum_{j=1}^{m}e^{q\textcolor{frozenblue}{K}(\textcolor{promptorange}{p_j})},&
 w&=\sum_{j=1}^{m}e^{q\textcolor{frozenblue}{K}(\textcolor{promptorange}{p_j})}\textcolor{frozenblue}{V}(\textcolor{promptorange}{p_j}).
 \end{aligned}
 \label{eq:aggregate}
\end{equation}
Both $Z_i$ and $\kappa$ are positive. \rev{Following the prefix-attention
decomposition of \citet{petrov2023promptingprefixtuning}, with}
$A=(A_i)_{i=1}^N$, the scalar output and its empirical law are
\begin{equation}
 z_i^A(\kappa,w)=\frac{A_i+w}{Z_i+\kappa},\qquad
 \mu_{\kappa,w}^A=\frac1N\sum_{i=1}^N\delta_{z_i^A(\kappa,w)},\qquad
 \mathcal M_A=\{\mu_{\kappa,w}^A:\kappa>0,\ w\in\mathbb R\}.
 \label{eq:family}
\end{equation}
Here $\delta_x$ is unit point mass at $x\in\mathbb R$ and
$z^A=(z_i^A)_{i=1}^N$ the output vector. Order samples so
$Z_1\geq\cdots\geq Z_N$, with $d_Z$ distinct values. For $A=0$, omit the
superscript and write $z,\mu_{\kappa,w},\mathcal M$.
Proposition~\ref{prop:realization} realizes every independent aggregate pair
and smooth path with a fixed head. We solve the $A=0$ core, then treat
nonzero $A$ in Section~\ref{sec:stability}. Interpolation time is $t\in[0,1]$.

\subsection{Distribution paths and kinetic action}
\label{sec:transport_setup}

For laws with finite second moments, $W_2$ is the quadratic Wasserstein
distance: $W_2^2(\mu,\nu)=\inf_{\gamma\in\Pi(\mu,\nu)}
\int_{\mathbb R^2}|x-y|^2\,d\gamma(x,y)$, where $\Pi(\mu,\nu)$ contains
their couplings. Let $\mathcal P_2(\mathbb R)$ denote these laws and
$\mathcal P_{2,N}(\mathbb R)$ the equally weighted $N$-atom laws, including
repeated atoms. The map $Q$ lists an empirical law's atoms in increasing
order. For $a,\widetilde a\in\mathbb R^N$, set
$\langle a,\widetilde a\rangle_N=N^{-1}\sum_i a_i\widetilde a_i$ and
$\|a\|_N=\sqrt{\langle a,a\rangle_N}$. Sorted matching gives
$W_2(\mu,\nu)=\|Q\mu-Q\nu\|_N$ for empirical laws
\citep{peyre2019computational}; the proof is in
Appendix~\ref{app:sorting}.

Write $AC^2([0,1];\mathcal P_2(\mathbb R))$ for paths absolutely
continuous in $W_2$ with square-integrable metric speed
$|\dot\mu_t|_{W_2}$. Equivalently, some nonnegative $g\in L^2(0,1)$
satisfies $W_2(\mu_s,\mu_t)\le\int_s^t g(r)\,dr$ for $s\le t$;
the metric speed is the almost-everywhere limit of distance quotients
\citep{ambrosio2008gradient}. For endpoint aggregates
$(\kappa_b,w_b)$, put $\mu_b^A=\mu_{\kappa_b,w_b}^A$ and
$D_A=W_2(\mu_0^A,\mu_1^A)$. We minimize kinetic action over paths
that remain attainable at every time:
\begin{equation}
 \begin{aligned}
 \mathcal E[\mu_\cdot]&=\int_0^1|\dot\mu_t|_{W_2}^2\,dt,\\
 \mathcal E_*(A)&=\inf\{\mathcal E[\mu_\cdot]:
 \mu_\cdot\in AC^2([0,1];\mathcal P_2(\mathbb R)),\quad
 \mu_t\in\mathcal M_A\ (0\le t\le1),\ \mu_b=\mu_b^A\}.
 \end{aligned}
 \label{eq:action_problem}
\end{equation}
The action uses no factor $1/2$ \citep{benamou2000computational}.
Without the constraint, the minimum is $D_A^2$. Thus
$\mathcal G_A=\mathcal E_*(A)-D_A^2$ measures the cost of shared attention;
write $D=D_0$ and $\mathcal G=\mathcal G_0$. We also use path length
$\operatorname{Len}[\mu_\cdot]=\int_0^1|\dot\mu_t|_{W_2}\,dt$.
An empirical stochastic interpolant
\citep{albergo2023stochasticinterpolants,ma2025conditionaldependentcoupling}
arises by drawing $I$ uniformly from $\{1,\ldots,N\}$ and setting
$X_t=z_I(\kappa_t,w_t)$, whose law is $\mu_{\kappa_t,w_t}$.

\subsection{Coordinates for the core result}
\label{sec:geometry_notation}

On the positive branch $w>0$ of $\mathcal M$, set
$u_\kappa=((Z_i+\kappa)^{-1})_{i=1}^N$,
$v_\kappa=u_\kappa/\|u_\kappa\|_N$, and
$R=w\|u_\kappa\|_N$. Then $Q\mu_{\kappa,w}=Rv_\kappa$.
For a fixed $\kappa_{\mathrm{ref}}>0$, define
\begin{equation}
 \theta(\kappa)=\int_{\kappa_{\mathrm{ref}}}^{\kappa}\|v'_s\|_N\,ds,
 \qquad J=\theta((0,\infty)).
 \label{eq:angle_coordinate}
\end{equation}
Primes here denote derivatives in $\kappa$; dots denote time derivatives.
For $d_Z\ge2$, $\theta$ is strictly increasing and has inverse
$\theta^{-1}:J\to(0,\infty)$ (Appendix~\ref{app:cone}). For either
endpoint sign define $R_b=|w_b|\|u_{\kappa_b}\|_N$,
$\theta_b=\theta(\kappa_b)$,
$\Delta=|\theta_1-\theta_0|$, and
$\alpha=\arccos\langle v_{\kappa_0},v_{\kappa_1}\rangle_N$.
$\Delta$ measures travel along the direction curve; $\alpha$ is the
direct angle between its endpoints. The subscript $\mathrm E$ denotes
the Euclidean norm in planar or vector feature spaces.

For perturbation bounds, put
\begin{equation}
 \eta=\left(\frac1N\sum_i\frac{A_i^2}{Z_i^2}\right)^{1/2},
 \qquad h_A=\frac{Q\mu_0^A+Q\mu_1^A}{2},
 \qquad \rho=\operatorname{dist}_N(h_0,Q(\mathcal M)).
 \label{eq:stability_notation}
\end{equation}
Here $\operatorname{dist}_N(a,S)=\inf_{s\in S}\|a-s\|_N$ and
$[x]_+=\max\{x,0\}$. Remaining symbols are defined where they are used.

\section{Method: Optimal Interpolation within Shared Attention}
\label{sec:attention_geometry}

\begin{figure}[!ht]
  \centering
  \includegraphics[width=\linewidth]{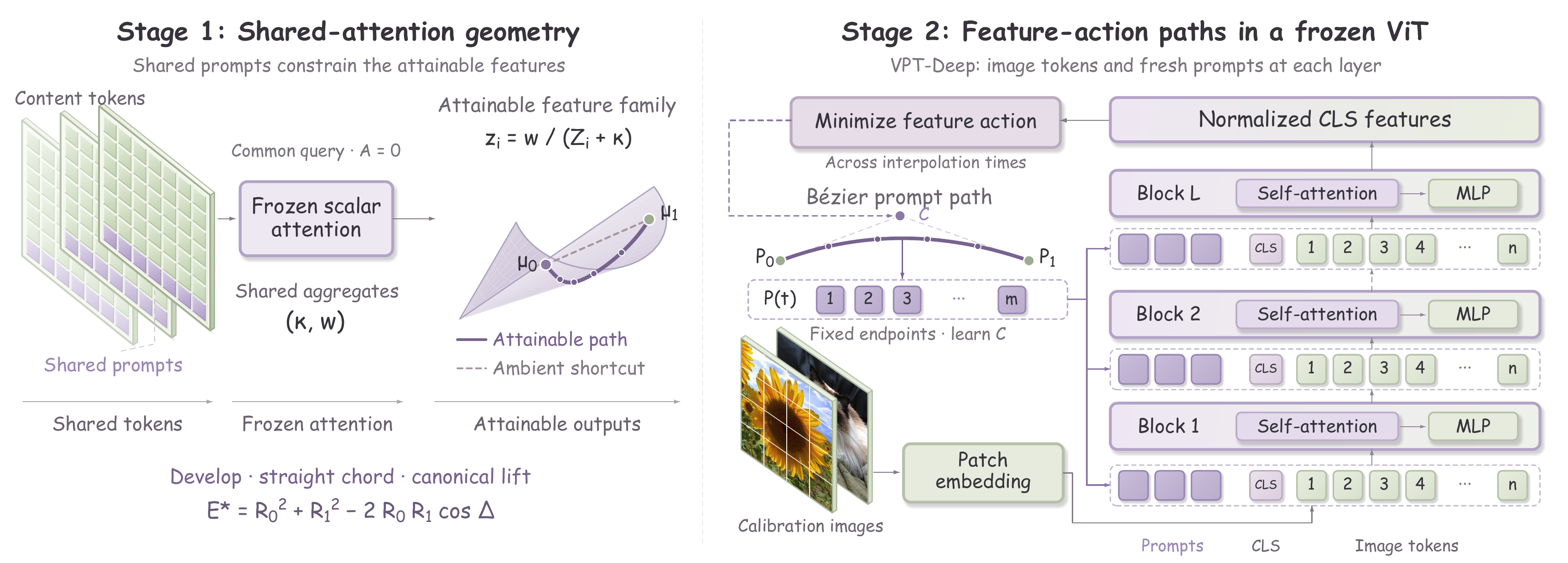}
  \caption{\textbf{Geometry-guided prompt interpolation.}
  \emph{Left:} in the scalar core, we construct a minimum-action path
  within the attainable feature family.
  \emph{Right:} in VPT-Deep, a B\'ezier path supplies prompts to each layer
  of a frozen ViT. We minimize feature action by optimizing its control
  tensor with both endpoints fixed.}
  \label{fig:method_overview}
\end{figure}

Given normalizers $(Z_i)_{i=1}^N$, prompt count $m$, and endpoint aggregates
$(\kappa_b,w_b)$, we develop the $A=0$ geometry into the plane and lift its
optimal path to shared prompts. We then bound the effect of content
perturbations. Figure~\ref{fig:method_overview} links the construction to
frozen-ViT feature-action optimization.

\subsection{An exactly realizable attention family}
\label{sec:realization}

A frozen head independently controls $\kappa,w$ in
(\ref{eq:aggregate})--(\ref{eq:family}), realizing each path with shared tokens.

\begin{samepage}
\begin{proposition}[Exact realization for every prompt length]
\label{prop:realization}
For any positive $(Z_i)_{i=1}^N$ and fixed $A\in\mathbb R^N$, there is one
frozen scalar attention head on two-dimensional tokens, without LayerNorm,
whose attainable family equals $\mathcal M_A$ for every integer $m\geq1$.
Every smooth aggregate path with $\kappa_t>0$ is realized by
\begin{equation}
 \textcolor{promptorange}{p_j}(t)=\textcolor{promptorange}{\bigl(\log(\kappa_t/m),\,w_t/\kappa_t\bigr)},\qquad j=1,\ldots,m.
 \label{eq:token_lift}
\end{equation}
\end{proposition}
\end{samepage}

Choose $\beta\ne0$ with $e^\beta<\min_i Z_i$ and frozen maps
$\textcolor{frozenblue}{K}(p)=p^{(1)}$, $\textcolor{frozenblue}{V}(p)=p^{(2)}$, and
$\textcolor{frozenblue}{H}(p)=p^{(1)}/\beta$ for $p\in\mathbb R^2$.
Fixed content tokens supply $Z_i,A_i$ and common query one.
Prompt count changes token representations, not attainable outputs.
Please see Appendix~\ref{app:realization} for more details.

\subsection{The attainable family has a cone metric}
\label{sec:cone}

Set $A=0$. The scalar $w$ scales outputs, while $\kappa$ changes relative
sample responses. \rev{Pulling back the quadratic Wasserstein metric to this
two-parameter statistical family \citep{li2018natural,li2023wasserstein}
resolves radial and directional motions into} a planar
polar-coordinate metric.

\begin{proposition}[Exact cone coordinates]
\label{prop:cone}
If $d_Z\geq2$, the positive branch of $\mathcal M$ is a smooth embedded
surface in its sorted-vector representation. Its coordinates
$(R,\theta)\in(0,\infty)\times J$ induce the Wasserstein line element
\begin{equation}
 ds^2=dR^2+R^2d\theta^2.
 \label{eq:cone_metric}
\end{equation}
Every admissible path remaining in this branch has
$\mathcal E[\mu_\cdot]=\int_0^1(\dot R_t^2+R_t^2\dot\theta_t^2)\,dt$.
\end{proposition}

Because $v_\kappa$ has unit norm, its change is perpendicular to itself.
Thus radial and angular motion contribute the two terms in
(\ref{eq:cone_metric}). Mapping $(R,\theta)$ to
$R(\cos\theta,\sin\theta)$ preserves their action.
Please see Appendix~\ref{app:cone} for more details.

\subsection{Exact minimum action and the normalization threshold}
\label{sec:action}

Reachable endpoints need not admit a reachable straight path. Comparing
travel along the cone with the ambient chord in sorted output space gives
the exact constraint cost.

\begin{theorem}[Global minimum over all attainable paths]
\label{prop:action}
Assume $d_Z\geq2$, $w_0,w_1>0$, $\kappa_0\ne\kappa_1$, and $\Delta<\pi$.
The infimum in (\ref{eq:action_problem}) for $A=0$ is attained, and
\begin{equation}
 \begin{aligned}
 \mathcal E_*(0)&=R_0^2+R_1^2-2R_0R_1\cos\Delta,\\
 D^2&=R_0^2+R_1^2-2R_0R_1\cos\alpha,\\
 \mathcal G&=2R_0R_1(\cos\alpha-\cos\Delta).
 \end{aligned}
 \label{eq:exact_action}
\end{equation}
The infimum includes sign changes, zero crossings, and changes in atom matching.
\end{theorem}

The short-arc chord gives a smooth prompt optimum. For $d_Z\geq2$, the
global action is
\begin{equation}
 \mathcal E_*(0)=
 \begin{cases}
 R_0^2+R_1^2-2R_0R_1\cos(\min\{\Delta,\pi\}),&w_0w_1>0,\\
 (R_0+R_1)^2,&w_0w_1\leq0.
 \end{cases}
 \label{eq:global_action}
\end{equation}
For $\Delta\geq\pi$ or opposite signs, a path through $\delta_0$ attains
the infimum among distribution paths; smooth prompt paths approach it.
Please see Appendix~\ref{app:action} for the proof and endpoint distances.

The number of distinct normalizers determines whether the direction curve
exceeds the shortest spherical arc between its endpoints.

\begin{corollary}[Two versus three distinct normalizers]
\label{prop:threshold}
For $w_0,w_1>0$, if $d_Z\leq2$, then $\mathcal G=0$ for every pair
$\kappa_0,\kappa_1>0$. If $d_Z\geq3$ and $\kappa_0\ne\kappa_1$,
then $\mathcal G>0$ at every angular span. For the head in
Proposition~\ref{prop:realization}, these statements hold for every $m\geq1$.
\end{corollary}

With two normalizers, the direction curve follows its shortest spherical
arc, so $\Delta=\alpha$. A third distinct normalizer bends that curve
away from the arc, giving $\Delta>\alpha$; the global formula covers
long arcs through the apex as well.
Please see Appendix~\ref{app:threshold} for more details.

\subsection{Constructing the optimal prompt interpolant}
\label{sec:construction}

Write $e(\vartheta)=(\cos\vartheta,\sin\vartheta)$ and
$c(t)=(1-t)R_0e(\theta_0)+tR_1e(\theta_1)$.
Under Theorem~\ref{prop:action}, this developed chord never vanishes.
Let $\vartheta(t)$ be its continuous angle between the endpoint angles.
Recover the prompt aggregates by
\begin{equation}
 R_t=\|c(t)\|_{\mathrm E},\qquad
 \kappa_t=\theta^{-1}(\vartheta(t)),\qquad
 w_t=\frac{R_t}{\|u_{\kappa_t}\|_N}.
 \label{eq:optimal_interpolant}
\end{equation}
Equation~(\ref{eq:token_lift}) then gives every prompt. Computing the path
requires one-dimensional quadrature for $\theta$ and monotone inversion
between $\kappa_0$ and $\kappa_1$.

The lifted chord also gives the empirical interpolant $X_t$ from the
preliminaries. It has a bounded velocity field that is uniformly
Lipschitz in space and satisfies the weak continuity equation.
When $\kappa_0=\kappa_1$ or $d_Z=1$, simple linear paths suffice.
The formal statement, construction, and degenerate cases are in
Proposition~\ref{prop:interpolant} and Appendix~\ref{app:interpolant}.

\subsection{Stability beyond zero content numerators}
\label{sec:stability}

At fixed endpoint aggregates, small content perturbations preserve the
core's midpoint separation from the attainable family. This separation
bounds every perturbed path's action.

\begin{theorem}[A neighborhood with positive excess action]
\label{prop:stability}
Suppose $d_Z\geq3$, $w_0,w_1>0$, and $\kappa_0\ne\kappa_1$. The midpoint
distance $\rho$ in (\ref{eq:stability_notation}) is strictly positive.
For every $A\in\mathbb R^N$,
\begin{equation}
 \mathcal E_*(A)\geq D_A^2+4[\rho-2\eta]_+^2.
 \label{eq:stability_bound}
\end{equation}
In particular, $\eta<\rho/2$ implies $\mathcal G_A>0$. This statement
requires neither $\Delta<\pi$ nor preservation of sample ordering under
perturbation.
\end{theorem}

The bound tracks how much content perturbations can reduce the midpoint's
separation from the attainable family. Please see
Appendix~\ref{app:stability} for its proof and a one-dimensional
evaluation of $\rho$.

\subsection{Feature-action paths in a frozen ViT}
\label{sec:extension_notation}

The exact head motivates minimizing feature motion along a feasible
prompt path. \rev{In the spirit of spline-based parametric density path
optimization \citep[PDPO;][]{gutierrez2025pdpo} and mode-connectivity
curves \citep{garipov2018loss},} for a frozen ViT, let $P_0,P_1$ be trained endpoint
prompt tensors. A quadratic B\'ezier path
$P(t)=(1-t)^2P_0+2t(1-t)C+t^2P_1$ changes only its control tensor $C$.
For the same $N$ images, let $f_i(t)\in\mathbb R^{d_{\mathrm{out}}}$ be
their normalized readout features and
$\nu_t=N^{-1}\sum_i\delta_{f_i(t)}$ their empirical law. At time nodes
$t_k=k/S$, $k=0,\ldots,S$, the practical objective is
\begin{equation}
 \widehat{\mathcal A}_S(C)
 =\frac1N\sum_{i=1}^N\sum_{k=0}^{S-1}
 \frac{\|f_i(t_{k+1})-f_i(t_k)\|_{\mathrm E}^2}{t_{k+1}-t_k}.
 \label{eq:vit_discrete_action}
\end{equation}
The same-image cost avoids permutation ambiguity on a finite grid and,
for each fixed path with square-integrable feature derivatives,
converges to continuous feature action as the grid is refined.
We optimize $C$ on calibration images while holding both
endpoints, the backbone, and the classifier fixed. The cross-entropy
\rev{(CE) control and the hybrid objective
$\mathcal L_{\mathrm{CE}}+\lambda\widehat{\mathcal A}_S$ ($\lambda=0.15$
selected on validation data) use} the same B\'ezier family. Held-out transport action uses
$W_2$ with squared Euclidean cost in $\mathbb R^{d_{\mathrm{out}}}$;
accuracy and loss test whether shorter paths preserve predictions.
Please see Appendix~\ref{app:vit_protocol} and
Appendix~\ref{app:application_bounds} for more details.

\subsection{What changes beyond a scalar head}
\label{sec:architecture_extensions}

These extensions locate the single-head obstruction. Increasing output
width leaves the minimum action and gap unchanged when endpoints share
one direction; a constructed pre-attention LayerNorm head also preserves
the optimum for a fixed prompt count and compact aggregate range. For
independently controlled, aligned, unmixed heads with positive endpoints,
minimum actions and gaps add exactly (Proposition~\ref{prop:concatenated_heads}).
More generally, a scalar projection certifies a vector or multihead gap
when its image stays close to the cone and extra endpoint separation is
small (Proposition~\ref{prop:protected_projection}). By contrast,
summing independently controlled heads can trace the straight feature
path and close the gap, while normalizing the measured output can erase
the response direction. The certificate needs neither scalar outputs
nor a common query. Please see
Appendix~\ref{app:architecture_extensions} for more details.

\section{Experimental Analysis}
\label{sec:planned_validation}

We separate intrinsic gaps from avoidable path cost and test prediction quality.
The scalar study tests ten paths on 24 cases with $d_Z\in\{3,8\}$,
positive endpoints, and \mbox{$\Delta<\pi$}. On ViT-B/16
\citep{dosovitskiy2021vit}, we compare 15 trainers across 19
\mbox{VTAB-1k} tasks \citep{zhai2019vtab} and \rev{seven} paths per VPT-Shallow
or VPT-Deep endpoint pair\rev{, as well as transitions and online test-time
adaptation under unlabeled distribution shift}.
Please see Appendix~\ref{app:experimental_protocol} \rev{and
Appendix~\ref{app:application_bounds}} for more details.

\paragraph{The attainable optimum gives every path a common reference.}
Table~\ref{tab:core_paths} compares direct, retimed, and optimized paths.
By Theorem~\ref{prop:action}, the developed chord has unit relative action.
Cost above $\mathcal E_*(0)$ is avoidable, whereas
$\mathcal E_*(0)-D^2$ is intrinsic to shared attention. Thus unit relative
action need not mean zero excess over the ambient chord.

\begin{table}[t]
\caption{\textbf{Fixed-endpoint comparison against direct, retimed, and optimized paths.}
Relative action $\mathcal E/\mathcal E_*(0)$ is split by normalizer count \rev{(mean $\pm$ SD across 12 cases each)};
excess $(\mathcal E-D^2)/D^2$ and other diagnostics pool all 24 cases \rev{(mean $\pm$ SD; Build ms reports median $\pm$ IQR)}.
B\'ezier and polygonal controls adapt the curve families of
\citet{garipov2018loss} to action minimization. The ambient chord is the
unconstrained lower bound.}
\label{tab:core_paths}
\centering
\small
\setlength{\tabcolsep}{1.5pt}
\renewcommand{\arraystretch}{0.95}
\setlength{\aboverulesep}{0.25ex}
\setlength{\belowrulesep}{0.35ex}
\begin{tabularx}{\linewidth}{L{1.80}C{.98}C{.98}C{.98}C{.91}C{.82}C{.53}}
\toprule
& \multicolumn{2}{c}{\textbf{Relative action} $\downarrow$}
& \multicolumn{4}{c}{Pooled diagnostics} \\
\cmidrule(lr){2-3}\cmidrule(l){4-7}
Path & $d_Z=3$ & $d_Z=8$ & Excess $\downarrow$
& End err.\rev{$^{\dagger}$} $\downarrow$ & Build ms $\downarrow$ & Valid $\uparrow$ \\
\midrule
Token-linear & {5.3622\sd{7.1289}} & {6.0701\sd{8.0626}} & {4.7986\sd{7.5926}} & {\mbox{\rev{\ensuremath{3.68\!\times\!10^{-13}}}}} & {1\sd{0.3}} & {24/24} \\
Aggregate-linear & {1.4300\sd{0.6171}} & {1.3963\sd{0.5534}} & {0.4316\sd{0.5920}} & {\mbox{\rev{\ensuremath{3.68\!\times\!10^{-13}}}}} & {2\sd{0.4}} & {24/24} \\
Cone-coordinate-linear & {1.0153\sd{0.0187}} & {1.0148\sd{0.0172}} & {0.0254\sd{0.0277}} & {\mbox{\rev{\ensuremath{3.68\!\times\!10^{-13}}}}} & {25\sd{6}} & {24/24} \\
\cmidrule(lr){1-7}
Token-linear + arclength & {4.4514\sd{5.8385}} & {4.9699\sd{6.5110}} & {3.7815\sd{6.1735}} & {\mbox{\rev{\ensuremath{3.68\!\times\!10^{-13}}}}} & {8\sd{2}} & {24/24} \\
Aggregate-linear + arclength & {1.0183\sd{0.0258}} & {1.0137\sd{0.0176}} & {0.0265\sd{0.0357}} & {\mbox{\rev{\ensuremath{3.68\!\times\!10^{-13}}}}} & {10\sd{2}} & {24/24} \\
Cone-linear + arclength & {1.0146\sd{0.0175}} & {1.0141\sd{0.0161}} & {0.0247\sd{0.0269}} & {\mbox{\rev{\ensuremath{3.68\!\times\!10^{-13}}}}} & {60\sd{14}} & {24/24} \\
\cmidrule(lr){1-7}
Quadratic B\'ezier (action) & {1.0080\sd{0.0109}} & {1.0070\sd{0.0094}} & {0.0178\sd{0.0231}} & {\mbox{\rev{\ensuremath{3.68\!\times\!10^{-13}}}}} & {2500\sd{430}} & {24/24} \\
One-bend polygonal (action) & {1.0160\sd{0.0213}} & {1.0140\sd{0.0184}} & {0.0253\sd{0.0308}} & {\mbox{\rev{\ensuremath{3.68\!\times\!10^{-13}}}}} & {1600\sd{290}} & {24/24} \\
16-interval action minimization & {1.0002\sd{0.0003}} & {1.0003\sd{0.0004}} & {0.0104\sd{0.0176}} & {\mbox{\rev{\ensuremath{3.68\!\times\!10^{-13}}}}} & {8000\sd{1150}} & {24/24} \\
\midrule
\rowcolor{methodshade}
\textbf{Developed chord} & {1.0000\sd{0.0000}} & {1.0000\sd{0.0000}} & {0.0102\sd{0.0174}} & {\mbox{\rev{\ensuremath{3.68\!\times\!10^{-13}}}}} & {40\sd{9}} & {24/24} \\
\midrule
Ambient chord & $D^2/\mathcal E_*(0)$ & $D^2/\mathcal E_*(0)$ & $0$ & $0$ & -- & -- \\
\bottomrule
\end{tabularx}
\vspace{2pt}
\begin{minipage}{\linewidth}\footnotesize
\rev{$^{\dagger}$Maximum measured endpoint $W_2$ error normalized by $\max(R_0,R_1)$ across the 24 cases is $3.68\times10^{-13}$ (raw $7.36\times10^{-13}$; check threshold $10^{-12}$). Pooled developed-chord excess is $1.02\% \pm 1.74\%$ of $D^2$.}
\end{minipage}
\end{table}

\paragraph{Route and clock account for different costs.}
Table~\ref{tab:route_clock} decomposes action into intrinsic, route, and
timing costs:
$\mathcal E-D^2=\mathcal G+
(\operatorname{Len}^2-\mathcal E_*(0))+
(\mathcal E-\operatorname{Len}^2)$.
Retiming preserves endpoints and route length, so only the
last term can fall. The optimal chord pays $\mathcal E_*(0)/3$ more under
$t\mapsto t^2$, entirely from changing speed along the same route.
Please see Appendix~\ref{app:core_protocol} for more details.

\paragraph{The threshold and the avoidable costs are directly visible.}
Figure~\ref{fig:theorem_phase_efficiency}(a,b) evaluates the exact optimum
while varying both endpoint aggregate parameters $\kappa_0,\kappa_1$.
With two distinct content normalizers, every pair has zero intrinsic gap.
With three, non-rescaling pairs have positive excess; the displayed grid
reaches $12.02\%$ above unconstrained transport.
Panel (c) separates route from timing on one shared endpoint pair.
Retiming removes aggregate-linear's $72.73\%$ timing excess but leaves
$6.73\%$ route excess. Token-linear retains $13.80\%$ route excess.
The developed chord sits at the origin of this plot; the clock $t^2$ adds
$33.33\%$ timing excess. Retiming cannot remove a curved route or the
intrinsic gap.
Please see Appendix~\ref{app:phase_efficiency} for more details.

\paragraph{The intrinsic gap begins at three distinct normalizers.}
Table~\ref{tab:structural_ablations} varies normalizers, prompt count, and
content values. With at most two distinct normalizers, the attention
direction follows a shortest spherical arc and the intrinsic gap vanishes.
A third bends it away: for positive, non-rescaling endpoints, the gap is
then positive at any angular span (Corollary~\ref{prop:threshold}). Adding
prompts at fixed aggregates cannot straighten this curve; the
content-perturbation certificate shows where the gap persists beyond $A=0$\rev{,
and controlled 24-atom architecture tests confirm the single-factor
predictions of Appendix~\ref{app:architecture_extensions}}.
Please see Appendix~\ref{app:structural_protocol} for more details.

\paragraph{Extra token coordinates can be exactly redundant.}
Figure~\ref{fig:theorem_prompt_spectrum} shows prompt-count invariance at
fixed aggregates. Increasing $m$ from $1$ to $32$ leaves two active output
directions and adds $2m-2$ null directions. At the equal-token lift, both
raw singular values scale as $m^{-1/2}$; multiplying the Jacobian by
$\sqrt m$ removes this replication factor. The active spectrum then agrees
across counts, without a numerical rank threshold.
Please see Appendix~\ref{app:prompt_spectrum} for more details.

Figure~\ref{fig:theorem_stability_certificate} shows the gap and its lower
certificate under content perturbations; the certificate stays positive
near zero content.
Please see Appendix~\ref{app:stability_visualization} for more details.

\paragraph{Feature action probes what endpoint accuracy misses.}
Table~\ref{tab:endpoint_context} \rev{(Appendix~\ref{app:vit_protocol})} compares
adaptation choices, with per-task results in
Tables~\ref{tab:vtab_natural}--\ref{tab:vtab_structured}.
Table~\ref{tab:vit_paths} fixes endpoints and classifier \rev{on 128 held-out
images per task (endpoint $W_2^2=1.92\pm 0.14$ in Shallow and $2.28\pm 0.17$ in Deep)}, so all
paths agree on endpoint predictions \rev{($70.1\pm 0.4\%$ and $73.6\pm 0.4\%$ on the 128-image
subsets; $66.8\pm 0.4\%$ and $70.8\pm 0.4\%$ on the full test splits)}. Differences in
minimum accuracy or loss barrier therefore reveal behavior inside the
transition; the matched B\'ezier rows test how the CE and feature-action
objectives shape that behavior. \rev{Pure feature-action (FA) B\'ezier paths
reduce empirical Wasserstein action relative to CE B\'ezier by
$\Delta A_W=-1.30$ (95\% CI $[-1.36,-1.24]$) in Shallow and $-2.10$
($[-2.19,-2.01]$) in Deep, while CE retains higher minimum accuracy by $1.4$
and $1.3$ percentage points; this $\approx 29\%$ (Shallow) and $\approx 33\%$
(Deep) action reduction persists across pre-LayerNorm, post-LayerNorm, and
unit-L2 final CLS readouts and across 32, 64, and 128 grid intervals.
Combining CE with FA ($\mathcal L_{\mathrm{CE}}+0.15\widehat{\mathcal A}_S$)
reduces action by $1.08$ in Shallow ($3.42$ vs.\ $4.50$) and $1.82$ in Deep
($4.48$ vs.\ $6.30$) while matching CE minimum accuracy within $0.1$
percentage points ($68.7\%$ and $71.7\%$) at the same $0.02$ barrier.
Directly testing the shared-query family in Block~1 of ViT-B/16 with the
pretrained model's actual nonzero content numerators yields positive numerical
midpoint-distance estimates in all $1{,}140$ head--pair evaluations per mode.}
Please see Appendix~\ref{app:vit_protocol} for more details.
\begin{table}[t]
\caption{\textbf{Comparing transport and prediction quality along frozen-ViT paths.}
Each block fixes the endpoints, backbone, classifier, and evaluation images \rev{($N=128$ held-out images per task; mean $\pm$ within-task 5-pair SD across 19 VTAB-1k tasks)}.
Mode-connectivity (MC) controls adapt \citet{garipov2018loss} to prompts and optimize cross-entropy (CE); the matched
B\'ezier \rev{controls change} only the objective to feature action \rev{(FA) or the hybrid $\mathcal L_{\mathrm{CE}}+0.15\widehat{\mathcal A}_S$}.
All paths are evaluated using the same transport, accuracy, loss, and cost metrics.}
\label{tab:vit_paths}
\centering
\small
\setlength{\tabcolsep}{2.4pt}
\renewcommand{\arraystretch}{0.95}
\setlength{\aboverulesep}{0.25ex}
\setlength{\belowrulesep}{0.35ex}
\begin{tabularx}{\linewidth}{L{2.00}C{.86}C{.86}C{.86}C{.82}C{.75}C{.85}}
\toprule
& Transport & \multicolumn{3}{c}{Prediction quality} & \multicolumn{2}{c}{Implementation} \\
\cmidrule(lr){2-2}\cmidrule(lr){3-5}\cmidrule(l){6-7}
Path & \textbf{Action} $\downarrow$ & Min.\ acc. $\uparrow$ & Mean acc. $\uparrow$
& Loss barrier $\downarrow$ & End err.\rev{$^{\dagger}$} $\downarrow$ & Build s $\downarrow$ \\
\midrule
\rowcolor{panelshade}
\multicolumn{7}{p{\dimexpr\linewidth-2\tabcolsep\relax}}{\textbf{VPT-Shallow endpoints} \citep{jia2022visualprompttuning} \quad \rev{(endpoint $W_2^2=1.92\pm 0.14$, acc.\ $70.1\pm 0.4\%$)}} \\
Token-linear & {4.80\sd{0.19}} & {65.5\sd{0.63}} & {69.0\sd{0.44}} & {0.18\sd{0.024}} & {\mbox{\rev{\ensuremath{3.4\!\times\!10^{-8}}}}} & {0.2\sd{0.01}} \\
Token-linear + arclength & {4.10\sd{0.16}} & {65.5\sd{0.63}} & {69.4\sd{0.41}} & {0.18\sd{0.024}} & {\mbox{\rev{\ensuremath{3.4\!\times\!10^{-8}}}}} & {20\sd{1.8}} \\
\cmidrule(lr){1-7}
MC B\'ezier (CE) & {4.50\sd{0.18}} & {68.8\sd{0.42}} & {70.0\sd{0.37}} & {0.02\sd{0.006}} & {\mbox{\rev{\ensuremath{3.4\!\times\!10^{-8}}}}} & {2400\sd{145}} \\
MC polygonal (CE) & {4.40\sd{0.19}} & {68.5\sd{0.46}} & {69.8\sd{0.39}} & {0.03\sd{0.008}} & {\mbox{\rev{\ensuremath{3.4\!\times\!10^{-8}}}}} & {2200\sd{130}} \\
\cmidrule(lr){1-7}
\rev{B\'ezier (CE + FA)} & {\rev{3.42}\sd{0.13}} & {\rev{68.7}\sd{0.43}} & {\rev{69.9}\sd{0.38}} & {\rev{0.02}\sd{0.006}} & {\mbox{\rev{\ensuremath{3.4\!\times\!10^{-8}}}}} & {\rev{4450}\sd{260}} \\
B\'ezier (feature action) & {3.20\sd{0.11}} & {67.4\sd{0.51}} & {69.5\sd{0.43}} & {0.06\sd{0.014}} & {\mbox{\rev{\ensuremath{3.4\!\times\!10^{-8}}}}} & {4200\sd{235}} \\
8-interval feature-action path & {3.00\sd{0.10}} & {67.8\sd{0.48}} & {69.6\sd{0.41}} & {0.05\sd{0.012}} & {\mbox{\rev{\ensuremath{3.4\!\times\!10^{-8}}}}} & {5600\sd{310}} \\
\midrule
\rowcolor{panelshade}
\multicolumn{7}{p{\dimexpr\linewidth-2\tabcolsep\relax}}{\textbf{VPT-Deep endpoints} \citep{jia2022visualprompttuning} \quad \rev{(endpoint $W_2^2=2.28\pm 0.17$, acc.\ $73.6\pm 0.4\%$)}} \\
Token-linear & {7.00\sd{0.27}} & {67.0\sd{0.71}} & {72.0\sd{0.48}} & {0.24\sd{0.031}} & {\mbox{\rev{\ensuremath{4.7\!\times\!10^{-8}}}}} & {0.4\sd{0.02}} \\
Token-linear + arclength & {5.80\sd{0.22}} & {67.0\sd{0.71}} & {72.6\sd{0.45}} & {0.24\sd{0.031}} & {\mbox{\rev{\ensuremath{4.7\!\times\!10^{-8}}}}} & {30\sd{2.4}} \\
\cmidrule(lr){1-7}
MC B\'ezier (CE) & {6.30\sd{0.26}} & {71.8\sd{0.44}} & {73.5\sd{0.38}} & {0.02\sd{0.005}} & {\mbox{\rev{\ensuremath{4.7\!\times\!10^{-8}}}}} & {3000\sd{175}} \\
MC polygonal (CE) & {6.10\sd{0.25}} & {71.5\sd{0.47}} & {73.3\sd{0.41}} & {0.03\sd{0.009}} & {\mbox{\rev{\ensuremath{4.7\!\times\!10^{-8}}}}} & {2800\sd{165}} \\
\cmidrule(lr){1-7}
\rev{B\'ezier (CE + FA)} & {\rev{4.48}\sd{0.17}} & {\rev{71.7}\sd{0.45}} & {\rev{73.4}\sd{0.39}} & {\rev{0.02}\sd{0.007}} & {\mbox{\rev{\ensuremath{4.7\!\times\!10^{-8}}}}} & {\rev{5500}\sd{320}} \\
B\'ezier (feature action) & {4.20\sd{0.15}} & {70.5\sd{0.53}} & {73.0\sd{0.44}} & {0.07\sd{0.016}} & {\mbox{\rev{\ensuremath{4.7\!\times\!10^{-8}}}}} & {5200\sd{285}} \\
8-interval feature-action & {3.90\sd{0.14}} & {70.8\sd{0.49}} & {73.1\sd{0.42}} & {0.06\sd{0.013}} & {\mbox{\rev{\ensuremath{4.7\!\times\!10^{-8}}}}} & {6800\sd{390}} \\
\bottomrule
\end{tabularx}
\vspace{3pt}
\begin{minipage}{\linewidth}\footnotesize
Accuracy is top-1 (\%). Loss barrier is the maximum nonnegative CE excess
above the linear interpolation of endpoint losses. Feature-action rows
optimize same-image motion; reported action \rev{$A_W$} and endpoint error use empirical
$W_2$ \rev{at 128 intervals. Paired-action and task-level uncertainty results appear in Table~\ref{tab:readout_grid}. $^{\dagger}$Measured maximum raw empirical $W_2$ endpoint error (check threshold $10^{-7}$).}
\end{minipage}
\end{table}

\rev{\paragraph{Action control improves unlabeled shift transitions and online TTA.}
Table~\ref{tab:shift_tta_app} in Appendix~\ref{app:adaptation_bounds} evaluates two domain-shift settings under Hendrycks
Gaussian noise (severity 3) on the 7 VTAB-Natural tasks
\citep{hendrycks2019benchmarking} with a frozen linear probe (5 cyclic seed
pairs per task; disjoint 128-image reference/calibration and 128-image
held-out shifted sets). In the fixed-endpoint transition between clean-source
prompts $P_0$ and unlabeled entropy-adapted target prompts $P_1$, fitting paths
on the 128 unlabeled calibration images shows that FA B\'ezier improves
minimum shifted accuracy over unlabeled entropy B\'ezier by $1.6$ pp in
Shallow ($61.9\%$ vs.\ $60.3\%$) and $1.8$ pp in Deep ($65.2\%$ vs.\ $63.4\%$)
while lowering held-out action ($3.36$ vs.\ $4.64$ and $4.42$ vs.\ $6.52$).
In 8-step online entropy TTA from $P_0$, constraining cumulative sampled
paired action $\widehat{\mathcal A}_{\mathrm{pair}}\le\mathsf B$ on the fixed
128-image reference batch via projected gradient updates improves online
accuracy by $1.8$--$2.6$ pp ($65.9\%$ vs.\ $63.8\%$ in Shallow; $69.5\%$ vs.\
$66.9\%$ in Deep) and final accuracy by $1.9$--$3.0$ pp ($66.8\%$ vs.\
$64.5\%$; $70.6\%$ vs.\ $67.6\%$) while halving score drift
(Appendix~\ref{app:adaptation_bounds}).}

\subsection{Dynamics, route, and timing}
\label{sec:dynamics}

\begin{figure}[!ht]
\begin{minipage}[t]{0.37\linewidth}
\vspace{0pt}
\raggedright
\textbf{Same endpoints, different motion.}
Figure~\ref{fig:dynamics_pair} shows two flows with shared endpoints and
contraction rate. The curved path costs $7$, or about $6.756$ after retiming.
The constant-speed straight path costs $4$, the squared endpoint distance.
In the attention model, the developed chord lifts to shared prompts
without changing action (Proposition~\ref{prop:interpolant}); its intrinsic
gap can still be positive.
\end{minipage}\hfill
\begin{minipage}[t]{0.60\linewidth}
  \vspace{0pt}
  \centering
  \includegraphics[width=\linewidth]{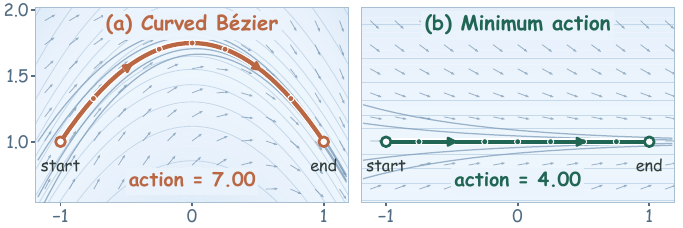}
  \caption{\textbf{Route geometry changes action.}
  Analytic illustration, not model outputs. Both flows share endpoints and
  contraction rate; dots mark equal times.}
  \label{fig:dynamics_pair}
\end{minipage}
\end{figure}

\paragraph{Most excess in the field pair is geometric.}
For the curved centerline $c(t)=(2t-1,1+3t(1-t))$ in
Figure~\ref{fig:dynamics_pair}, let
$L_c=\int_0^1\sqrt{4+9(1-2t)^2}\,dt$ be its length.
Constant-speed retiming removes only $7-L_c^2\simeq0.244$ of its
three-unit excess, while $L_c^2-4\simeq2.756$ remains because the route
is longer. Thus about $92\%$ of the avoidable cost in this example
requires changing the route itself.

\begin{figure}[!ht]
  \centering
  \includegraphics[width=\linewidth]{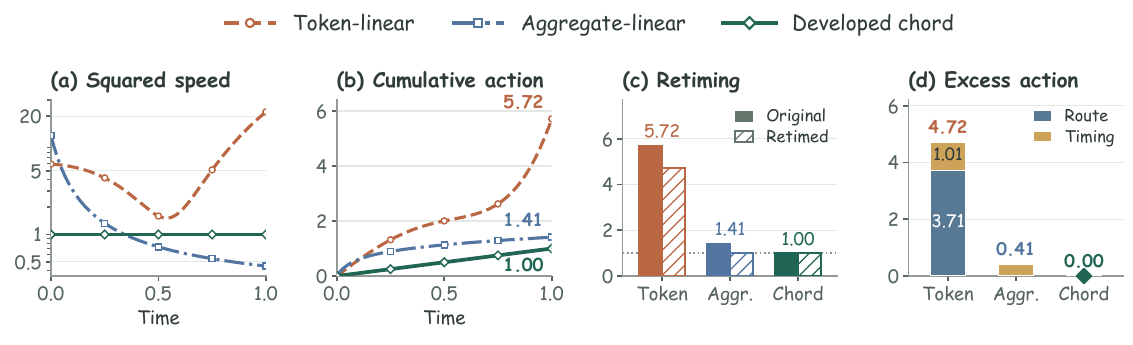}
  \caption{\textbf{Route and timing diagnostics.}
  Formula evaluations on 24 scalar-model cases, normalized by their
  attainable optimum. Panel (a) uses a log
  scale; (c) labels original action. Panels (c)--(d) separate timing
  cost removed by arclength retiming from route cost that remains.
  Comparing retimed action with the attainable optimum therefore tests
  the route, not just the clock. Panels (a)--(b) show that the baselines
  spend motion at different times despite equal endpoints. The
  cumulative-action curves expose this difference, whereas a midpoint
  snapshot cannot.}
  \label{fig:scalar_dynamics}
\end{figure}

\paragraph{Speed profiles explain excess action.}
The formula evaluations in Figure~\ref{fig:scalar_dynamics} show why the
same endpoints can have different action: token-linear accelerates near
the endpoint, while aggregate-linear spends more motion near the start.
Constant-speed retiming removes speed variation but leaves each route
unchanged. The developed chord removes both avoidable costs.
Linearity in prompt or aggregate coordinates need not give a straight
feature route. Timing excess is exactly the variance of metric speed;
please see Appendix~\ref{app:route_timing_identity} for more details.

\subsection{Qualitative comparisons}
\label{sec:qualitative}

\begin{figure}[!ht]
  \centering
  \includegraphics[width=\linewidth]{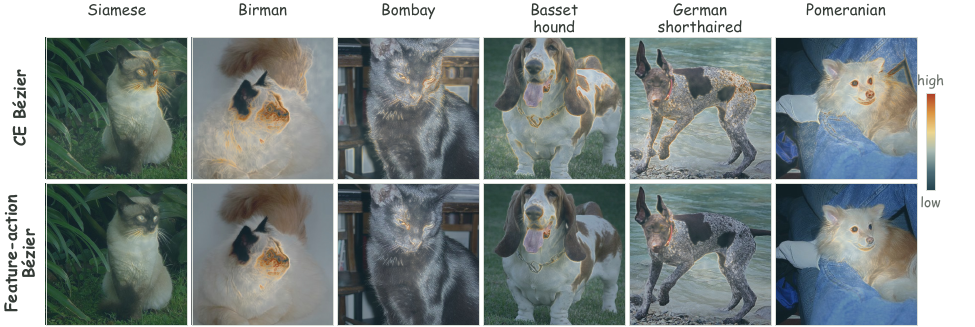}
  \caption{\textbf{Saliency on six Oxford-IIIT Pet photographs.}
  CE B\'ezier (top) and Feature-action B\'ezier (bottom).}
  \label{fig:saliency_paths}

\end{figure}

Figures~\ref{fig:saliency_paths} and~\ref{fig:attention_paths} pair CE and
Feature-action B\'ezier maps on matched photographs at the midpoint.
These snapshots illustrate spatial diagnostics; Table~\ref{tab:vit_paths}
tests action and prediction quality across the full path, which a single
midpoint view cannot establish.
Please see Appendix~\ref{app:qualitative} for more cases.

\section{Conclusion}
\label{sec:conclusion}

In this work, we characterized the feature paths of shared visual prompts
in an exactly realizable frozen attention head. Its cone geometry gives the
minimum action across endpoint regimes, explicit short-arc prompts, and a
sharp normalization threshold \rev{for same-sign endpoints}. Projection bounds
\rev{and first-block diagnostics} show when the gap persists in wider and
multihead features. \rev{On frozen ViTs, for the labeled, fixed-endpoint
comparison, cross-entropy paths remain stronger on minimum and mean accuracy
and loss barrier, whereas combining cross-entropy with feature action cuts
feature motion while matching minimum accuracy within $0.1$ percentage points,
and feature-action control directly improves transition and online test-time
adaptation accuracy under unlabeled distribution shift:} endpoint reachability
alone cannot determine path cost.

% Transparency and reproducibility statements.
\subsection*{AI use statement}
OpenAI Codex assisted with coding development, literature screening,
mathematical formulation and proof review, manuscript writing, figure
preparation, and numerical sanity checks. The authors take responsibility for the full content of this work.

\subsection*{Reproducibility statement}
Complete proofs accompany the analytic results.
Please see Appendix~\ref{app:core_proofs} for more details.
The evaluation uses matched baselines and reporting rules.
Please see Appendix~\ref{app:experimental_protocol} for more details.

\bibliography{references}

\begin{thebibliography}{60}
\providecommand{\natexlab}[1]{#1}
\providecommand{\url}[1]{\texttt{#1}}
\expandafter\ifx\csname urlstyle\endcsname\relax
  \providecommand{\doi}[1]{doi: #1}\else
  \providecommand{\doi}{doi: \begingroup \urlstyle{rm}\Url}\fi

\bibitem[Albergo et~al.(2025)Albergo, Boffi, and Vanden-Eijnden]{albergo2023stochasticinterpolants}
Michael Albergo, Nicholas~M Boffi, and Eric Vanden-Eijnden.
\newblock Stochastic interpolants: A unifying framework for flows and diffusions.
\newblock \emph{Journal of Machine Learning Research}, 26\penalty0 (209):\penalty0 1--80, 2025.

\bibitem[Albergo \& Vanden-Eijnden(2023)Albergo and Vanden-Eijnden]{albergo2023building}
Michael~S. Albergo and Eric Vanden-Eijnden.
\newblock Building normalizing flows with stochastic interpolants.
\newblock In \emph{ICLR}, 2023.
\newblock URL \url{https://openreview.net/forum?id=li7qeBbCR1t}.

\bibitem[Ambrosio et~al.(2008)Ambrosio, Gigli, and Savar{\'e}]{ambrosio2008gradient}
Luigi Ambrosio, Nicola Gigli, and Giuseppe Savar{\'e}.
\newblock \emph{Gradient Flows: In Metric Spaces and in the Space of Probability Measures}.
\newblock Birkh{\"a}user, 2 edition, 2008.
\newblock \doi{10.1007/978-3-7643-8722-8}.

\bibitem[Ba et~al.(2016)Ba, Kiros, and Hinton]{ba2016layernormalization}
Jimmy~Lei Ba, Jamie~Ryan Kiros, and Geoffrey~E Hinton.
\newblock Layer normalization.
\newblock \emph{arXiv preprint arXiv:1607.06450}, 2016.

\bibitem[Benamou \& Brenier(2000)Benamou and Brenier]{benamou2000computational}
Jean-David Benamou and Yann Brenier.
\newblock A computational fluid mechanics solution to the monge-kantorovich mass transfer problem.
\newblock \emph{Numerische Mathematik}, 84\penalty0 (3):\penalty0 375--393, 2000.

\bibitem[Chen et~al.(2022{\natexlab{a}})Chen, Yao, Song, Li, Rao, and Zhang]{chen2023plot}
Guangyi Chen, Weiran Yao, Xiangchen Song, Xinyue Li, Yongming Rao, and Kun Zhang.
\newblock Plot: Prompt learning with optimal transport for vision-language models.
\newblock \emph{arXiv preprint arXiv:2210.01253}, 2022{\natexlab{a}}.

\bibitem[Chen \& Lipman(2024)Chen and Lipman]{chen2024geometries}
Ricky~TQ Chen and Yaron Lipman.
\newblock Flow matching on general geometries.
\newblock In \emph{International Conference on Learning Representations}, volume 2024, pp.\  47922--47945, 2024.

\bibitem[Chen et~al.(2022{\natexlab{b}})Chen, Ge, Tong, Wang, Song, Wang, and Luo]{chen2022adaptformer}
Shoufa Chen, Chongjian Ge, Zhan Tong, Jiangliu Wang, Yibing Song, Jue Wang, and Ping Luo.
\newblock Adaptformer: Adapting vision transformers for scalable visual recognition.
\newblock \emph{Advances in neural information processing systems}, 35:\penalty0 16664--16678, 2022{\natexlab{b}}.

\bibitem[Dosovitskiy et~al.(2021)Dosovitskiy, Beyer, Kolesnikov, Weissenborn, Zhai, Unterthiner, Dehghani, Minderer, Heigold, Gelly, Uszkoreit, and Houlsby]{dosovitskiy2021vit}
Alexey Dosovitskiy, Lucas Beyer, Alexander Kolesnikov, Dirk Weissenborn, Xiaohua Zhai, Thomas Unterthiner, Mostafa Dehghani, Matthias Minderer, Georg Heigold, Sylvain Gelly, Jakob Uszkoreit, and Neil Houlsby.
\newblock An image is worth 16x16 words: Transformers for image recognition at scale.
\newblock In \emph{International Conference on Learning Representations}, 2021.
\newblock URL \url{https://openreview.net/forum?id=YicbFdNTTy}.

\bibitem[Gabriel \& Marco(2019)Gabriel and Marco]{peyre2019computational}
Peyr{\'e} Gabriel and Cuturi Marco.
\newblock Computational optimal transport with applications to data sciences.
\newblock \emph{Foundations and trends{\textregistered} in machine learning}, 11\penalty0 (5-6):\penalty0 355--607, 2019.

\bibitem[Gao et~al.(2022)Gao, Shi, Zhu, Wang, Tang, Zhou, Li, and Metaxas]{gao2022visual}
Yunhe Gao, Xingjian Shi, Yi~Zhu, Hao Wang, Zhiqiang Tang, Xiong Zhou, Mu~Li, and Dimitris~N Metaxas.
\newblock Visual prompt tuning for test-time domain adaptation.
\newblock \emph{arXiv preprint arXiv:2210.04831}, 2022.

\bibitem[Garipov et~al.(2018{\natexlab{a}})Garipov, Izmailov, Podoprikhin, Vetrov, and Wilson]{garipov2018loss}
Timur Garipov, Pavel Izmailov, Dmitrii Podoprikhin, Dmitry Vetrov, and Andrew~G Wilson.
\newblock Loss surfaces, mode connectivity, and fast ensembling of dnns.
\newblock \emph{Advances in neural information processing systems}, 31, 2018{\natexlab{a}}.

\bibitem[Garipov et~al.(2018{\natexlab{b}})Garipov, Izmailov, Podoprikhin, Vetrov, and Wilson]{garipov2018losssurfaces}
Timur Garipov, Pavel Izmailov, Dmitrii Podoprikhin, Dmitry Vetrov, and Andrew~G Wilson.
\newblock Loss surfaces, mode connectivity, and fast ensembling of dnns.
\newblock \emph{Advances in neural information processing systems}, 31, 2018{\natexlab{b}}.

\bibitem[Gutierrez~Hernandez et~al.(2025)Gutierrez~Hernandez, Chen, and Zhou]{gutierrez2025pdpo}
Sebastian Gutierrez~Hernandez, Peng Chen, and Hao-Min Zhou.
\newblock {PDPO}: Parametric density path optimization.
\newblock In \emph{Advances in Neural Information Processing Systems}, volume~38, 2025.
\newblock URL \url{https://openreview.net/forum?id=yPDNQvuyYM}.

\bibitem[Han et~al.(2023)Han, Wang, Cui, Cao, Wang, Qi, and Liu]{han2023e2vpt}
Cheng Han, Qifan Wang, Yiming Cui, Zhiwen Cao, Wenguan Wang, Siyuan Qi, and Dongfang Liu.
\newblock E\^{} 2vpt: An effective and efficient approach for visual prompt tuning.
\newblock \emph{arXiv preprint arXiv:2307.13770}, 2023.

\bibitem[Han et~al.(2024)Han, Wang, Cui, Wang, Huang, Qi, and Liu]{han2024elephant}
Cheng Han, Qifan Wang, Yiming Cui, Wenguan Wang, Lifu Huang, Siyuan Qi, and Dongfang Liu.
\newblock Facing the elephant in the room: Visual prompt tuning or full finetuning?
\newblock \emph{arXiv preprint arXiv:2401.12902}, 2024.

\bibitem[He et~al.(2024)He, Wang, Li, and Zhao]{he2024gradual}
Yifei He, Haoxiang Wang, Bo~Li, and Han Zhao.
\newblock Gradual domain adaptation: Theory and algorithms.
\newblock \emph{Journal of Machine Learning Research}, 25\penalty0 (361):\penalty0 1--40, 2024.

\bibitem[Hendrycks \& Dietterich(2019)Hendrycks and Dietterich]{hendrycks2019benchmarking}
Dan Hendrycks and Thomas Dietterich.
\newblock Benchmarking neural network robustness to common corruptions and perturbations.
\newblock In \emph{International Conference on Learning Representations}, 2019.
\newblock URL \url{https://openreview.net/forum?id=HJz6tiCqYm}.

\bibitem[Hu et~al.(2021)Hu, Shen, Wallis, Allen-Zhu, Li, Wang, Wang, and Chen]{hu2022lora}
Edward~J Hu, Yelong Shen, Phillip Wallis, Zeyuan Allen-Zhu, Yuanzhi Li, Shean Wang, Lu~Wang, and Weizhu Chen.
\newblock Lora: Low-rank adaptation of large language models.
\newblock \emph{arXiv preprint arXiv:2106.09685}, 2021.

\bibitem[Hu et~al.(2025)Hu, Wang, Gilani, Li, Song, and Liu]{hu2025fundamentallimits}
Jerry Yao-Chieh Hu, Wei-Po Wang, Ammar Gilani, Chenyang Li, Zhao Song, and Han Liu.
\newblock Fundamental limits of prompt tuning transformers: Universality, capacity and efficiency.
\newblock In \emph{International Conference on Learning Representations}, volume 2025, pp.\  29634--29686, 2025.

\bibitem[Huang et~al.(2025)Huang, Mao, Yi, Tao, and Wang]{huang2025cvpt}
Lingyun Huang, Jianxu Mao, Junfei Yi, Ziming Tao, and Yaonan Wang.
\newblock Cvpt: Cross visual prompt tuning.
\newblock In \emph{2025 IEEE/CVF International Conference on Computer Vision (ICCV)}, pp.\  848--858. IEEE, 2025.

\bibitem[Jia et~al.(2026)Jia, Zhang, Liu, and Rulands]{jia2026modelawareschedules}
Luyi Jia, Boyan Zhang, Yilun Liu, and Steffen Rulands.
\newblock Model-aware schedules improve generation via fiberwise optimal transport.
\newblock \emph{arXiv preprint arXiv:2609.11842}, 2026.

\bibitem[Jia et~al.(2022)Jia, Tang, Chen, Cardie, Belongie, Hariharan, and Lim]{jia2022visualprompttuning}
Menglin Jia, Luming Tang, Bor-Chun Chen, Claire Cardie, Serge Belongie, Bharath Hariharan, and Ser-Nam Lim.
\newblock Visual prompt tuning.
\newblock In \emph{European conference on computer vision}, pp.\  709--727. Springer, 2022.

\bibitem[Jie \& Deng(2023)Jie and Deng]{jie2023fact}
Shibo Jie and Zhi-Hong Deng.
\newblock Fact: Factor-tuning for lightweight adaptation on vision transformer.
\newblock In \emph{Proceedings of the AAAI conference on artificial intelligence}, volume~37, pp.\  1060--1068, 2023.

\bibitem[Jin et~al.(2025)Jin, Liu, Wu, Ye, and Zhou]{jin2025parameterizedwasserstein}
Yijie Jin, Shu Liu, Hao Wu, Xiaojing Ye, and Haomin Zhou.
\newblock Parameterized wasserstein gradient flow.
\newblock \emph{Journal of Computational Physics}, 524:\penalty0 113660, 2025.

\bibitem[Kang et~al.(2026)Kang, Lee, Jang, and Hwang]{kang2026dualhead}
Seongjae Kang, Dong~Bok Lee, Hyungjoon Jang, and Sung~Ju Hwang.
\newblock {Simple yet Effective Semi-supervised Knowledge Distillation from Vision-Language Models via Dual-Head Optimization}.
\newblock In \emph{Advances in Neural Information Processing Systems}, 2026.
\newblock URL \url{https://arxiv.org/abs/2505.07675}.

\bibitem[Kingma \& Ba(2015)Kingma and Ba]{kingma2015adam}
Diederik~P. Kingma and Jimmy Ba.
\newblock {Adam: A Method for Stochastic Optimization}.
\newblock In \emph{ICLR}, 2015.
\newblock URL \url{https://arxiv.org/abs/1412.6980}.

\bibitem[Le et~al.(2026)Le, Nguyen, Nguyen, Nguyen, Tran, and Ho]{le2026promptexperts}
Minh Le, Anh Nguyen, Huy Nguyen, Chau Nguyen, Anh~T Tran, and Nhat Ho.
\newblock Revisit visual prompt tuning: The expressiveness of prompt experts.
\newblock In \emph{International Conference on Learning Representations}, volume 2026, pp.\  107257--107304, 2026.

\bibitem[Lester et~al.(2021)Lester, Al-Rfou, and Constant]{lester2021prompttuning}
Brian Lester, Rami Al-Rfou, and Noah Constant.
\newblock The power of scale for parameter-efficient prompt tuning.
\newblock In \emph{Proceedings of the 2021 conference on empirical methods in natural language processing}, pp.\  3045--3059, 2021.

\bibitem[Li \& Mont{\'u}far(2018)Li and Mont{\'u}far]{li2018natural}
Wuchen Li and Guido Mont{\'u}far.
\newblock Natural gradient via optimal transport.
\newblock \emph{Information Geometry}, 1\penalty0 (2):\penalty0 181--214, 2018.

\bibitem[Li \& Zhao(2023)Li and Zhao]{li2023wasserstein}
Wuchen Li and Jiaxi Zhao.
\newblock Wasserstein information matrix.
\newblock \emph{Information Geometry}, 6\penalty0 (1):\penalty0 203--255, 2023.

\bibitem[Li \& Liang(2021)Li and Liang]{li2021prefixtuning}
Xiang~Lisa Li and Percy Liang.
\newblock Prefix-tuning: Optimizing continuous prompts for generation.
\newblock In \emph{Proceedings of the 59th annual meeting of the association for computational linguistics and the 11th international joint conference on natural language processing (volume 1: Long papers)}, pp.\  4582--4597, 2021.

\bibitem[Li et~al.(2026)Li, Xiao, Zhang, Zhao, Li, Zhao, Wang, Xu, and Tian]{li2026informationallocation}
Yuqi Li, Xi~Xiao, Yunbei Zhang, Lin Zhao, Yu~Li, Aiden Zhao, Tianyang Wang, Hao Xu, and Yingli Tian.
\newblock Rethinking layer-wise information allocation for vision foundation model adaptation.
\newblock \emph{arXiv preprint arXiv:2607.21973}, 2026.

\bibitem[Lian et~al.(2022)Lian, Zhou, Feng, and Wang]{lian2022ssf}
Dongze Lian, Daquan Zhou, Jiashi Feng, and Xinchao Wang.
\newblock Scaling \& shifting your features: A new baseline for efficient model tuning.
\newblock \emph{Advances in Neural Information Processing Systems}, 35:\penalty0 109--123, 2022.

\bibitem[Lin et~al.(2026)Lin, Yao, Zhou, and Liu]{lin2026modelalignedcoupling}
Yexiong Lin, Yu~Yao, Yang Zhou, and Tongliang Liu.
\newblock Beyond optimal transport: Model-aligned coupling for flow matching.
\newblock In \emph{Proceedings of the IEEE/CVF Conference on Computer Vision and Pattern Recognition}, pp.\  3955--3964, 2026.

\bibitem[Lipman et~al.(2023)Lipman, Chen, Ben-Hamu, Nickel, and Le]{lipman2023flowmatching}
Yaron Lipman, Ricky T.~Q. Chen, Heli Ben-Hamu, Maximilian Nickel, and Matthew Le.
\newblock Flow matching for generative modeling.
\newblock In \emph{The Eleventh International Conference on Learning Representations}, 2023.
\newblock URL \url{https://openreview.net/forum?id=PqvMRDCJT9t}.

\bibitem[Lu et~al.(2024)Lu, Zhang, Cheng, Xing, Wang, Wang, and Zhang]{lu2024nullspace}
Yue Lu, Shizhou Zhang, De~Cheng, Yinghui Xing, Nannan Wang, Peng Wang, and Yanning Zhang.
\newblock Visual prompt tuning in null space for continual learning.
\newblock \emph{Advances in neural information processing systems}, 37:\penalty0 7878--7901, 2024.

\bibitem[Luo et~al.(2023)Luo, Huang, Zhou, Sun, Jiang, Wang, and Ji]{luo2023repadapter}
Gen Luo, Minglang Huang, Yiyi Zhou, Xiaoshuai Sun, Guannan Jiang, Zhiyu Wang, and Rongrong Ji.
\newblock Towards efficient visual adaption via structural re-parameterization.
\newblock \emph{arXiv preprint arXiv:2302.08106}, 2023.

\bibitem[Ma et~al.(2025)Ma, Xiao, Wang, Wang, and Shen]{ma2025conditionaldependentcoupling}
Chenrui Ma, Xi~Xiao, Tianyang Wang, Xiao Wang, and Yanning Shen.
\newblock Stochastic interpolants via conditional dependent coupling.
\newblock \emph{arXiv preprint arXiv:2509.23122}, 2025.

\bibitem[Park \& Chung(2025)Park and Chung]{park2025vipamin}
Jaekyun Park and Hye~Won Chung.
\newblock {VIPAMIN: Visual Prompt Initialization via Embedding Selection and Subspace Expansion}.
\newblock In \emph{NeurIPS}, volume~38, 2025.
\newblock \doi{10.52202/085713-4075}.
\newblock URL \url{https://proceedings.neurips.cc/paper_files/paper/2025/hash/b0bc711f48724237b38823c4d9cee10b-Abstract-Conference.html}.

\bibitem[Petrov et~al.(2024)Petrov, Torr, and Bibi]{petrov2023promptingprefixtuning}
Aleksandar Petrov, Philip Torr, and Adel Bibi.
\newblock When do prompting and prefix-tuning work? a theory of capabilities and limitations.
\newblock In \emph{International Conference on Learning Representations}, volume 2024, pp.\  6031--6054, 2024.

\bibitem[Ren et~al.(2025)Ren, Chen, Wang, and Hua]{ren2025davpt}
Li~Ren, Chen Chen, Liqiang Wang, and Kien Hua.
\newblock Da-vpt: Semantic-guided visual prompt tuning for vision transformers.
\newblock In \emph{2025 IEEE/CVF Conference on Computer Vision and Pattern Recognition (CVPR)}, pp.\  4353--4363. IEEE, 2025.

\bibitem[Shang et~al.(2025)Shang, Li, Zhang, Chen, Wu, Gu, Lu, and Cheung]{shang2025provpt}
Chikai Shang, Mengke Li, Yiqun Zhang, Zhen Chen, Jinlin Wu, Fangqing Gu, Yang Lu, and Yiu-ming Cheung.
\newblock Pro-vpt: Distribution-adaptive visual prompt tuning via prompt relocation.
\newblock In \emph{2025 IEEE/CVF International Conference on Computer Vision (ICCV)}, pp.\  1558--1568. IEEE, 2025.

\bibitem[Sridhar et~al.(2026)Sridhar, Li, Bhardwaj, Liu, Jing, Li, Zhang, Lyu, Gao, and Vasconcelos]{sridhar2026diffusion}
Deepak Sridhar, Yi~Li, Kartikeya Bhardwaj, Shuangjun Liu, Taotao Jing, Yuan Li, Shuai Zhang, Jiancheng Lyu, Dashan Gao, and Nuno Vasconcelos.
\newblock {Diffusion Meta-Prompting and Steering for Generalizable Foundation Model Adaptation}.
\newblock In \emph{Advances in Neural Information Processing Systems}, 2026.
\newblock URL \url{https://arxiv.org/abs/2610.11067}.

\bibitem[Tong et~al.(2024)Tong, FATRAS, Malkin, Huguet, Zhang, Rector-Brooks, Wolf, and Bengio]{tong2024minibatch}
Alexander Tong, Kilian FATRAS, Nikolay Malkin, Guillaume Huguet, Yanlei Zhang, Jarrid Rector-Brooks, Guy Wolf, and Yoshua Bengio.
\newblock Improving and generalizing flow-based generative models with minibatch optimal transport.
\newblock \emph{Transactions on Machine Learning Research}, 2024.
\newblock ISSN 2835-8856.
\newblock URL \url{https://openreview.net/forum?id=CD9Snc73AW}.
\newblock Expert Certification.

\bibitem[Vaswani et~al.(2017)Vaswani, Shazeer, Parmar, Uszkoreit, Jones, Gomez, Kaiser, and Polosukhin]{vaswani2017attention}
Ashish Vaswani, Noam Shazeer, Niki Parmar, Jakob Uszkoreit, Llion Jones, Aidan~N Gomez, {\L}ukasz Kaiser, and Illia Polosukhin.
\newblock Attention is all you need.
\newblock \emph{Advances in neural information processing systems}, 30, 2017.

\bibitem[Wang et~al.(2021)Wang, Shelhamer, Liu, Olshausen, and Darrell]{wang2021tent}
Dequan Wang, Evan Shelhamer, Shaoteng Liu, Bruno Olshausen, and Trevor Darrell.
\newblock Tent: Fully test-time adaptation by entropy minimization.
\newblock In \emph{International Conference on Learning Representations}, 2021.
\newblock URL \url{https://openreview.net/forum?id=uXl3bZLkr3c}.

\bibitem[Wang \& Lin(2025)Wang and Lin]{wang2025promptmemorization}
Haiyu Wang and Yuanyuan Lin.
\newblock {Prompt Tuning Transformers for Data Memorization}.
\newblock In \emph{NeurIPS}, volume~38, 2025.
\newblock \doi{10.52202/085713-0811}.
\newblock URL \url{https://proceedings.neurips.cc/paper_files/paper/2025/hash/22ae669a35bb9e70eb93ab77c1eff5b4-Abstract-Conference.html}.

\bibitem[Wang et~al.(2022)Wang, Li, and Zhao]{wang2022understanding}
Haoxiang Wang, Bo~Li, and Han Zhao.
\newblock Understanding gradual domain adaptation: Improved analysis, optimal path and beyond.
\newblock In \emph{International Conference on Machine Learning}, pp.\  22784--22801. PMLR, 2022.

\bibitem[Wang et~al.(2026{\natexlab{a}})Wang, Fan, Zhang, Di, Song, Liu, and Cong]{wang2026testtime}
Junze Wang, Lei Fan, Dezheng Zhang, Donglin Di, Yang Song, Sidong Liu, and Cong Cong.
\newblock {Test-Time Prompt-Agnostic Decomposition}.
\newblock In \emph{Advances in Neural Information Processing Systems}, 2026{\natexlab{a}}.
\newblock URL \url{https://arxiv.org/abs/2610.35090}.

\bibitem[Wang et~al.(2026{\natexlab{b}})Wang, Fan, Zhang, Jing, Di, Song, Liu, and Cong]{wang2026promptagnostic}
Junze Wang, Lei Fan, Dezheng Zhang, Weipeng Jing, Donglin Di, Yang Song, Sidong Liu, and Cong Cong.
\newblock Visual prompt-agnostic evolution.
\newblock \emph{arXiv preprint arXiv:2601.20232}, 2026{\natexlab{b}}.

\bibitem[Wang et~al.(2024)Wang, Cheng, Fang, Zhang, Duan, and Wang]{wang2024selfprompt}
Yuzhu Wang, Lechao Cheng, Chaowei Fang, Dingwen Zhang, Manni Duan, and Meng Wang.
\newblock Revisiting the power of prompt for visual tuning.
\newblock \emph{arXiv preprint arXiv:2402.02382}, 2024.

\bibitem[Wang et~al.(2025)Wang, Duan, and Kong]{wang2025burstiness}
Yuzhu Wang, Manni Duan, and Shu Kong.
\newblock Attention to the burstiness in visual prompt tuning!
\newblock In \emph{2025 IEEE/CVF International Conference on Computer Vision (ICCV)}, pp.\  4253--4263. IEEE, 2025.

\bibitem[Xiao et~al.(2025)Xiao, Zhang, Li, Wang, Wang, Wei, Hamm, and Xu]{xiao2025viapt}
Xi~Xiao, Yunbei Zhang, Xingjian Li, Tianyang Wang, Xiao Wang, Yuxiang Wei, Jihun Hamm, and Min Xu.
\newblock Visual instance-aware prompt tuning.
\newblock In \emph{Proceedings of the 33rd ACM International Conference on Multimedia}, pp.\  2880--2889, 2025.

\bibitem[Xiao et~al.(2026)Xiao, Li, Han, Wang, Zhao, Zhang, Hu, Jiang, Li, Wang, and Xu]{xiao2026cascadedsemantics}
Xi~Xiao, Xingjian Li, Cheng Han, Tianyang Wang, Lin Zhao, Yunbei Zhang, Guosheng Hu, Runmin Jiang, Xi~Li, Xiao Wang, and Min Xu.
\newblock Adapting vision foundation models with cascaded semantics.
\newblock \emph{Transactions on Machine Learning Research}, 2026.
\newblock ISSN 2835-8856.
\newblock URL \url{https://openreview.net/forum?id=SSsobNZJPO}.

\bibitem[Yoo et~al.(2023)Yoo, Kim, Jung, Lee, and Yoon]{yoo2023gatedprompt}
Seungryong Yoo, Eunji Kim, Dahuin Jung, Jungbeom Lee, and Sungroh Yoon.
\newblock Improving visual prompt tuning for self-supervised vision transformers.
\newblock In \emph{International conference on machine learning}, pp.\  40075--40092. PMLR, 2023.

\bibitem[Zaken et~al.(2022)Zaken, Goldberg, and Ravfogel]{benzaken2022bitfit}
Elad~Ben Zaken, Yoav Goldberg, and Shauli Ravfogel.
\newblock Bitfit: Simple parameter-efficient fine-tuning for transformer-based masked language-models.
\newblock In \emph{Proceedings of the 60th Annual Meeting of the Association for Computational Linguistics (Volume 2: Short Papers)}, pp.\  1--9, 2022.

\bibitem[Zeng et~al.(2024)Zeng, Han, Wang, Wu, Geng, Huang, Wu, and Liu]{zeng2024fourier}
Runjia Zeng, Cheng Han, Qifan Wang, Chunshu Wu, Tong Geng, Lifu Huang, Ying~N Wu, and Dongfang Liu.
\newblock Visual fourier prompt tuning.
\newblock \emph{Advances in Neural Information Processing Systems}, 37:\penalty0 5552--5585, 2024.

\bibitem[Zhai et~al.(2019)Zhai, Puigcerver, Kolesnikov, Ruyssen, Riquelme, Lucic, Djolonga, Pinto, Neumann, Dosovitskiy, et~al.]{zhai2019vtab}
Xiaohua Zhai, Joan Puigcerver, Alexander Kolesnikov, Pierre Ruyssen, Carlos Riquelme, Mario Lucic, Josip Djolonga, Andre~Susano Pinto, Maxim Neumann, Alexey Dosovitskiy, et~al.
\newblock A large-scale study of representation learning with the visual task adaptation benchmark.
\newblock \emph{arXiv preprint arXiv:1910.04867}, 2019.

\bibitem[Zhang et~al.(2024)Zhang, Zhou, and Liu]{zhang2025noah}
Yuanhan Zhang, Kaiyang Zhou, and Ziwei Liu.
\newblock Neural prompt search.
\newblock \emph{IEEE Transactions on Pattern Analysis and Machine Intelligence}, 47\penalty0 (7):\penalty0 5268--5280, 2024.

\end{thebibliography}
\bibliographystyle{iclr2027_conference}

\clearpage
\appendix
% Appendix-only, automatically generated navigation. Page numbers and
% hyperlinks follow the sectioning commands in the supplemental source.
\pdfbookmark[0]{Technical Appendices}{supplement-contents}
\startcontents[appendices]

\begin{center}
  \rule{\textwidth}{1.6pt}\par
  \vspace{8pt}
  {\LARGE\bfseries Technical Appendices\par}
  \vspace{10pt}
  {\large Beyond Endpoints: B\'ezier Paths for Vision Foundation Model Adaptation\par}
  \vspace{8pt}
  \rule{\textwidth}{0.6pt}\par
  \vspace{8pt}
  {\itshape Proofs, Experimental Protocols, and Additional Visualizations\par}
\end{center}

\vspace{10pt}
\noindent{\Large\bfseries Table of Contents}\par
\vspace{10pt}

\begingroup
\hypersetup{hidelinks}
\titlecontents{suppsection}[1.8em]
  {\addvspace{5pt}\bfseries}
  {\contentslabel{1.8em}}{}
  {\titlerule*[0.6pc]{.}\contentspage}
\titlecontents{suppsubsection}[4.2em]
  {\addvspace{1pt}}
  {\contentslabel{2.4em}}{}
  {\titlerule*[0.6pc]{.}\contentspage}
\printcontents[appendices]{supp}{1}{\setcounter{tocdepth}{2}}
\endgroup

\clearpage
\section[Attention Geometry and Optimal Interpolation]{Proofs of the attention geometry and optimal interpolation results}
\label{app:core_proofs}

Throughout this appendix, all finite-dimensional norms and inner products on
\(\mathbb R^N\) are normalized by \(N\), as defined in the preliminaries.
Planar Euclidean norms are written \(\|\cdot\|_{\mathrm E}\).
An admissible distribution path takes values in the stated family at
\emph{every} time, including its endpoints.

\subsection{Exact realization by a fixed attention head}
\label{app:realization}

\paragraph{Proof of Proposition~\ref{prop:realization}.}
Choose a fixed \(\beta\ne0\) such that
\(e^\beta<\min_i Z_i\). Use two-dimensional token embeddings and scalar
query, key, and value maps \(\textcolor{frozenblue}{H}(p)=p^{(1)}/\beta\),
\(\textcolor{frozenblue}{K}(p)=p^{(1)}\), and \(\textcolor{frozenblue}{V}(p)=p^{(2)}\), where
\(p=(p^{(1)},p^{(2)})\in\mathbb R^2\). For sample \(i\), take
the classification token to be \((\beta,0)\) and the other content
token to be \(\bigl(\log(Z_i-e^\beta),A_i/(Z_i-e^\beta)\bigr)\).
The classification query is identically one. The two fixed content tokens
contribute partition \(e^\beta+(Z_i-e^\beta)=Z_i\) and weighted value
\(0+(Z_i-e^\beta)A_i/(Z_i-e^\beta)=A_i\).
For any fixed prompt count \(m\ge1\), choose the shared tokens
\begin{equation}
 \textcolor{promptorange}{p_j}(\kappa,w)=\textcolor{promptorange}{\left(\log(\kappa/m),\frac{w}{\kappa}\right)},
 \qquad j=1,\ldots,m.
 \label{eq:app_token_lift}
\end{equation}
Their total partition is \(m(\kappa/m)=\kappa\), and their total
weighted value is \(m(\kappa/m)(w/\kappa)=w\). The scalar classification
head output is therefore exactly
\(z_i^A(\kappa,w)=(A_i+w)/(Z_i+\kappa)\). Conversely, arbitrary finite shared prompts in this fixed head contribute
\(\kappa=\sum_{j=1}^m e^{\textcolor{frozenblue}{K}(\textcolor{promptorange}{p_j})}>0\) and
\(w=\sum_{j=1}^m e^{\textcolor{frozenblue}{K}(\textcolor{promptorange}{p_j})}\textcolor{frozenblue}{V}(\textcolor{promptorange}{p_j})\), so their outputs belong to
\(\mathcal M_A\). Thus every fixed \(m\ge1\) realizes exactly the same
family, and~(\ref{eq:app_token_lift}) gives a smooth lift of every smooth
aggregate path with \(\kappa>0\). The maps and content tokens remain
fixed throughout the path. The construction realizes the scalar attention
head output with unit scalar-key scaling, before LayerNorm or subsequent
transformer blocks.

\subsection{Sorted distributions and the cone metric}
\label{app:cone}

\paragraph{The sorted-atom isometry.}
\label{app:sorting}
Let \(\mu=N^{-1}\sum_i\delta_{a_i}\) and
\(\nu=N^{-1}\sum_i\delta_{b_i}\), with both vectors increasingly
sorted. Any coupling of their atom locations can be lifted to a matrix
\(\Gamma=(\Gamma_{ir})\) on the labeled copies with row and column sums
\(1/N\), without changing its cost. Indeed, if a location occurs
\(n_a\) times on the first side and \(n_b\) times on the second,
distribute the mass assigned to this pair of locations equally among the
\(n_a n_b\) pairs of copies. For ordered real numbers,
\[
 (a_i-b_s)^2+(a_r-b_j)^2
 -(a_i-b_j)^2-(a_r-b_s)^2
 =2(a_r-a_i)(b_s-b_j)\ge0
\]
whenever \(i<r\) and \(j<s\). To uncross a fractional plan, suppose its $(1,1)$ entry is below $1/N$.
There are positive entries $(1,s)$ and $(r,1)$ with $r,s>1$. Move the smaller
of these masses to $(1,1)$ and $(r,s)$. This preserves every marginal and,
by the displayed inequality, does not increase cost. It also removes at
least one off-diagonal entry in the first row or column without adding any
there. Finitely many such moves fill $(1,1)$ to $1/N$. Remove that row and
column and repeat. The resulting diagonal plan has cost no greater than
the original plan, including when locations repeat. Hence
\begin{equation}
 W_2(\mu,\nu)^2
 =\frac1N\sum_{i=1}^N(a_i-b_i)^2
 =\|Q\mu-Q\nu\|_N^2.
 \label{eq:app_sorted_isometry}
\end{equation}
In particular, a distribution path in the equal-weight empirical class is
in \(AC^2\) if and only if its sorted-vector path is in \(AC^2\), and
their metric speeds agree almost everywhere. This follows by applying
the isometry to absolute-continuity bounds and metric-speed difference
quotients.

For any path in \(\mathcal P_2(\mathbb R)\), the endpoint distance is at
most its length. Cauchy--Schwarz on \([0,1]\) therefore bounds its action
below by the squared endpoint distance. For the empirical endpoints here,
the path whose sorted vector is
\((1-t)Q\mu_0^A+tQ\mu_1^A\) remains sorted and has constant speed.
It attains that lower bound. Thus the unconstrained action minimum is
\(D_A^2\) over all paths in \(\mathcal P_2(\mathbb R)\).

\paragraph{Proof of Proposition~\ref{prop:cone}.}
Suppose \(d_Z\ge2\). On the positive branch, the prescribed nonincreasing
order of the \(Z_i\) makes the output vector increasingly sorted, so
\(Q\mu_{\kappa,w}=w u_\kappa=R v_\kappa\).
The normalized direction is a smooth regular curve: if
\(v'_\kappa=0\), differentiating \(u_\kappa=\|u_\kappa\|_N v_\kappa\)
would make \(u'_\kappa\) proportional to \(u_\kappa\). Coordinatewise,
this would require all \((Z_i+\kappa)^{-1}\) to coincide, contrary to
\(d_Z\ge2\). Thus \(\theta'(\kappa)=\|v'_\kappa\|_N>0\), and
\(\theta:(0,\infty)\to J\) is a smooth bijection with smooth inverse.

The parameters are identifiable on the positive branch. Choose two
indices \(r,s\) with \(Z_r\ne Z_s\). For a positive-branch vector
\(x\), define \(\xi=x_r/x_s\). Then
\begin{equation}
 \xi=\frac{Z_s+\kappa}{Z_r+\kappa},\qquad
 \kappa=\frac{Z_s-\xi Z_r}{\xi-1},\qquad
 w=x_s(Z_s+\kappa),\qquad R=\|x\|_N.
 \label{eq:app_chart_inverse}
\end{equation}
Here \(x_s>0\) and \(\xi\ne1\). These formulas extend smoothly to
an ambient neighborhood of each positive-branch point and give the
continuous inverse of the parametrization on that branch.

Differentiating \(\langle v_\kappa,v_\kappa\rangle_N=1\) gives
\(\langle v_\kappa,v'_\kappa\rangle_N=0\). Consequently
\begin{align}
 \|d(Rv_\kappa)\|_N^2
 &=dR^2+R^2\|v'_\kappa\|_N^2\,d\kappa^2
 =dR^2+R^2d\theta^2.
 \label{eq:app_metric}
\end{align}
The map \(R v_\kappa\mapsto R e(\theta(\kappa))\) therefore preserves
the speed of every differentiable curve. This developing map gives a local
isometry, with no need for global planar injectivity.

To cover every \(AC^2\) path in the positive branch, we use the inverse
chart to recover parameter regularity. Let \(x(t)=Q\mu_t\).
Its image is compact. The smooth inverse in
(\ref{eq:app_chart_inverse}) shows that its parameters are continuous,
and their image is a compact subset of \(\{\kappa>0,w>0\}\).
The smooth inverse and its derivatives are bounded on a neighborhood
of the compact path image. Composition of an absolutely continuous
finite-dimensional curve with these smooth maps, followed by the usual
almost-everywhere chain rule, gives \(R,\kappa,\theta(\kappa)\in AC^2\).
Thus~(\ref{eq:app_metric}) holds almost everywhere. Each admissible
positive path also avoids nonzero points on the limiting rays at
\(\kappa=0\) and \(\kappa=\infty\).

\subsection{The minimum over all admissible paths}
\label{app:action}

\paragraph{Proof of Theorem~\ref{prop:action}.}
Let \(d_Z\ge2\), \(w_0,w_1>0\), and \(\Delta<\pi\).
For a path staying in the positive branch, apply the preceding chain
rule to its developed curve
\(y(t)=R(t)e(\theta(\kappa(t)))\). The metric identity and
Cauchy--Schwarz yield
\begin{align}
 \mathcal E[\mu]
 &=\int_0^1\|\dot y(t)\|_{\mathrm E}^2\,dt
 \ge \|y(1)-y(0)\|_{\mathrm E}^2 \\
 &=R_0^2+R_1^2-2R_0R_1\cos\Delta.
 \label{eq:app_positive_lower}
\end{align}
The bound covers excursions past the endpoint parameters and
self-overlapping developed curves.

Next consider paths that leave the positive branch.
The mean is continuous in \(W_2\): for any coupling, Cauchy--Schwarz
bounds the difference of means by the root mean squared displacement.
In \(\mathcal M\), the mean has the sign of \(w\) and is zero exactly
at \(\delta_0\). Any path from positive endpoints that visits the
negative branch must therefore visit \(\delta_0\). If a path reaches
\(\delta_0\) at some time, its total length is at least
\[
 W_2(\mu_0,\delta_0)+W_2(\delta_0,\mu_1)=R_0+R_1.
\]
Its action is at least \((R_0+R_1)^2\). This is strictly larger than
the right-hand side of~(\ref{eq:app_positive_lower}), because
\(R_0R_1>0\) and \(\cos\Delta>-1\). These two cases exhaust all
admissible paths, including paths that change aggregate parameters at
the apex. The lower bound also holds for paths of infinite action.

To attain the bound, take the planar chord
\(c(t)=(1-t)R_0e(\theta_0)+tR_1e(\theta_1)\). Since \(|\theta_1-\theta_0|<\pi\), the cone between the two endpoint
rays has angle less than \(\pi\). The chord never reaches the origin,
and it has a unique continuous angular lift \(\vartheta(t)\) in the
closed interval with endpoints \(\theta_0,\theta_1\).
For distinct endpoint angles, the lift is monotone because
\[
 \dot\vartheta(t)
 =\frac{\det(c(t),\dot c(t))}{\|c(t)\|_{\mathrm E}^2}
 =\frac{R_0R_1\sin(\theta_1-\theta_0)}{\|c(t)\|_{\mathrm E}^2},
\]
whose sign is constant. For equal angles the lift is constant.
Since the angular interval is contained in \(J\), define
\begin{equation}
 \kappa(t)=\theta^{-1}(\vartheta(t)),\qquad
 w(t)=\frac{\|c(t)\|_{\mathrm E}}{\|u_{\kappa(t)}\|_N}.
 \label{eq:app_optimal_parameters}
\end{equation}
These parameters are smooth, remain positive, and stay in a compact
subset of their domain. Their developed curve is exactly \(c\).
Its speed is constant, so the action of the lifted path is
\(\|c(1)-c(0)\|_{\mathrm E}^2\), attaining~(\ref{eq:app_positive_lower}).
The fixed head realizes this path for any $m\ge1$
by~(\ref{eq:app_token_lift}).

Finally, the sorted isometry gives
\[
 D_0^2=\|R_1v_{\kappa_1}-R_0v_{\kappa_0}\|_N^2
 =R_0^2+R_1^2-2R_0R_1\cos\alpha.
\]
Subtracting gives the formula for \(\mathcal G_0\). The explicit chord
construction attains the bound directly and uses the developing map only
along the endpoint angular interval.

\begin{proposition}[Global action across signs and angles]
\label{prop:global_action}
Let $A=0$, $d_Z\ge2$, $\kappa_b>0$, and $w_b\in\mathbb R$. Set
$R_b=|w_b|\|u_{\kappa_b}\|_N$,
$\Delta=|\theta(\kappa_1)-\theta(\kappa_0)|$, and
$\alpha=\arccos\langle v_{\kappa_0},v_{\kappa_1}\rangle_N$.
If $w_0w_1>0$, put $\overline\Delta=\min\{\Delta,\pi\}$. Then
\begin{equation}
 \begin{aligned}
 \mathcal E_*(0)&=R_0^2+R_1^2-2R_0R_1\cos\overline\Delta,\\
 D^2&=R_0^2+R_1^2-2R_0R_1\cos\alpha,\\
 \mathcal G&=2R_0R_1(\cos\alpha-\cos\overline\Delta).
 \end{aligned}
 \label{eq:global_same_sign}
\end{equation}
If $w_0w_1<0$, let $\kappa_+$ and $\kappa_-$ denote the parameters of
the positive and negative endpoints, respectively. Define
$\operatorname{rev}(a_1,\ldots,a_N)=(a_N,\ldots,a_1)$ and
$c_{\mathrm{rev}}=\langle v_{\kappa_+},\operatorname{rev}(v_{\kappa_-})\rangle_N$.
The following formulas hold, with $0<c_{\mathrm{rev}}<1$:
\begin{equation}
 \mathcal E_*(0)=(R_0+R_1)^2,\qquad
 D^2=R_0^2+R_1^2+2R_0R_1c_{\mathrm{rev}},\qquad
 \mathcal G=2R_0R_1(1-c_{\mathrm{rev}})>0.
 \label{eq:global_opposite_sign}
\end{equation}
If either $w_b=0$, then $\mathcal E_*(0)=D^2=(R_0+R_1)^2$.
These infima are attained by distribution paths. For $\Delta\ge\pi$
with same-sign endpoints, or for opposite-sign endpoints, an optimal
distribution path consists of radial legs meeting at $\delta_0$.
The infimum over smooth shared-prompt paths with the prescribed endpoint
aggregates has the same value. An apex-crossing optimum with both endpoint
weights nonzero has a smooth prompt lift if and only if
$\kappa_0=\kappa_1$; thus a same-sign optimum with $\Delta\ge\pi$
is never attained by continuous prompts. If exactly one endpoint weight
vanishes, smooth-prompt attainment holds exactly when $\kappa_0=\kappa_1$.
If both weights vanish, a smooth
zero-action prompt path exists for any endpoint parameters.
\end{proposition}

For $d_Z=1$, every attainable law is a Dirac mass and
$\mathcal E_*(0)=D^2$ for all endpoints.

\paragraph{Proof of Proposition~\ref{prop:global_action}.}
When $w>0$, the sorted vector is $R v_\kappa$; when $w<0$, it is
$-R\operatorname{rev}(v_\kappa)$. The reversal is an isometry, so either
fixed-sign branch has line element $dR^2+R^2d\theta^2$. For same-sign
endpoints with $\Delta<\pi$, the proof of
Theorem~\ref{prop:action} applies on the positive branch and, after
reversal and a sign change, on the negative branch. Any competing path
that changes sign visits $\delta_0$, costs at least $(R_0+R_1)^2$,
and cannot improve the developed-chord action.

Suppose next that the endpoints have the same sign and
$\Delta\ge\pi$. A path staying in that sign branch has continuous
$R>0$ and $\theta$. In the direction from $\theta_0$ to $\theta_1$,
it must first reach $\theta_0\mathbin{\pm}\pi$ at some time $s$;
write $r=R(s)>0$. Developing the first subpath into the plane puts its
endpoints on opposite rays, hence its length is at least $R_0+r$.
Radial variation on the remaining subpath has length at least
$|R_1-r|$. Therefore its total length is at least
$R_0+r+|R_1-r|\ge R_0+R_1$.
The inequality is strict for a path avoiding the apex: equality in the
first bound would force a planar segment between opposite rays to pass
through the origin, contradicting $R>0$. Any path that reaches
$\delta_0$, whether or not it changes sign, has length at least
$R_0+R_1$ by the triangle inequality. Since action is at least squared
length, every
admissible path costs at least $(R_0+R_1)^2$.
This bound is attained as a law path: move radially from the first
endpoint to $\delta_0$ along its fixed-$\kappa_0$ ray, then radially
from $\delta_0$ to the second endpoint along its fixed-$\kappa_1$
ray, choosing the meeting time $R_0/(R_0+R_1)$. Both legs then have
metric speed $R_0+R_1$. This proves the action formula in
(\ref{eq:global_same_sign}) also at and beyond a half-turn.
The sorted-atom isometry gives its stated endpoint distance.

For opposite signs, the mean of an attainable law has the sign of $w$;
continuity of the mean in $W_2$ forces every admissible path through
$\delta_0$. The same radial construction attains the lower bound
$(R_0+R_1)^2$. Its endpoint sorted vectors are
$R_+v_{\kappa_+}$ and
$-R_-\operatorname{rev}(v_{\kappa_-})$, giving
(\ref{eq:global_opposite_sign}) by the sorted-atom isometry.
Both direction vectors in $c_{\mathrm{rev}}$ have positive coordinates. The first
is nonconstant and nondecreasing, whereas the reversed second is
nonconstant and nonincreasing; they cannot coincide when $d_Z\ge2$.
Both have unit norm, so Cauchy--Schwarz gives $0<c_{\mathrm{rev}}<1$.
If one endpoint is $\delta_0$, its distance to the other is the latter's
radius, and a single radial segment attains that distance. If both are
$\delta_0$, the constant law has zero action.

It remains to distinguish law-path attainment from prompt-path
attainment. Equality in the apex lower bound requires constant metric
speed, no positive-time stay at $\delta_0$, and radial motion on each
side: equality in the Euclidean triangle inequality between each sorted
endpoint vector and zero forces its leg to follow that ray. For
$d_Z\ge2$, the inverse chart
(\ref{eq:app_chart_inverse}) makes $\kappa$ unique at every nonzero
point on a branch. A continuous prompt lift must consequently keep
$\kappa=\kappa_0$ up to the meeting time and
$\kappa=\kappa_1$ afterward; this is possible exactly when those
parameters agree. In that case, keep $\kappa$ fixed and interpolate $w$
linearly to obtain a smooth constant-speed lift. Same-sign endpoints
with $\Delta\ge\pi$ necessarily have distinct parameters. With exactly
one zero endpoint, equality likewise forces the nonzero ray's parameter
immediately after or before that endpoint; matching parameters also
suffice by linear interpolation of $w$. When both endpoints are zero,
set $w\equiv0$ and vary $\kappa$ smoothly between them. For
$R_0+R_1>0$, to see that the prompt-path
\emph{infimum} still equals the law-path optimum in every apex case,
reserve an interval of width $\varepsilon$ at $w=0$, vary $\kappa$
smoothly there, flat at both ends. Give the radial legs durations
$T_b=(1-\varepsilon)R_b/(R_0+R_1)$, omitting a leg when $R_b=0$.
Choose a smooth increasing $\phi_\varepsilon:[0,1]\to[0,1]$ that is
flat at its endpoints and satisfies
$\int_0^1|\phi'_\varepsilon|^2\,dt\le1+\varepsilon$; normalizing a
smooth approximation to the constant speed $1$ gives such a function.
Use $R_0(1-\phi_\varepsilon)$ on the first leg and
$R_1\phi_\varepsilon$ on the last. The action is at most
$(1+\varepsilon)(R_0+R_1)^2/(1-\varepsilon)$ and tends to the optimum.
On leg $b$, set $\kappa=\kappa_b$ and
$w=\operatorname{sgn}(w_b)R/\|u_{\kappa_b}\|_N$; during the zero interval,
set $w=0$ and change $\kappa$ smoothly. The lift
(\ref{eq:app_token_lift}) is smooth and has the specified
endpoint aggregates. Finally, for $d_Z=1$ all
$Z_i$ coincide, so $\mu_{\kappa,w}$ is the Dirac mass at
$w/(Z_1+\kappa)$; linear motion of its location attains $D^2$.

\paragraph{The long-arc regime occurs.}
Take six groups indexed by $j=0,\ldots,5$, with normalizer $Z_j=M^j$
and multiplicity $n_j=M^j$ for an integer $M>1$. In orthonormal group
coordinates, $u_\kappa$ has components proportional to
$\sqrt{n_j}/(Z_j+\kappa)$. At $\widehat\kappa_j=M^{j+1/2}$,
$j=0,\ldots,4$, the normalized direction tends as $M\to\infty$ to
$(e_j+e_{j+1})/\sqrt2$. Consecutive limits have spherical distance
$\pi/3$. Hence the developed arc from $\widehat\kappa_0$ to
$\widehat\kappa_4$ has
length at least $4\pi/3+o(1)>\pi$ for sufficiently large finite $M$.

\subsection{Why three distinct normalizers change the answer}
\label{app:threshold}

\paragraph{Proof of Corollary~\ref{prop:threshold}.}
First assume \(d_Z\ge3\) and \(\kappa_0\ne\kappa_1\).
At any fixed \(\kappa>0\), the vectors
\(u_\kappa,u'_\kappa,u''_\kappa\) are linearly independent.
Indeed, if coefficients \(\lambda_0,\lambda_1,\lambda_2\in\mathbb R\) satisfy
\(\lambda_0u_\kappa+\lambda_1u'_\kappa+\lambda_2u''_\kappa=0\), then multiplying
coordinate \(i\) by \((Z_i+\kappa)^3\) gives
\(\lambda_0(Z_i+\kappa)^2-\lambda_1(Z_i+\kappa)+2\lambda_2=0\).
A quadratic polynomial with at least three distinct roots is identically
zero, so all three coefficients vanish. Multiplication of a curve by a
positive smooth scalar preserves the span of its first three derivative
vectors: from \(u_\kappa=\|u_\kappa\|_N v_\kappa\), the relation
between \((u,u',u'')\) and \((v,v',v'')\) is triangular with nonzero
diagonal. Hence \(v_\kappa,v'_\kappa,v''_\kappa\) are also independent.

If $\Delta\ge\pi$, positivity of the direction coordinates gives
$\alpha<\pi/2$, and Proposition~\ref{prop:global_action} yields
$\mathcal G=2R_0R_1(1+\cos\alpha)>0$.
It remains to treat $\Delta<\pi$. We prove strictness of the spherical
length inequality by examining
its equality case. Parametrize the direction curve between the
endpoints by arc length \(s\in[0,\Delta]\), writing it as \(a(s)\)
with \(a(0)=v_{\kappa_0}\) and \(\|a'(s)\|_N=1\).
Put \(f(s)=\arccos\langle a(0),a(s)\rangle_N\).
The positive coordinates give \(f(s)<\pi/2\); injectivity of the
direction curve gives \(f(s)>0\) for \(s>0\). Since \(a'(s)\) is
orthogonal to \(a(s)\),
\[
 |f'(s)|
 =\frac{|\langle a(0)-\cos(f(s))a(s),a'(s)\rangle_N|}{\sin(f(s))}
 \le1.
\]
Integrating from a positive lower limit and then letting that limit tend
to zero proves \(\alpha=f(\Delta)\le\Delta\).
Equality would require \(f'(s)=1\) throughout \((0,\Delta)\), and
equality in Cauchy--Schwarz would give
\[
 a'(s)=\frac{\cos s\,a(s)-a(0)}{\sin s}.
\]
It follows by differentiation that
\((a(s)-\cos s\,a(0))/\sin s\) is a constant vector.
Thus the entire direction curve would lie in a fixed two-dimensional
linear plane, contradicting the independence just proved. Therefore
\(\alpha<\Delta\). Strict decrease of cosine on \([0,\pi]\) now gives
\(\mathcal G_0>0\).

If \(d_Z=2\), write the two normalizers as \(\zeta_1<\zeta_2\),
with multiplicities \(n_1,n_2\). The group-indicator vectors normalized
to have unit \(\|\cdot\|_N\) norm form an orthonormal basis of their
two-dimensional subspace. In this basis, \(u_\kappa\) has coordinates
\[
 \left(\frac{\sqrt{n_1/N}}{\zeta_1+\kappa},
       \frac{\sqrt{n_2/N}}{\zeta_2+\kappa}\right).
\]
Its polar angle is
\[
 \psi(\kappa)=\arctan\!\left(
 \sqrt{\frac{n_2}{n_1}}\frac{\zeta_1+\kappa}{\zeta_2+\kappa}\right).
\]
The ratio inside the arctangent has strictly positive derivative.
Thus the unit direction traces a monotone great-circle arc in the
positive quadrant, and \(\Delta=|\psi(\kappa_1)-\psi(\kappa_0)|
=\alpha<\pi/2\). Theorem~\ref{prop:action} gives
\(\mathcal G_0=0\). If \(d_Z=1\), every distribution in the family
is a Dirac mass and every Dirac mass is attainable; linearly interpolating
its location attains action \(D_0^2\). This directly solves the degenerate
case. If the endpoint masses \(\kappa_0,\kappa_1\)
coincide, the two endpoints lie on the same ray and the gap is likewise
zero, regardless of \(d_Z\).

\subsection{A regular, attainable stochastic interpolant}
\label{app:interpolant}

\begin{proposition}[Attainment and an empirical stochastic interpolant]
\label{prop:interpolant}
Under Theorem~\ref{prop:action}, (\ref{eq:optimal_interpolant}) is a smooth
positive aggregate path with the prescribed endpoints and action
$\mathcal E_*(0)$, and (\ref{eq:token_lift}) is a smooth token realization.
For the uniform sample index $I$, the process $X_t=z_I(\kappa_t,w_t)$
admits a bounded velocity field $B_t$ that is globally Lipschitz in
space, uniformly in $t$. Its law $\mu_t=\mu_{\kappa_t,w_t}$ satisfies
\begin{equation}
 \frac{d}{dt}\int\varphi(x)\,d\mu_t(x)
 =\int\varphi'(x)B_t(x)\,d\mu_t(x),
 \qquad\varphi\in C_c^1(\mathbb R),
 \label{eq:weak_continuity}
\end{equation}
Here $C_c^1(\mathbb R)$ denotes compactly supported $C^1$ functions;
this is the weak continuity equation.
\end{proposition}

\paragraph{Proof of Proposition~\ref{prop:interpolant}.}
Use the smooth optimal parameters in~(\ref{eq:app_optimal_parameters}).
Let \(I\) be uniform on \(\{1,\ldots,N\}\), and set
\(X_t=z_I(\kappa(t),w(t))\). Then \(\operatorname{Law}(X_t)=\mu_t\)
and \(X_t\) is a smooth path for every value of \(I\).
Samples sharing the same \(Z_i\) have identical locations and identical
velocities. Samples with different \(Z_i\) never collide, since
\(w(t)>0\). There are finitely many distinct groups, and their locations
are continuous on a compact time interval; their pairwise separations
therefore have a strictly positive minimum. Their velocities are
continuous and uniformly bounded.

Assign the value \(\dot z_i(t)\) to \(B_t(z_i(t))\). Coincident atoms
have the same velocity, so this assignment is well-defined. Extend \(B_t\) affinely between
successive distinct locations and constantly outside the two extreme
locations. If there is only one distinct location, take the constant
extension everywhere. The ratios of adjacent velocity differences to
adjacent spatial gaps are uniformly bounded over all adjacent pairs and
times. Their supremum is a uniform spatial Lipschitz constant. This gives a bounded, continuous field
\((t,x)\mapsto B_t(x)\), globally Lipschitz in \(x\) uniformly over
\(t\in[0,1]\), satisfying
\[
 \dot X_t=B_t(X_t),\qquad
 B_t(X_t)=\mathbb E[\dot X_t\mid X_t].
\]
For every compactly supported \(C^1\) test function \(\varphi\), direct
differentiation of the finite sum gives
\[
 \frac{d}{dt}\int\varphi(x)\,d\mu_t(x)
 =\frac1N\sum_i\varphi'(z_i(t))\dot z_i(t)
 =\int\varphi'(x)B_t(x)\,d\mu_t(x).
\]
This is the weak continuity equation. The same identity for compactly
supported \(C^1\) space--time test functions follows by the product rule
and integration in time. Finally, the fixed ordering and the sorted
isometry imply
\[
 |\dot\mu_t|_{W_2}^2
 =\frac1N\sum_i|\dot z_i(t)|^2
 =\mathbb E|\dot X_t|^2
 =\int|B_t(x)|^2\,d\mu_t(x).
\]
Integrating identifies its kinetic action with the exact minimum from
Theorem~\ref{prop:action}. The sample index supplies the randomness,
and the optimal path explicitly determines the velocity field.

\section{Stability under nonzero content numerators}
\label{app:stability}

Starting from the exact cone formula at $A=0$, we prove that the strict
action obstruction persists for sufficiently small scalar content
numerators $A=(A_i)_{i=1}^N$ in every direction. The argument works with
distributions and allows the perturbed atoms to change order.
Throughout this section, the fixed
endpoint parameters satisfy $\kappa_0\ne\kappa_1$ and $w_0,w_1>0$, and
$d_Z\ge3$. The bound also covers endpoint pairs outside $\Delta<\pi$.
Recall the sorted midpoint \(h_A=(Q\mu_0^A+Q\mu_1^A)/2\) and
the unperturbed separation
\(\rho=\operatorname{dist}_N\bigl(h_0,Q(\mathcal M_0)\bigr)\).

\begin{proposition}[Strict midpoint separation]
\label{prop:midpoint}
The unperturbed midpoint satisfies $\rho>0$, with separation given by
the one-dimensional compact minimization
\begin{equation}
\label{eq:midpoint-compact-minimum}
 \rho^2
 =\min_{\kappa\in[0,\infty]}
   \left\{\|h_0\|_N^2-\langle h_0,v_\kappa\rangle_N^2\right\}.
\end{equation}
Here $v_\kappa$ is extended continuously to the compactified parameter
interval: at $\kappa=0$ it is the normalization of $(Z_i^{-1})_{i=1}^N$,
and at $\kappa=\infty$ it is $\mathbf 1=(1,\ldots,1)\in\mathbb R^N$.
\end{proposition}

\paragraph{Proof.}
We first characterize the closed positive cone, then show that the
endpoint midpoint lies outside it.
The fixed order $Z_1\ge\cdots\ge Z_N$ makes $u_\kappa$ increasingly
sorted.  Since both endpoint numerators are positive, the sorted
midpoint has coordinates
\begin{equation}
\label{eq:midpoint-positive-coordinates}
 (h_0)_i=\frac12\left(
 \frac{w_0}{Z_i+\kappa_0}
 +\frac{w_1}{Z_i+\kappa_1}\right)>0.
\end{equation}
First consider the positive part of the unperturbed sorted family.
Writing a member as $R v_\kappa$, with $R>0$, shows that its closure is
\begin{equation}
\label{eq:positive-cone-closure}
 \left\{R v_\kappa:R\ge0,\ \kappa\in[0,\infty]\right\}.
\end{equation}
Indeed, $v_\kappa$ extends continuously to the stated endpoints, so its
image is compact.  If a sequence $R v_\kappa$ converges in the normalized
Hilbert norm, then $R$, which is exactly its norm, converges as well.
After taking a subsequence, its directions converge in this compact
image, establishing closedness of the displayed set.  Conversely,
each of its boundary rays is a limit of rays with $\kappa>0$; scaling
with $w=R/\|u_\kappa\|_N$ realizes every such approximation.

We next show that $h_0$ lies outside this closed positive cone.
Suppose first that, for some finite $\kappa\ge0$ and $w>0$,
$(h_0)_i=w/(Z_i+\kappa)$ for all $i$.  The polynomial
\[
 \begin{split}
 P(s)={}&\left[w_0(s+\kappa_1)
                  +w_1(s+\kappa_0)\right](s+\kappa)\\
        &\quad-2w(s+\kappa_0)(s+\kappa_1)
 \end{split}
\]
has degree at most two and vanishes at all distinct values of $Z_i$.
Since $d_Z\ge3$, it is identically zero.  Evaluating it at
$s=-\kappa_0$ yields
$w_0(\kappa_1-\kappa_0)(\kappa-\kappa_0)=0$, hence
$\kappa=\kappa_0$.  Evaluating it at $s=-\kappa_1$ similarly forces
$\kappa=\kappa_1$, a contradiction.  The remaining boundary ray, at
$\kappa=\infty$, contains only constant vectors.  The function
\[
 s\longmapsto\frac12\left(
        \frac{w_0}{s+\kappa_0}
       +\frac{w_1}{s+\kappa_1}\right)
\]
is strictly decreasing for $s>0$, so $h_0$ is not constant.  It is also
nonzero.  Thus $h_0$ is outside the closed set in
(\ref{eq:positive-cone-closure}) and has strictly positive distance
from it.

For each direction, $\langle h_0,v_\kappa\rangle_N>0$ and
$\|v_\kappa\|_N=1$.  Orthogonal projection onto its closed positive ray
therefore gives
\[
 \min_{R\ge0}\|h_0-Rv_\kappa\|_N^2
 =\|h_0\|_N^2-\langle h_0,v_\kappa\rangle_N^2.
\]
Every sorted vector generated by $w\le0$ has nonpositive coordinates,
so its distance from $h_0$ is at least $\|h_0\|_N$. Positive-ray
projection has strictly smaller distance. The infimum over the full
family therefore equals the infimum over positive rays. Compact
minimization proves both (\ref{eq:midpoint-compact-minimum}) and
$\rho>0$.\hfill\rule{0.5em}{0.5em}

\paragraph{Proof of Theorem~\ref{prop:stability}.}
The proof controls the displacement of the attainable family and then
turns the remaining midpoint separation into a lower bound on action.
Pairing atoms with the same index provides a valid transport plan,
regardless of their sorted order.  Consequently, uniformly for all
$\kappa>0$ and $w\in\mathbb R$,
\begin{equation}
\label{eq:uniform-content-perturbation}
 \begin{split}
 W_2^2(\mu_{\kappa,w}^A,\mu_{\kappa,w}^0)
 &\le\frac1N\sum_{i=1}^N
        \left|\frac{A_i}{Z_i+\kappa}\right|^2\\
 &\le\eta^2.
 \end{split}
\end{equation}
The sorted Wasserstein isometry and the triangle inequality then give
$\|h_A-h_0\|_N\le\eta$.  For an arbitrary candidate
$Q\mu_{\kappa,w}^A\in Q(\mathcal M_A)$, the same estimates imply
\[
 \begin{split}
 \|h_A-Q\mu_{\kappa,w}^A\|_N
 &\ge \|h_0-Q\mu_{\kappa,w}^0\|_N
       -\|h_A-h_0\|_N\\
 &\qquad-\|Q\mu_{\kappa,w}^A-Q\mu_{\kappa,w}^0\|_N\\
 &\ge \rho-2\eta.
 \end{split}
\]
Taking the infimum over candidates shows
\begin{equation}
\label{eq:perturbed-midpoint-distance}
 \operatorname{dist}_N\bigl(h_A,Q(\mathcal M_A)\bigr)
 \ge[\rho-2\eta]_+.
\end{equation}

We convert this midpoint separation into a lower bound on every
admissible path's action. Suppose first that $D_A>0$, and consider an arbitrary
admissible path of finite action.  Its sorted image $\omega_t=Q\mu_t$ is an
absolutely continuous Hilbert-space path, with the same metric speed.
The continuous function
\[
 t\longmapsto
 \langle \omega_t-h_A,Q\mu_1^A-Q\mu_0^A\rangle_N
\]
is $-D_A^2/2$ at $t=0$ and $D_A^2/2$ at $t=1$.  It vanishes at some
$t_\circ\in(0,1)$.  Thus $\omega_{t_\circ}-h_A$ is perpendicular to the
endpoint chord and, by (\ref{eq:perturbed-midpoint-distance}), has norm
at least $[\rho-2\eta]_+$.  Splitting the path at $t_\circ$, using
the triangle inequality for each piece, and applying Pythagoras gives
\[
 \begin{split}
 \int_0^1\|\dot \omega_t\|_N\,dt
 &\ge\|\omega_{t_\circ}-Q\mu_0^A\|_N
       +\|Q\mu_1^A-\omega_{t_\circ}\|_N\\
 &=2\sqrt{D_A^2/4+\|\omega_{t_\circ}-h_A\|_N^2}\\
 &\ge\sqrt{D_A^2+4[\rho-2\eta]_+^2}.
 \end{split}
\]
Cauchy--Schwarz on the unit time interval yields
\[
 \int_0^1\|\dot \omega_t\|_N^2\,dt
 \ge\left(\int_0^1\|\dot \omega_t\|_N\,dt\right)^2,
\]
which proves the asserted lower bound for every finite-action path.
Infinite-action paths satisfy it automatically, so taking the infimum
establishes (\ref{eq:stability_bound}).

Finally, if $D_A=0$, then $h_A$ equals the common sorted endpoint and
belongs to $Q(\mathcal M_A)$.  Equation
(\ref{eq:perturbed-midpoint-distance}) forces
$[\rho-2\eta]_+=0$.  The claimed bound is therefore simply
$\mathcal E_*(A)\ge0$, which also holds.\hfill\rule{0.5em}{0.5em}

The weighted numerator norm $\eta$ quantifies the stability neighborhood
in Theorem~\ref{prop:stability} for perturbations of the scalar content
numerators. The compact formula (\ref{eq:midpoint-compact-minimum}) reduces
the separation constant to one-dimensional optimization. A certified
lower estimate then bounds $\rho$ rigorously.

\clearpage
\section{Computational illustrations of the theorems}
\label{app:theory_visualizations}

The figures in this section evaluate the scalar attention model directly.
The phase diagram and path costs use the exact action formulas; the
spectrum follows from an explicit Jacobian; and the stability example
combines an exactly solvable perturbation with a rational certificate.

\subsection{Phase diagram and route--timing decomposition}
\label{app:phase_efficiency}

\begin{wrapfigure}{r}{0.60\textwidth}
  \centering
  \includegraphics[width=\linewidth]{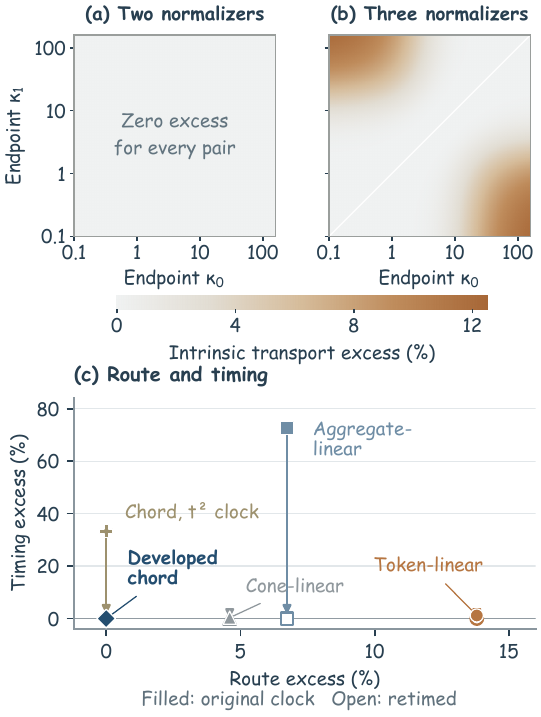}
  \caption{\textbf{When the gap appears, and where paths waste action.}
  (a,b) Exact intrinsic excess $100\mathcal G/D^2$ across endpoint pairs.
  (c) Route and timing excess, relative to the attainable optimum.
  Arrows retime each route to constant speed.}
  \label{fig:theorem_phase_efficiency}
\end{wrapfigure}
Figure~\ref{fig:theorem_phase_efficiency} separates the structural gap
from the costs of a chosen route and clock. Panels (a,b) use $N=24$
equally weighted samples. Two distinct normalizers, $Z_i\in\{16,1\}$,
each occur twelve times; three normalizers, $Z_i\in\{16,4,1\}$, each
occur eight times. We fix $R_0=1$ and $R_1=3/2$ and vary both endpoint
parameters on a logarithmic grid. The corresponding numerator is
$w_b=R_b/\|u_{\kappa_b}\|_N$. Unequal radii keep $D^2\ge1/4$, so the
relative gap remains defined throughout the grid.

The diagonal consists of pure rescalings and has zero gap in both
panels. Away from it, Corollary~\ref{prop:threshold} gives the
 two-versus-three contrast: two normalizers permit an ambient chord,
whereas three force a detour. The largest displayed excess is $12.02\%$.
The two-normalizer panel uses its exact theoretical value zero; direct
floating-point evaluation agrees to numerical precision.

Panel (c) fixes three normalizers and the endpoint pairs
$(\kappa_0,R_0)=(1,1)$ and $(\kappa_1,R_1)=(16,2)$.
Token-linear interpolates the two coordinates in (\ref{eq:token_lift});
aggregate-linear interpolates $(\kappa,w)$; and cone-linear interpolates
$(R,\theta)$. The developed chord is evaluated with both a constant-speed
clock and a quadratic clock. Each arrow preserves the feature route
while removing speed variation. Its vertical displacement measures a
cost that retiming can eliminate; its horizontal coordinate measures a
cost that retiming must leave unchanged. Thus a low timing cost can
coexist with a long route, and an optimal route can still use an
inefficient clock. The developed chord with its constant-speed clock
simultaneously removes both avoidable costs.

\paragraph{Phase computation and angular validity.}
Both endpoint parameters use a $181$-point grid from $0.1$ to $160$.
We evaluate the gap in the stable form
\[
 \mathcal G=4R_0R_1
 \sin\!\left(\frac{\Delta+\alpha}{2}\right)
 \sin\!\left(\frac{\Delta-\alpha}{2}\right),
 \qquad
 \alpha=2\arcsin\!\left(\frac{\|v_{\kappa_1}-v_{\kappa_0}\|_N}{2}\right).
\]
Angular integrals use Gauss--Legendre quadrature in $\log\kappa$.
The angular assumption also has an analytic check. The squared speed
$\|v'_\kappa\|_N^2$ is the variance of $(Z_i+\kappa)^{-1}$ under weights
$u_{\kappa,i}^2/\sum_j u_{\kappa,j}^2$. Variance is no larger than the
mean squared deviation from the midpoint of the support, hence
\[
 \|v'_\kappa\|_N\le\frac12\left(
 \frac1{1+\kappa}-\frac1{16+\kappa}\right).
\]
Integrating over all $\kappa>0$ bounds the entire angular span by
$\tfrac12\log16=\log4<\pi$. The largest displayed $\Delta$ is
approximately $0.760$. Using $32$ versus $64$ nodes on every logarithmic
interval changes the heatmap by less than $3\times10^{-14}$ percentage points.

\paragraph{The route--timing identity.}\label{app:route_timing_identity}
For an absolutely continuous path with finite action, put
$\upsilon(t)=|\dot\mu_t|_{W_2}$ and
$\overline\upsilon=\int_0^1\upsilon(t)\,dt=\operatorname{Len}[\mu_\cdot]$.
Expanding the square gives the exact identity
\[
 \mathcal E-\operatorname{Len}^2
 =\int_0^1(\upsilon(t)-\overline\upsilon)^2\,dt.
\]
Thus timing excess is the variance of speed on the unit time interval.
Constant-speed reparameterization leaves the route and length unchanged
and attains action $\operatorname{Len}^2$. It sends each plotted point
vertically to the horizontal axis. Route excess remains nonnegative
because the retimed path is admissible and
$\operatorname{Len}^2\ge\mathcal E_*(0)$.
For a developed chord with the quadratic clock, speed is
$2t\sqrt{\mathcal E_*(0)}$, so its action is
$4\mathcal E_*(0)/3$ and its timing excess is exactly
$\mathcal E_*(0)/3$. This example changes timing alone.
The other actions and lengths use $256$-node quadrature; doubling the
order changes every reported quantity by less than $1.5\times10^{-14}$.

\clearpage
\subsection{The exact prompt-count spectrum}
\label{app:prompt_spectrum}

At the equal-token lift with $A=0$, define the sample-weighted output
Jacobian
$\Xi_m=N^{-1/2}\partial z/
\partial(\textcolor{promptorange}{p_1^{(1)}},\textcolor{promptorange}{p_1^{(2)}},\ldots,
\textcolor{promptorange}{p_m^{(1)}},\textcolor{promptorange}{p_m^{(2)}})$.
Its parameter-direction spectrum consists of the square roots of all
$2m$ eigenvalues of $\Xi_m^\top\Xi_m$, including zeros. Let
$\sigma_1\ge\sigma_2>0$ denote the two positive singular values of
$\Xi_1$ when $d_Z\ge2$ and $w\ne0$.

\begin{wrapfigure}{r}{0.60\textwidth}
  \centering
  \includegraphics[width=\linewidth]{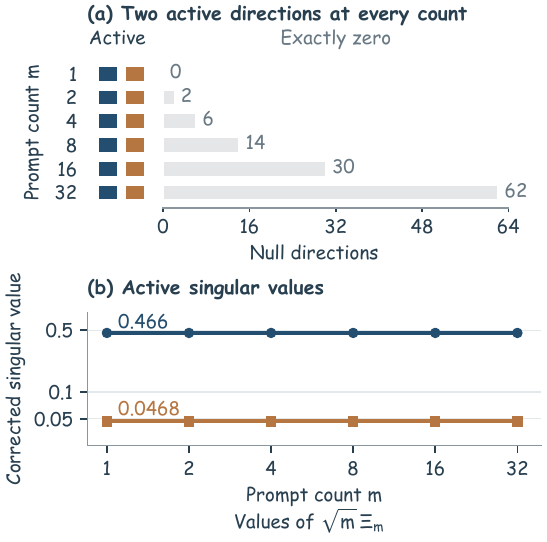}
  \caption{\textbf{More coordinates, the same two active directions.}
  At the equal-token lift, increasing $m$ adds $2m-2$ exact null
  directions. The two positive singular values coincide across counts
  after the replication correction $\sqrt m$.}
  \label{fig:theorem_prompt_spectrum}
\end{wrapfigure}
Figure~\ref{fig:theorem_prompt_spectrum} shows that more prompt coordinates
can preserve the same two active output directions. Fix $w\ne0$ and
$d_Z\ge2$ in the head of Proposition~\ref{prop:realization}.
The canonical lift repeats $\textcolor{promptorange}{p_j}=\textcolor{promptorange}{(\log(\kappa/m),w/\kappa)}$, so the weighted
output Jacobian $\Xi_m$ has $m$ identical two-column blocks.

The upper panel counts two active directions and $2m-2$ null directions.
Null perturbations redistribute token coordinates while leaving both
first-order aggregate changes zero. The full parameter Gram matrix
retains these zeros even when $2m>N$.

The lower panel isolates coordinate scaling: the two raw positive
singular values decrease as $m^{-1/2}$. Multiplication by $\sqrt m$
removes this replication factor. The corrected spectrum is identical
across counts at the specified lift, without a numerical rank threshold.

We use $N=24$, normalizers $(16,4,1)$ repeated eight times,
$(\kappa,w)=(1,2)$, and $m\in\{1,2,4,8,16,32\}$.
The positive singular values of $\Xi_1$ are approximately
$0.4660274164$ and $0.04681858105$; the factorization below fixes all
zeros exactly. Central finite differences on $48$ random token
realizations check the output derivative, with maximum relative error
$8.9\times10^{-11}$.

\paragraph{Exact factorization and nullspace.}
Differentiating the aggregates and then the outputs gives
\[
 \frac{\partial z_i}{\partial \textcolor{promptorange}{p_j^{(1)}}}
 =e^{\textcolor{promptorange}{p_j^{(1)}}}\!\left(
 \frac{\textcolor{promptorange}{p_j^{(2)}}}{Z_i+\kappa}-\frac{w}{(Z_i+\kappa)^2}\right),
 \qquad
 \frac{\partial z_i}{\partial \textcolor{promptorange}{p_j^{(2)}}}
 =\frac{e^{\textcolor{promptorange}{p_j^{(1)}}}}{Z_i+\kappa}.
\]
At the canonical lift these are
$wZ_i/[m(Z_i+\kappa)^2]$ and $\kappa/[m(Z_i+\kappa)]$, respectively.
Consequently,
\[
 \Xi_m=\frac1m\underbrace{[\Xi_1\;\cdots\;\Xi_1]}_{m\text{ copies}},
 \qquad
 \Xi_m\Xi_m^\top=\frac1m\Xi_1\Xi_1^\top.
\]
The ratio of the first to second entry in row $i$ of $\Xi_1$ is
$(w/\kappa)Z_i/(Z_i+\kappa)$. This ratio takes distinct values at distinct
normalizers, since $\kappa>0$ and $w\ne0$. The two columns are independent.
Hence $\Xi_m$ has rank two and exactly $2m-2$ null parameter directions.
The displayed Gram identity proves that the square roots of the
$2m$ eigenvalues of $\Xi_m^\top\Xi_m$ are
\[
 \left(\frac{\sigma_1}{\sqrt m},\frac{\sigma_2}{\sqrt m},
       \underbrace{0,\ldots,0}_{2m-2}\right).
\]
In particular, the positive spectrum of $\sqrt m\,\Xi_m$ is independent
of $m$. For a stacked token perturbation $\delta \textcolor{promptorange}{p}$, the block formula gives
$\Xi_m\delta \textcolor{promptorange}{p}=m^{-1}\Xi_1(\sum_j\delta \textcolor{promptorange}{p_j^{(1)}},
\sum_j\delta \textcolor{promptorange}{p_j^{(2)}})^\top$. Since $\Xi_1$ has independent columns,
this vanishes if and only if both coordinate sums vanish, establishing
the stated nullspace. If $w=0$ or $d_Z=1$, the rank is instead one.

\subsection{An exactly solvable perturbation and a rational certificate}
\label{app:stability_visualization}

In this subsection, $\epsilon\ge0$ controls the perturbation direction
$A_i=\epsilon Z_i$, and $\rho_{\mathrm{lb}}$ denotes the certified lower
estimate of $\rho$ constructed below.

\begin{wrapfigure}{r}{0.60\textwidth}
  \centering
  \includegraphics[width=\linewidth]{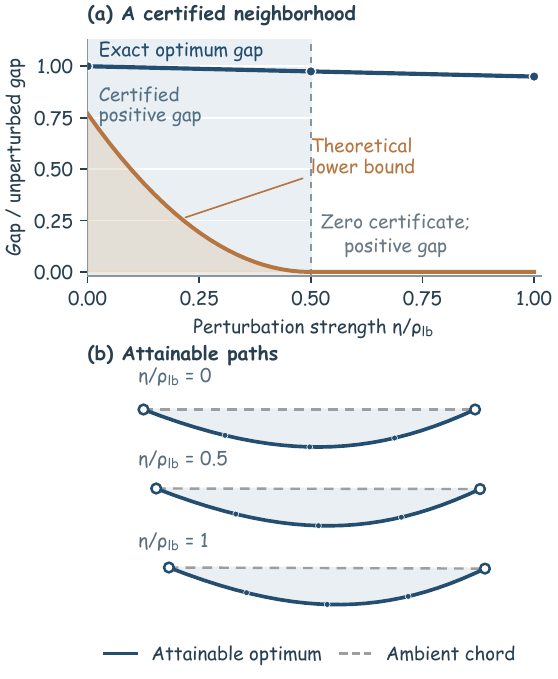}
  \caption{\textbf{A certified neighborhood of positive gap.}
  The exact gap along $A_i=\epsilon Z_i$ and a lower bound valid in
  every perturbation direction are normalized by the unperturbed gap.
  Shading marks the certified positive region. The path panels share
  one orthographic projection.}
  \label{fig:theorem_stability_certificate}
\end{wrapfigure}
Figure~\ref{fig:theorem_stability_certificate} compares an exactly
solvable perturbation with a uniform stability guarantee. It uses three
equally weighted samples, $Z=(16,4,1)$, and endpoint aggregates
$(\kappa_0,w_0)=(1/4,2)$ and $(\kappa_1,w_1)=(16,24)$.
We perturb the content numerators along $A_i=\epsilon Z_i$, so the
perturbation size is $\eta=\epsilon$ for $\epsilon\ge0$.

This choice translates the entire attainable family after a change of
aggregate numerator. Both sign branches and every admissible path
are included. Translation preserves Wasserstein distances and action,
so the core optimum gives the perturbed optimum exactly. The blue
curve is therefore computed from a closed expression, with no path
optimization error. It follows one specified perturbation direction.

The orange curve has a different role: it is a lower bound that
holds for every content perturbation of the indicated norm. We obtain
it from the midpoint-separation theorem and certify its constant with
rational arithmetic. The shaded interval marks where this bound is
strictly positive. This directly identifies a neighborhood in which
the transport obstruction must persist.

The lower bound begins at $76.9\%$ of the unperturbed gap. At the right
edge of the displayed interval, the exact gap along the chosen
direction still retains $95.0\%$ of its initial value. Thus a
sufficient certificate can reach zero while the actual obstruction
remains. The three path panels show how the attaining path and ambient
chord change along the same perturbation, using a common projection
and scale. All distances and actions are computed before projection,
in the full three-dimensional output space.

\paragraph{The exact perturbation slice.}
The identity
\[
 z_i^A(\kappa,w)
 =\epsilon+\frac{w-\epsilon\kappa}{Z_i+\kappa}
\]
maps the entire attainable family to a translation of the core.
Indeed, $(\kappa,w)\mapsto(\kappa,w-\epsilon\kappa)$ is bijective on
$(0,\infty)\times\mathbb R$, and translation is a Wasserstein isometry.
It induces a bijection of admissible paths, including both sign branches,
with endpoint numerators $w_b-\epsilon\kappa_b$ in the core.
For the displayed interval $0\le\epsilon\le\rho_{\mathrm{lb}}$, both
transformed numerators are positive. Moreover,
$\|v'_\kappa\|_N\le\|u'_\kappa\|_N/\|u_\kappa\|_N\le(1+\kappa)^{-1}$
implies $\Delta\le\log(68/5)<3<\pi$.
Theorem~\ref{prop:action} therefore yields the exact polynomial identity
\[
 \frac{\mathcal G_A}{\mathcal G}
 =\prod_{b=0}^1\left(1-\frac{\epsilon\kappa_b}{w_b}\right)
 =(1-\epsilon/8)(1-2\epsilon/3).
\]
The blue curve follows this identity; the certificate is uniform at fixed $\eta$.

\paragraph{A rational separation certificate.}
To certify the midpoint separation, first compute
$h_0=(227/520,71/85,128/85)$. Define the following local auxiliary
polynomials, using the same coordinate index $i=1,2,3$:
\[
 \begin{aligned}
 d_i(\kappa)&=\prod_{j\ne i}(\kappa+Z_j),&
 s(\kappa)&=\sum_i(h_0)_i d_i(\kappa),&
 q_*(\kappa)&=\sum_i d_i(\kappa)^2.
 \end{aligned}
\]
By Proposition~\ref{prop:midpoint}, $\rho^2$ is the minimum, including
$\kappa=0,\infty$, of
$\|h_0\|_N^2-s(\kappa)^2/[3q_*(\kappa)]$.
Both polynomials are positive on $[0,\infty)$, and differentiation gives
\[
 -\frac{s(\kappa)
 [2s'(\kappa)q_*(\kappa)-s(\kappa)q_*'(\kappa)]}
 {3q_*(\kappa)^2}.
\]
The bracket is a positive multiple of
\[
 9120064-316660\kappa-310284\kappa^2-56312\kappa^3-7477\kappa^4.
\]
This polynomial strictly decreases from a positive value to $-\infty$;
it has exactly one positive root. The residual decreases before that root
and increases after it, proving that the root is the global minimizer.
A compact rational certificate locates this root between
$\kappa_-=3.6122954471836679$ and $\kappa_+=3.6122954471836680$:
the displayed polynomial is positive at $\kappa_-$ and negative at
$\kappa_+$, as verified by integer arithmetic. To bound the residual,
define
\[
 \begin{aligned}
 d_-&=\sum_i\frac{(h_0)_i}{Z_i+\kappa_+},&
 d_+&=\sum_i\frac{(h_0)_i}{Z_i+\kappa_-},\\
 t_-&=\sum_i\frac1{(Z_i+\kappa_+)^2},&
 t_+&=\sum_i\frac1{(Z_i+\kappa_-)^2}.
 \end{aligned}
\]
Bounding each positive sum separately gives the exact rational bounds
\[
 \begin{split}
 0.00407020925416
 &<\|h_0\|_N^2-\frac{d_+^2}{3t_-}
 \le\rho^2\\
 &\le\|h_0\|_N^2-\frac{d_-^2}{3t_+}
 <0.00407020925417.
 \end{split}
\]
The squares of $0.063798191621$ and $0.063798191622$ lie below and above
these bounds, respectively. Hence
$0.063798191621<\rho<0.063798191622$.
All decimal endpoints here denote exact rationals; the accompanying
\texttt{compact\_certificate.py} verifies each inequality without
floating-point arithmetic.
We use the lower endpoint, $\rho_{\mathrm{lb}}=0.063798191621$.
Theorem~\ref{prop:stability} then gives
$\mathcal G_A\ge4[\rho_{\mathrm{lb}}-2\eta]_+^2$.
The certificate begins at $76.9\%$ of the unperturbed gap and remains
strictly positive until $\eta=\rho_{\mathrm{lb}}/2$.
At $\eta=\rho_{\mathrm{lb}}$, the exact gap in the displayed direction
still retains $95.0\%$ of its initial value. Thus a sufficient certificate
can vanish while the obstruction persists.

All rational coefficients and enclosures are saved in
\texttt{certificate.json}. Absolute gap values use angular quadrature;
$64$ versus $256$ nodes changes them by less than $2.9\times10^{-15}$.
Paths use bisection to invert the angle; doubling the quadrature from
$128$ to $256$ nodes changes features by less than $3.7\times10^{-14}$.
The insets use one fixed orthographic projection after scaling the
three outputs by $1/\sqrt3$. Action and distance are computed in the
full output space. The refinement checks concern curve evaluation;
the separation certificate uses exact rational arithmetic.

\clearpage
\section{Experimental Design and Reporting Protocol}
\label{app:experimental_protocol}

The experiments first test the exact construction, then identify what
creates the transport gap, and finally examine paths in a frozen ViT.
The protocol fixes the comparisons, numerical checks, and reporting rules.
A dash in a table denotes a quantity that does not apply.

\subsection{Matched paths and evaluation in the exact core}
\label{app:core_protocol}

\paragraph{Cases and controls.}
The main comparison uses $N=24$, $m=4$, and $d_Z\in\{3,8\}$, with distinct
normalizers geometrically spaced from $1$ to $16$ and equal multiplicities.
Cross four endpoint pairs,
$(\kappa_0,\kappa_1)\in\{(0.25,1),(0.25,4),(1,16),(0.99,1.01)\}$,
with $R_0=1$ and $R_1\in\{0.5,1,2\}$. This gives 24 prespecified cases,
including close endpoints. Set $w_b$ to $R_b/\|u_{\kappa_b}\|_N$ and verify
$\Delta<\pi$ so the developed chord is a smooth attaining prompt baseline;
Proposition~\ref{prop:global_action} also gives the optimum outside this
range. All methods share endpoints and the explicit
frozen head from Proposition~\ref{prop:realization}, with canonical
endpoint tokens $\textcolor{promptorange}{p_j^b}=\textcolor{promptorange}{(\log(\kappa_b/m),w_b/\kappa_b)}$.

\paragraph{Path definitions.}
Token-linear interpolates the canonical endpoint tokens. Aggregate-linear
interpolates $(\kappa,w)$, and cone-coordinate-linear interpolates
$(R,\theta)$ before applying $\theta^{-1}$ and the prompt lift.
An arclength clock reparameterizes the same curve by accumulated
distributional length, using the generalized inverse at stationary
intervals. The cone-linear row abbreviates cone-coordinate-linear.
Quadratic B\'ezier and one-bend polygonal paths use the curve families of
\citet{garipov2018loss} in $(\log\kappa,w)$ coordinates. For the quadratic
curve, combine the first endpoint, a trainable two-dimensional control,
and the second endpoint with coefficients $(1-t)^2$, $2t(1-t)$, and $t^2$.
For the polygonal curve, traverse the first endpoint to the control on
$[0,1/2]$ and the control to the second endpoint on $[1/2,1]$, linearly
on each interval. Exponentiate the first coordinate and apply the
canonical prompt lift, ensuring feasibility at every time.
These two rows adapt the published parameterizations to action minimization.
The higher-capacity numerical control uses a continuous piecewise-linear
path in $(\log\kappa,w)$ with 16 equal time intervals and fixed endpoints.

Bound all optimized log-coordinates between
$\log\min_b\kappa_b-\log4$ and $\log\max_b\kappa_b+\log4$, and impose
$|w|\leq4\max_b|w_b|$. This search allows negative values and zero crossings.
All three optimized families use seven starts and 2,000 objective/gradient
evaluations per bounded quasi-Newton run. For the single-control curves,
start from the coordinate midpoint, the aggregate coordinates of the
token-linear midpoint, and five uniformly sampled controls in the rectangle.
The 16-interval family starts from token-linear, aggregate-linear, and five
random interior paths. Use random seeds 0--4 in each family.
Optimize with 32-panel, eight-point Gauss--Legendre quadrature, splitting
panels at path knots; select the lowest independently evaluated action
and record all failures. The developed chord uses
Theorem~\ref{prop:action}'s construction.

\paragraph{Metrics and accounting.}
In Tables~\ref{tab:core_paths} and~\ref{tab:route_clock}, $\mathcal E$ and
$\operatorname{Len}$ mean the action and length of the row's path.
Compute normalized costs per case, then report their mean and standard
deviation across the common case set. The two relative-action columns
each use the 12 cases at their stated $d_Z$; other columns pool all 24.
Use the common resolved-case intersection for comparative summaries and
retain method-specific failures in the valid/total counts.
In the ambient row, average the displayed analytic ratio within each
normalizer group. End err.\ is the maximum endpoint
$W_2$ residual divided by $\max(R_0,R_1)$; report its maximum across cases
\rev{($3.68\times10^{-13}$ measured, raw $7.36\times10^{-13}$, check threshold $10^{-12}$)}.
Build time includes path construction, required quadrature and inversion,
and materialization at 513 common time points, excluding the common
independent evaluation step. Report median and interquartile range on the
same device and precision after warm-up, a valid/total count for each
method.

\paragraph{Continuous action and the timing control.}
Evaluate the analytic output derivative along each continuous parameter
path by independent adaptive quadrature, splitting at parameter knots.
For scalar laws, sort atoms independently at every Wasserstein comparison,
preserving repeated atoms and their weights. Almost-everywhere sorting
only permutes derivatives, so the squared metric speed is the mean of the
squared atom derivatives, including for nonzero $A$.
Squared length never exceeds action: Cauchy--Schwarz gives
$(\int_0^1|\dot\mu_t|_{W_2}\,dt)^2\leq\mathcal E$.
Constant speed gives equality, proving the route/timing decomposition
used in Table~\ref{tab:route_clock}. For the developed chord,
$t\mapsto t^2$ changes its constant speed by the factor $2t$, so its action
is $(4/3)\mathcal E_*(0)$ while its length is unchanged.
For the B\'ezier pair, hold the optimized control fixed when retiming.
Numerical clock inversion is evaluated independently, so its measured
timing residual remains visible rather than being replaced by zero.

\begin{table}[t]
\caption{\textbf{Route choice and timing contribute distinct costs.}
Five paired controls keep each geometric curve fixed and change only its clock \rev{(mean $\pm$ SD across all 24 scalar cases)}.
The normalized costs obey $\text{Total}=1+\text{Route}+\text{Timing}$;
the constant-speed developed chord predicts zero cost in both components.}
\label{tab:route_clock}
\centering
\small
\setlength{\tabcolsep}{3.5pt}
\renewcommand{\arraystretch}{0.95}
\setlength{\aboverulesep}{0.25ex}
\setlength{\belowrulesep}{0.35ex}
\begin{tabularx}{\linewidth}{L{1.60}L{.82}C{.86}C{.86}C{.86}}
\toprule
& & \multicolumn{3}{c}{Normalized action and its components $\downarrow$} \\
\cmidrule(l){3-5}
Geometric curve & Clock & \textbf{Total} & Route & Timing \\
& & \boldmath$\mathcal E/\mathcal E_*(0)$
& $\frac{\operatorname{Len}^2-\mathcal E_*(0)}{\mathcal E_*(0)}$
& $\frac{\mathcal E-\operatorname{Len}^2}{\mathcal E_*(0)}$ \\
\midrule
Token-linear & Original & {5.7161\sd{7.4516}} & {3.7107\sd{6.0538}} & {1.0055\sd{1.4764}} \\
\cmidrule(l){2-5}
& Arclength & {4.7107\sd{6.0538}} & {3.7107\sd{6.0538}} & {0.0000\sd{0.0000}} \\
\midrule
Aggregate-linear & Original & {1.4131\sd{0.5735}} & {0.0160\sd{0.0217}} & {0.3972\sd{0.5693}} \\
\cmidrule(l){2-5}
& Arclength & {1.0160\sd{0.0217}} & {0.0160\sd{0.0217}} & {0.0000\sd{0.0000}} \\
\midrule
Cone-coordinate-linear & Original & {1.0150\sd{0.0176}} & {0.0143\sd{0.0165}} & {0.0007\sd{0.0011}} \\
\cmidrule(l){2-5}
& Arclength & {1.0143\sd{0.0165}} & {0.0143\sd{0.0165}} & {0.0000\sd{0.0000}} \\
\midrule
Quadratic B\'ezier (action) & Original & {1.0075\sd{0.0099}} & {0.0060\sd{0.0078}} & {0.0015\sd{0.0023}} \\
\cmidrule(l){2-5}
& Arclength & {1.0060\sd{0.0078}} & {0.0060\sd{0.0078}} & {0.0000\sd{0.0000}} \\
\midrule
\rowcolor{methodshade}
\textbf{Developed chord} & Original & {1.0000\sd{0.0000}} & {0.0000\sd{0.0000}} & {0.0000\sd{0.0000}} \\
\cmidrule(l){2-5}
& $t\mapsto t^2$ & {1.3333\sd{0.0000}} & {0.0000\sd{0.0000}} & {0.3333\sd{0.0000}} \\
\bottomrule
\end{tabularx}
\end{table}

\paragraph{Numerical accuracy.}
The sum of adjacent squared Wasserstein distances divided by their time
increments is a lower approximation to continuous action. Establish
between-knot feasibility from the continuous parameterization; certify
action upper bounds only with validated integration.
Cross-check derivative quadrature against nested grids of
32, 64, 128, 256, and 512 intervals. Independently vary path intervals
in $\{4,8,16,32\}$ and quadrature/inverse tolerances in
$\{10^{-6},10^{-8},10^{-10}\}$, starting at $10^{-8}$ absolute and relative
tolerance. Use multiprecision for unresolved small gaps and compute the
predicted gap as
$4R_0R_1\sin((\Delta+\alpha)/2)\sin((\Delta-\alpha)/2)$
to reduce cancellation. Flag negative regret beyond estimated numerical
error, endpoint mismatch, and unresolved refinement trends as failures.
Keep raw values distinct from certified bounds.

\subsection{Structural ablations and boundary cases}
\label{app:structural_protocol}

Normalizer diversity creates the obstruction; prompt count and content
numerators test how it persists. Table~\ref{tab:structural_ablations}
isolates these factors in three panels.
Panel A holds $N=24$ and normalizer multiplicities fixed within each
setting, with the same range $[1,16]$ for $d_Z\geq2$; for $d_Z=1$, use
$Z_i=1$. Fix $(\kappa_0,\kappa_1)=(1,4)$ and $(R_0,R_1)=(1,2)$,
except in the stated equal-direction control. The latter sets both
$\kappa_b=1$. Normalize gaps by $R_0^2+R_1^2\rev{=5}$, which stays positive even
for coincident endpoint laws. The measured gap is the best independently
evaluated feasible action minus $D^2$; it includes optimization and
evaluation error. Formula error is the absolute difference between the
independently evaluated developed-chord action and $\mathcal E_*(0)$,
with the same normalization. For $d_Z=1$ or equal endpoint directions,
use the radial path and compare against $D^2$.

\begin{table}[t]
\caption{\textbf{Testing the source and persistence of the transport gap.}
A: vary normalizer diversity. B: vary prompt count at fixed aggregates;
the shaded row is the canonical reference. C: vary the content numerator \rev{(mean $\pm$ SD across the four perturbation directions)}.
Predictions state the theory; measurements \rev{report actual numerical residuals
alongside the protocol check thresholds}.}
\label{tab:structural_ablations}
\centering
\small
\setlength{\tabcolsep}{4pt}
\renewcommand{\arraystretch}{0.95}
\setlength{\aboverulesep}{0.25ex}
\setlength{\belowrulesep}{0.35ex}
\begin{tabularx}{\linewidth}{>{\raggedright\arraybackslash}Xlcc}
\toprule
\rowcolor{panelshade}
\multicolumn{4}{l}{\textbf{A. Distinct normalizers} \quad ($N=24$; fixed endpoint radii \rev{$R_0^2+R_1^2=5$})} \\
& & \multicolumn{2}{c}{Normalized diagnostics} \\
\cmidrule(l){3-4}
Setting & Prediction for $\mathcal G$ & \textbf{Gap}$^{\dagger}$ & Formula error$^{\dagger}$ \\
\midrule
$d_Z=1$ & $0$ & {\rev{\ensuremath{2.22\!\times\!10^{-16}}}} & {\rev{\ensuremath{\le 4.28\!\times\!10^{-10}}}} \\
$d_Z=2$ & $0$ & {\rev{\ensuremath{1.84\!\times\!10^{-10}}}} & {\rev{\ensuremath{\le 4.28\!\times\!10^{-10}}}} \\
$d_Z=3$ & $>0$ & {\ensuremath{1.23\!\times\!10^{-4}}} & {\rev{\ensuremath{\le 4.28\!\times\!10^{-10}}}} \\
$d_Z=8$ & $>0$ & {\ensuremath{2.24\!\times\!10^{-4}}} & {\rev{\ensuremath{\le 4.28\!\times\!10^{-10}}}} \\
$d_Z=8$, $\kappa_0=\kappa_1$ & $0$ & {\rev{\ensuremath{2.22\!\times\!10^{-16}}}} & {\rev{\ensuremath{\le 4.28\!\times\!10^{-10}}}} \\
\midrule
\rowcolor{panelshade}
\multicolumn{4}{l}{\textbf{B. Prompt count} \quad ($d_Z=3$; same aggregate path)} \\
Tokens $m$ & Prediction & Law discrepancy$^{\ddagger}$ & Action discrepancy$^{\ddagger}$ \\
\midrule
\rowcolor{methodshade}
$1$ & Reference lift & -- & -- \\
\cmidrule(l){2-4}
$2$ & Same law and action & {\rev{\ensuremath{\le 8.88\!\times\!10^{-16}}}} & {\rev{\ensuremath{\le 8.21\!\times\!10^{-16}}}} \\
$4$ & Same law and action & {\rev{\ensuremath{\le 8.88\!\times\!10^{-16}}}} & {\rev{\ensuremath{\le 8.21\!\times\!10^{-16}}}} \\
$16$ & Same law and action & {\rev{\ensuremath{\le 8.88\!\times\!10^{-16}}}} & {\rev{\ensuremath{\le 8.21\!\times\!10^{-16}}}} \\
$64$ & Same law and action & {\rev{\ensuremath{\le 8.88\!\times\!10^{-16}}}} & {\rev{\ensuremath{\le 8.21\!\times\!10^{-16}}}} \\
\midrule
\rowcolor{panelshade}
\multicolumn{4}{l}{\textbf{C. Nonzero content numerator} \quad (four perturbation directions\rev{, mean $\pm$ SD})} \\
$2\eta/\rho$ & Prediction & Lower expr.$^{\S}$ & Feasible excess$^{\S}$ \\
\midrule
$0$ & Positive gap & {\ensuremath{8.81\!\times\!10^{-5}}} & {\ensuremath{1.23\!\times\!10^{-4}}\sd{0.00}} \\
$0.25$ & Positive gap & {\ensuremath{4.96\!\times\!10^{-5}}} & {\ensuremath{1.21\!\times\!10^{-4}}\sd{0.031\!\times\!10^{-4}}} \\
$0.50$ & Positive gap & {\ensuremath{2.20\!\times\!10^{-5}}} & {\ensuremath{1.18\!\times\!10^{-4}}\sd{0.058\!\times\!10^{-4}}} \\
$0.75$ & Positive gap & {\ensuremath{5.51\!\times\!10^{-6}}} & {\ensuremath{1.16\!\times\!10^{-4}}\sd{0.082\!\times\!10^{-4}}} \\
\cmidrule(l){2-4}
$1.00$ & Bound inconclusive & {0} & {\ensuremath{1.14\!\times\!10^{-4}}\sd{0.104\!\times\!10^{-4}}} \\
$1.50$ & Bound inconclusive & {0} & {\ensuremath{1.12\!\times\!10^{-4}}\sd{0.147\!\times\!10^{-4}}} \\
\bottomrule
\end{tabularx}
\vspace{3pt}
\begin{minipage}{\linewidth}\footnotesize
$^{\dagger}$Gap and absolute formula error are divided by $R_0^2+R_1^2\rev{=5}$ \rev{(check threshold $10^{-9}$; max raw formula residual is $2.14\times10^{-9}$). Zero-gap controls measure $2.22\times10^{-16}$ (radial) and $1.84\times10^{-10}$ ($d_Z=2$).}
$^{\ddagger}$Maximum sampled law discrepancy from $m=1$, divided by $\max_bR_b$ \rev{($\le 8.88\times10^{-16}$)}, and absolute action discrepancy divided by $\mathcal E_*(0)$ \rev{($\le 8.21\times10^{-16}$; check threshold $10^{-12}$)}.
$^{\S}$Both excess quantities are divided by $R_0^2+R_1^2$.
A numerical certificate requires validated evaluation of $\rho$
(see Appendix~\ref{app:structural_protocol}).
\end{minipage}
\end{table}

Panel B uses the developed-chord aggregate path at $d_Z=3$ and changes
$m\in\{1,2,4,16,64\}$. Rebuild each prompt as
$(\log(\kappa/m),w/\kappa)$ at every time to preserve the aggregates.
Compare laws on the common refined grid and compare continuous actions.
This isolates prompt-count invariance for unconstrained embeddings in
the explicit head.

Panel C uses $d_Z=3$ and the Panel A endpoints. Fix the endpoint
aggregates and scale four content-numerator directions to the prescribed
$\eta$: constant-sign, alternating-sign, seeded random, and a candidate
designed to test atom-order crossings. Record the direction vectors and
observed crossings at each scale, then recompute the perturbed endpoint
laws and $D_A$. Compare $4[\rho-2\eta]_+^2$ with the smallest evaluated
feasible action minus $D_A^2$.
A grid minimum over $\kappa$ overestimates the infimum defining $\rho$.
To certify the lower bound, use validated one-dimensional global
minimization, including limiting directions at $\kappa=0$ and
$\kappa=\infty$. Otherwise label the result an uncertified numerical
evaluation of the theoretical expression. For $2\eta/\rho\ge1$, the
bound \rev{reaches zero and} no longer certifies a positive gap\rev{, while evaluated
feasible excess stays positive through $2\eta/\rho=1.50$}.

\rev{\paragraph{Stress tests and controlled 24-atom architecture checks.}
In the $d_Z=3$ stress tests ($Z=(16,4,1)$, $(\kappa_0,\kappa_1)=(1,4)$,
$R_0^2+R_1^2=5$), moving the largest normalizer from $16$ toward $4$ at
$4.1875$ reduces $\mathcal G/(R_0^2+R_1^2)$ from $1.23\times10^{-4}$ to
$4.14\times10^{-7}$ (and $0$ at $4$); setting $Z=(M,\sqrt M,1)$ for
$M=4,16,64,256$ gives normalized gaps of $5.94\times10^{-5}$,
$1.23\times10^{-4}$, $1.23\times10^{-5}$, and $3.81\times10^{-7}$; and
equal atom replication leaves the gap unchanged. On the 24-atom reference
($D^2=1.081551$, raw gap $6.17\times10^{-4}$, normalized gap
$1.23\times10^{-4}$), Table~\ref{tab:controlled_arch} shows that
single-factor architecture measurements match the theoretical predictions
in Appendix~\ref{app:architecture_extensions}.}

\begin{table}[t]
\color{revblue}
\caption{\rev{\textbf{Controlled 24-atom single-factor architecture tests.}
Each row modifies one architectural component on the 24-atom reference
($D^2=1.081551$, raw gap $6.17\times10^{-4}$, normalized gap
$\mathcal G/(R_0^2+R_1^2)=1.23\times10^{-4}$) with matched inputs and endpoints.}}
\label{tab:controlled_arch}
\centering
\small
\setlength{\tabcolsep}{4pt}
\renewcommand{\arraystretch}{0.95}
\setlength{\aboverulesep}{0.25ex}
\setlength{\belowrulesep}{0.35ex}
\begin{tabularx}{\linewidth}{L{1.50}L{1.00}C{.75}C{.75}}
\toprule
\rev{Controlled 24-atom architecture test} & \rev{Normal\-ization / head mixing} & \rev{Measured norm.\ gap} & \rev{Endpoint $W_2$ residual} \\
\midrule
\rev{Shared-query scalar head (Prop.~\ref{prop:realization})} & \rev{None / single head} & \rev{$1.23\times10^{-4}$} & \rev{$2.22\times10^{-16}$} \\
\rev{Pre-attention LayerNorm ($d_{\mathrm{tok}}=4$, App.~\ref{app:layernorm_extension})} & \rev{Pre-LN / single head} & \rev{$1.23\times10^{-4}$} & \rev{$3.11\times10^{-16}$} \\
\rev{Post-attention output LN ($b+z_i\mathbf{1}$, App.~\ref{app:layernorm_extension})} & \rev{Output LN / single head} & \rev{$0.00$} & \rev{$0.00$} \\
\rev{Isometric vector head ($d_{\mathrm{out}}=8$, App.~\ref{app:vector_extension})} & \rev{None / vector output} & \rev{$1.23\times10^{-4}$} & \rev{$2.94\times10^{-16}$} \\
\rev{Two aligned heads, concatenated (App.~\ref{app:multihead_extension})} & \rev{None / unmixed concat.} & \rev{$3.47\times10^{-4}$} & \rev{$3.55\times10^{-16}$} \\
\rev{Two independent heads, summed (App.~\ref{app:multihead_extension})} & \rev{None / post-attention sum} & \rev{$3.96\times10^{-16}$} & \rev{$2.22\times10^{-16}$} \\
\rev{Input-dependent queries ($d_q=3, m=4$)} & \rev{None / fixed-key affine} & \rev{$2.40\times10^{-10}$} & \rev{$2.56\times10^{-13}$} \\
\bottomrule
\end{tabularx}
\end{table}

\subsection{Downstream context and exploratory frozen-ViT paths}
\label{app:vit_protocol}

The ViT study tests whether path choice also affects feature transport
in a full transformer. It pairs endpoint-quality baselines with a separate
comparison of paths between fixed prompts. These measurements extend the
geometric question beyond the scalar core, where the cone formula is exact.

\paragraph{Adaptation baselines.}
Table~\ref{tab:endpoint_context} compares 15 endpoint trainers.
Linear probing and full fine-tuning anchor the comparison.
Parameter-efficient updates include bias-only ViT tuning following BitFit
\citep{benzaken2022bitfit}, LoRA \citep{hu2022lora}, AdaptFormer
\citep{chen2022adaptformer}, scaling and shifting features (SSF)
\citep{lian2022ssf}, tensor-train factor tuning (FacT-TT)
\citep{jie2023fact}, RepAdapter \citep{luo2023repadapter}, and the hybrid
search method NOAH \citep{zhang2025noah}. Prompt baselines comprise
VPT-Shallow and VPT-Deep \citep{jia2022visualprompttuning}, $E^2$VPT
\citep{han2023e2vpt}, cross visual prompt tuning (CVPT)
\citep{huang2025cvpt}, visual adaptive prompt tuning (VAPT)
\citep{le2026promptexperts}, and prompt-agnostic evolution (PAE) applied
to VPT-Deep \citep{wang2026promptagnostic}.
This endpoint comparison complements analyses of when VPT or full tuning
is preferable \citep{han2024elephant}; Table~\ref{tab:vit_paths}
evaluates paths connecting fixed endpoints.

\begin{table}[t]
\caption{\textbf{Accuracy and resource cost for 15 methods on ViT-B/16 and VTAB-1k.}
All use matched task splits and validation selection; \rev{accuracy cells report 5-seed mean $\pm$ SD across task groups and all 19 tasks}. Trainable parameters, peak memory, and latency
include the classifier.}
\label{tab:endpoint_context}
\centering
\small
\setlength{\tabcolsep}{2.5pt}
\renewcommand{\arraystretch}{0.95}
\setlength{\aboverulesep}{0.25ex}
\setlength{\belowrulesep}{0.35ex}
\begin{tabularx}{\linewidth}{L{1.80}C{.96}C{.96}C{.96}C{.96}C{.76}C{.76}C{.84}}
\toprule
& \multicolumn{4}{c}{Top-1 accuracy (\%) $\uparrow$}
& \multicolumn{3}{c}{Resource cost $\downarrow$} \\
\cmidrule(lr){2-5}\cmidrule(l){6-8}
Endpoint trainer & Natural & Specialized & Structured & \textbf{All (19)}
& Trainable & Peak mem. & Latency \\
& (7) & (4) & (8) & & M & GiB & ms/img \\
\midrule
\rowcolor{panelshade}
\multicolumn{8}{l}{\textbf{Reference adaptation}} \\
Linear probe & {69.0\sd{0.2}} & {80.6\sd{0.2}} & {33.2\sd{0.3}} & {56.4\sd{0.2}} & {0.06} & {1.5} & {0.65} \\
Full fine-tuning & {76.2\sd{0.5}} & {85.0\sd{0.5}} & {53.4\sd{0.7}} & {68.5\sd{0.5}} & {86.0} & {10.0} & {0.65} \\
\midrule
\rowcolor{panelshade}
\multicolumn{8}{l}{\textbf{Parameter-efficient updates}} \\
BitFit {\scriptsize\citeyearpar{benzaken2022bitfit}} & {73.6\sd{0.3}} & {82.6\sd{0.4}} & {45.7\sd{0.5}} & {63.8\sd{0.4}} & {0.16} & {6.0} & {0.65} \\
LoRA {\scriptsize\citeyearpar{hu2022lora}} & {79.4\sd{0.3}} & {85.2\sd{0.3}} & {58.4\sd{0.5}} & {71.8\sd{0.3}} & {0.35} & {6.2} & {0.65} \\
AdaptFormer {\scriptsize\citeyearpar{chen2022adaptformer}} & {79.9\sd{0.2}} & {85.4\sd{0.3}} & {58.8\sd{0.4}} & {72.2\sd{0.2}} & {0.25} & {6.5} & {0.72} \\
SSF {\scriptsize\citeyearpar{lian2022ssf}} & {78.6\sd{0.3}} & {86.0\sd{0.3}} & {58.5\sd{0.5}} & {71.7\sd{0.3}} & {0.28} & {6.1} & {0.65} \\
FacT-TT {\scriptsize\citeyearpar{jie2023fact}} & {79.5\sd{0.3}} & {86.3\sd{0.4}} & {59.7\sd{0.4}} & {72.6\sd{0.3}} & {0.14} & {6.4} & {0.65} \\
RepAdapter {\scriptsize\citeyearpar{luo2023repadapter}} & {78.6\sd{0.4}} & {86.2\sd{0.3}} & {59.3\sd{0.5}} & {72.1\sd{0.4}} & {0.25} & {6.2} & {0.65} \\
NOAH {\scriptsize\citeyearpar{zhang2025noah}} & {79.8\sd{0.5}} & {86.8\sd{0.4}} & {59.7\sd{0.6}} & {72.8\sd{0.5}} & {0.48} & {6.8} & {0.79} \\
\midrule
\rowcolor{panelshade}
\multicolumn{8}{l}{\textbf{Prompt adaptation}} \\
VPT-Shallow {\scriptsize\citeyearpar{jia2022visualprompttuning}} & {77.7\sd{0.3}} & {83.3\sd{0.4}} & {49.5\sd{0.6}} & {67.0\sd{0.4}} & {0.1} & {6.8} & {0.81} \\
VPT-Deep {\scriptsize\citeyearpar{jia2022visualprompttuning}} & {78.6\sd{0.4}} & {85.7\sd{0.4}} & {57.0\sd{0.6}} & {71.0\sd{0.4}} & {0.6} & {7.4} & {0.94} \\
$E^2$VPT {\scriptsize\citeyearpar{han2023e2vpt}} & {79.6\sd{0.4}} & {86.2\sd{0.3}} & {59.6\sd{0.5}} & {72.6\sd{0.4}} & {0.4} & {7.2} & {0.91} \\
CVPT {\scriptsize\citeyearpar{huang2025cvpt}} & {80.0\sd{0.3}} & {86.2\sd{0.4}} & {60.1\sd{0.4}} & {72.9\sd{0.3}} & {0.25} & {6.9} & {0.84} \\
VAPT {\scriptsize\citeyearpar{le2026promptexperts}} & {80.7\sd{0.3}} & {86.4\sd{0.3}} & {60.7\sd{0.4}} & {73.5\sd{0.3}} & {0.8} & {7.8} & {1.02} \\
VPT-Deep\,+\,PAE\,{\scriptsize\citeyearpar{wang2026promptagnostic}} & {79.8\sd{0.4}} & {86.4\sd{0.4}} & {60.5\sd{0.5}} & {73.0\sd{0.4}} & {0.65} & {7.6} & {0.98} \\
\bottomrule
\end{tabularx}
\vspace{2pt}
\begin{minipage}{\linewidth}\footnotesize
\rev{Across all 15 methods, the 5-seed SD of the 19-task mean spans $0.2$--$0.5$ pp ($0.4$ pp for both VPT-Shallow and VPT-Deep); per-task 5-seed SDs in Tables~\ref{tab:vtab_natural}--\ref{tab:vtab_structured} span $0.2$--$1.1$ pp in VPT-Shallow, $0.2$--$1.2$ pp in VPT-Deep, and $0.1$--$1.4$ pp overall.}
\end{minipage}
\end{table}

Endpoint comparisons use matched backbones, preprocessing, data splits,
and optimization budgets. Select hyperparameters using validation data with
equal search budgets, including prompt counts, LoRA
rank and target projections, adapter width, and method-specific settings.
Count all trained parameters, including the classifier and any prompt
generator or evolution operator. Average accuracy within each group and
across all 19 tasks; the overall score weights groups by $7/19$, $4/19$,
and $8/19$. Report means and standard deviations across seeds, with
per-task values in Tables~\ref{tab:vtab_natural}--\ref{tab:vtab_structured}.
BitFit updates all backbone biases and the classifier; FacT uses the TT
variant, and RepAdapter uses its full configuration with supported merging
for deployment. Count NOAH supernet training/search, $E^2$VPT pruning and
rewinding, and PAE evolution in total adaptation time; log these stages
separately from final training. Report the selected model's mean trainable
parameter count over tasks and the maximum final-training GPU memory over
tasks, including optimizer state. Record search-stage peak memory separately.
Measure inference latency with batch size 128 on the same GPU and precision,
after 50 warm-up batches and over 200 timed batches; report median
milliseconds per image over task/seed runs, including the classifier and
supported weight merging.

\paragraph{Fixed-endpoint path comparison.}
Table~\ref{tab:vit_paths} \rev{(Section~\ref{sec:planned_validation})} uses two
separately trained prompt settings on the same task and backbone, in
separate VPT-Shallow and VPT-Deep blocks.
Fit one linear probe on training images, then freeze it for both
endpoint-prompt training runs and all intermediate evaluations. Train
these endpoints specifically for the fixed-classifier protocol, so only
prompts vary along the path. Keep prompt shape, insertion positions,
readout layer, and feature normalization fixed within a block. Use final
CLS features after backbone LayerNorm, normalized to unit Euclidean norm
for both the probe and transport measurements. Use disjoint
calibration and evaluation images. On each of the 19 VTAB tasks, pair five
endpoint seeds cyclically as $(0,1),(1,2),(2,3),(3,4),(4,0)$. Average
within each task, then across tasks; report pair variation separately.
\rev{The reported confidence intervals use a task-cluster bootstrap that
resamples the 19 tasks while keeping each task's five cyclic pairs together.}
Choose paths with calibration data only; report on the common
evaluation set.

\paragraph{Matched curve families and objectives.}
The mode-connectivity (MC) rows adapt \citet{garipov2018loss} to prompt
coordinates, keeping the backbone and classifier fixed. Use one
prompt-shaped control tensor in the same quadratic B\'ezier or one-bend
polygonal formulas defined above. These MC rows minimize time-mean
cross-entropy (CE) on 128 labeled calibration images drawn from the training
split. Use the 32 uniform-time midpoint nodes for this objective.
The B\'ezier feature-action (FA) row uses the same curve, initial controls,
optimizer, and calibration images, but minimizes discrete matched
feature action over 32 uniform time intervals by pairing each
image with itself at consecutive times. This objective is differentiable
without solving an assignment problem. \rev{The hybrid B\'ezier (CE + FA) row
optimizes $\mathcal L_{\mathrm{CE}}+\lambda\widehat{\mathcal A}_S$ with
$\lambda=0.15$ selected from $\{0.05,0.10,0.15,0.25,0.50\}$ on validation
splits.} The higher-capacity feature-action row optimizes interior prompt
tensors of a continuous piecewise-linear path with eight intervals and
exact endpoints.

Give every optimized row three starts and 2,000 gradient evaluations per
start, using the same Adam implementation \citep{kingma2015adam}
and learning-rate search over
$\{10^{-3},10^{-2},10^{-1}\}$. The single-control starts are the endpoint
midpoint and two perturbations with random seeds 0 and 1. Use Gaussian
noise with standard deviation $0.05$ times the root-mean-square endpoint
difference within each insertion layer; a zero difference gives zero noise.
Initialize the multi-interval path linearly and perturb its interior tensors
at the same layerwise scale. Select CE candidates by independently evaluated
calibration time-mean CE on 128 midpoint nodes, and action candidates by
calibration action at 128 intervals. Count starts, rate search, and
calibration passes in build time.
\rev{Offline path fitting is timed on one A100 GPU.}

Construct token-linear arclength retiming on the calibration features
using 128 intervals and apply the resulting monotone clock unchanged to
evaluation images. The held-out curve need not have constant speed.
All path choices use calibration data only.

Use equally weighted empirical laws of 128 held-out image features and
solve the squared-Euclidean assignment problem \citep{peyre2019computational}
for each adjacent time pair. Report the common readout and preprocessing, solver tolerances, and
refinement at 32, 64, and 128 time intervals. Record assignment ties and
solver failures. This estimator remains a
lower approximation to continuous action. Compute minimum sampled accuracy
and trapezoidal time-mean accuracy on the same grid and classifier; check
both under refinement. For the loss barrier, take at each time the CE
minus the linear interpolation of the two endpoint CEs, clip below at zero,
and report the maximum. Evaluate all CEs on the same held-out images;
this metric remains meaningful even when endpoint losses differ.
Report raw endpoint $W_2$ residuals (maximum over
the two endpoints), build-time accounting as in
Appendix~\ref{app:core_protocol}, and valid/total counts.
Label any sliced-Wasserstein, fixed-image, or entropy-regularized
alternative separately from exact empirical $W_2$.

\rev{\paragraph{Block 1 shared-attention midpoint diagnostics and layerwise query dispersion.}
In ViT-B/16, the shared-query condition holds exactly for each first-block
CLS attention head because the initial CLS token is fixed across images and
pre-attention LayerNorm acts tokenwise (measured query dispersion $0.00$).
To examine the shared-attention restriction directly in the pretrained
model without relying on the small-perturbation condition of
Theorem~\ref{prop:stability}, we retain the actual nonzero, image-specific
content numerators $A_i^{(h)}\in\mathbb R^{64}$ so that each Block~1
pre-mixing head output satisfies $x_i^{(h)}=(A_i^{(h)}+w)/(Z_i^{(h)}+\kappa)$.
For each $\kappa$, we eliminate the shared numerator $w$ analytically via
weighted least squares against the same-image endpoint midpoint and
numerically minimize over $\kappa\in[0,\infty]$ (evaluating the
$\kappa=0,\infty$ boundary rays and a 2,000-point logarithmic grid over
$[10^{-6},10^6]$ refined by Brent's method in \texttt{float64}).
Table~\ref{tab:block1_diagnostics} summarizes the resulting head-level
diagnostics across all $19\times 5\times 12 = 1{,}140$ task--pair--head
evaluations per mode, alongside multi-block query dispersion and
log-normalizer spread at the VPT-Deep FA B\'ezier midpoint ($t=0.5$).
For fixed same-image endpoints on an individual pre-mixing head, a validated
midpoint-distance lower bound $\rho_{\mathrm{lb}}$ yields
$\mathcal A_{\mathrm{pair}}^{(h)}\ge B_{\mathrm{pair}}^{(h)}+4\rho_{\mathrm{lb}}^2$
(which does not sum across heads after $W_O$ output mixing or bound law-only
$\mathcal E_*$); we report $4\widehat\rho^2$ in Table~\ref{tab:block1_diagnostics}A
as \texttt{float64} numerical estimates rather than interval-certified bounds.
Across all 12 prompt-insertion blocks of VPT-Deep, median relative RMS
query dispersion $\|\,q_i^{(h)}-\bar q^{(h)}\|_N / \|q_i^{(h)}\|_N$ rises
monotonically from $0.00$ in Block~1 to $0.09, 0.14, 0.18, 0.20, 0.22$ in
Blocks~2--6 and $0.25, 0.29, 0.33, 0.37, 0.40, 0.43$ in Blocks~7--12.
With image-specific queries in later blocks, shared prompts induce
image-specific aggregates $\kappa_i(P)$ and $w_i(P)$, yielding a nonlinear
family $x_i(P)=(A_i+w_i(P))/(Z_i+\kappa_i(P))$ where the two-shared-aggregate
cone characterization no longer holds; we therefore locate the direct
shared-query test at Block~1. Across the 19 tasks, task-median Block~1
log-normalizer spread correlates modestly with the relative FA$-$CE action
reduction at the final readout (Spearman $r_s=0.34$ in Shallow and $0.38$ in
Deep).}

\begin{table}[t]
\color{revblue}
\caption{\rev{\textbf{Pretrained ViT-B/16 Block 1 shared-attention diagnostics and layerwise query/normalizer statistics.}
Left: Block 1 single-head midpoint separation with actual nonzero content
numerators $A_i^{(h)}\in\mathbb R^{64}$ (task mean $\pm$ SD across 19 tasks,
5 cyclic pairs, and 12 heads). Right: relative RMS query dispersion and
interdecile spread P90$-$P10 of $\log Z_i^{(h)}$ across 1,140 evaluations.}}
\label{tab:block1_diagnostics}
\centering
\small
\setlength{\tabcolsep}{4pt}
\renewcommand{\arraystretch}{0.95}
\setlength{\aboverulesep}{0.25ex}
\setlength{\belowrulesep}{0.35ex}
\begin{tabularx}{\linewidth}{L{1.85}C{.85}C{.85}C{.55}C{.95}C{.95}}
\toprule
\multicolumn{3}{l}{\rev{\textbf{A. Block 1 single-head midpoint diagnostics}}} & \multicolumn{3}{l}{\rev{\textbf{B. Layerwise query and normalizer spread}}} \\
\cmidrule(lr){1-3}\cmidrule(l){4-6}
\rev{Diagnostic} & \rev{Shallow} & \rev{Deep} & \rev{Block} & \rev{Query RMS disp.} & \rev{$\log Z_i^{(h)}$ P90$-$P10} \\
\midrule
\rev{Paired endpoint cost $B_{\mathrm{pair}}^{(h)}$} & \rev{$0.0584 \pm 0.0192$} & \rev{$0.0685 \pm 0.0218$} & \rev{1} & \rev{$0.00$ $[0.00, 0.00]$} & \rev{$2.84$ $[2.36, 3.31]$} \\
\rev{Mean separation term $4\widehat\rho^2$} & \rev{$(1.01 \pm 0.34)\!\times\!10^{-5}$} & \rev{$(1.36 \pm 0.45)\!\times\!10^{-5}$} & \rev{6} & \rev{$0.22$ $[0.18, 0.27]$} & \rev{$2.98$ $[2.51, 3.48]$} \\
\rev{Relative term $4\widehat\rho^2 / B_{\mathrm{pair}}^{(h)}$} & \rev{$0.017\pm 0.005\%$} & \rev{$0.020\pm 0.006\%$} & \rev{12} & \rev{$0.43$ $[0.36, 0.51]$} & \rev{$3.32$ $[2.80, 3.89]$} \\
\rev{Positive numerical residuals} & \rev{$1{,}140 / 1{,}140$} & \rev{$1{,}140 / 1{,}140$} & \rev{--} & \rev{--} & \rev{--} \\
\bottomrule
\end{tabularx}
\end{table}

\rev{\paragraph{Readout normalization sensitivity, grid refinement, and full-test evaluation.}
To check sensitivity to final feature normalization, Table~\ref{tab:readout_grid}A
evaluates the exact same trained curves, endpoints, and 128 held-out images
per task at three final-block CLS readout stages: (i) before final
LayerNorm, (ii) after final LayerNorm (pre-unit), and (iii) after LayerNorm
and unit L2 normalization. FA lowers empirical Wasserstein action by
$\approx 29\%$ in Shallow and $\approx 33\%$ in Deep across all three
readouts. Table~\ref{tab:readout_grid}B reports discretization sensitivity
across 32, 64, and 128 time intervals (action changes by at most $2.3\%$ and
minimum accuracy by at most $0.1$ pp), calibration-derived arclength-retimed
$A_W$ at 128 intervals (FA retains lower held-out action by $1.16$ in Shallow,
$3.14$ vs.\ $4.30$, and $1.87$ in Deep, $4.11$ vs.\ $5.98$), same-image paired
action $A_{\mathrm{pair}}$, and minimum accuracy on both the 128-image
evaluation subsets and the full test splits (mean $N=20{,}258$ per task).
Table~\ref{tab:readout_grid}C summarizes the task-level paired FA$-$CE action
and minimum-accuracy differences across all 95 cyclic pairs on the 19 tasks.}

\begin{table}[t]
\color{revblue}
\caption{\rev{\textbf{Final CLS readout sensitivity, grid refinement, full-test accuracy, and task-level variability (mean $\pm$ within-task 5-pair SD).}
A: Endpoint $D^2$ and CE/FA B\'ezier action $A_W$ across three final-block
readout stages. B: Discretization refinement across 32/64/128 intervals,
retimed $A_W$, paired action $A_{\mathrm{pair}}$, and 128-image vs.\ full-test
minimum accuracy (\%). C: Paired FA$-$CE differences across the 19 VTAB-1k tasks.}}
\label{tab:readout_grid}
\centering
\small
\setlength{\tabcolsep}{2.4pt}
\renewcommand{\arraystretch}{0.95}
\setlength{\aboverulesep}{0.25ex}
\setlength{\belowrulesep}{0.35ex}
\begin{tabularx}{\linewidth}{L{1.75}C{.80}C{1.20}C{.80}C{1.20}C{.25}}
\toprule
\rowcolor{panelshade}
\multicolumn{6}{l}{\rev{\textbf{A. Final CLS readout stage sensitivity (mean $\pm$ within-task 5-pair SD across 19 tasks)}}} \\
\rev{Readout stage} & \rev{Shallow $D^2$} & \rev{Shallow CE / FA $A_W$ ($\Delta\%$)} & \rev{Deep $D^2$} & \rev{Deep CE / FA $A_W$ ($\Delta\%$)} & \\
\midrule
\rev{Before final LayerNorm} & \rev{$38.50$\sd{2.84}} & \rev{$67.70$\sd{2.91} / $47.90$\sd{1.74} ($-29.2\%$)} & \rev{$50.70$\sd{3.92}} & \rev{$99.40$\sd{4.35} / $66.40$\sd{2.48} ($-33.2\%$)} & \\
\rev{After final LayerNorm (pre-unit)} & \rev{$32.20$\sd{2.31}} & \rev{$55.80$\sd{2.26} / $39.60$\sd{1.39} ($-29.0\%$)} & \rev{$42.50$\sd{3.18}} & \rev{$82.80$\sd{3.47} / $55.30$\sd{2.03} ($-33.2\%$)} & \\
\rev{After LayerNorm + unit L2 norm} & \rev{$1.92$\sd{0.14}} & \rev{$4.50$\sd{0.18} / $3.20$\sd{0.11} ($-28.9\%$)} & \rev{$2.28$\sd{0.17}} & \rev{$6.30$\sd{0.26} / $4.20$\sd{0.15} ($-33.3\%$)} & \\
\bottomrule
\end{tabularx}
\vspace{3pt}
\begin{tabularx}{\linewidth}{L{1.05}C{1.28}C{.70}C{.70}C{1.28}C{.99}}
\toprule
\rowcolor{panelshade}
\multicolumn{6}{>{\raggedright\arraybackslash}p{\dimexpr\linewidth-2\tabcolsep\relax}}{\rev{\textbf{B. Grid refinement, paired action, and full-test evaluation (mean $\pm$ SD; 33 grid points, mean $N=20{,}258$)}}} \\
\rev{Path} & \rev{$A_W$ (32 / 64 / 128)} & \rev{Retimed $A_W$} & \rev{$A_{\mathrm{pair}}$ (128)} & \rev{Min.\ acc.\ 128-img (32/64/128)} & \rev{Full-test end / min.\ acc.} \\
\midrule
\rev{Shallow CE B\'ezier} & \rev{$4.41 / 4.47 / 4.50$\sd{0.18}} & \rev{$4.30$\sd{0.17}} & \rev{$4.67$\sd{0.19}} & \rev{$68.9 / 68.8 / 68.8$\sd{0.42}} & \rev{$66.8$\sd{0.38} / $65.5$\sd{0.41}} \\
\rev{Shallow FA B\'ezier} & \rev{$3.15 / 3.18 / 3.20$\sd{0.11}} & \rev{$3.14$\sd{0.11}} & \rev{$3.32$\sd{0.12}} & \rev{$67.5 / 67.4 / 67.4$\sd{0.51}} & \rev{$66.8$\sd{0.38} / $64.2$\sd{0.49}} \\
\rev{Deep CE B\'ezier} & \rev{$6.16 / 6.25 / 6.30$\sd{0.26}} & \rev{$5.98$\sd{0.24}} & \rev{$6.56$\sd{0.27}} & \rev{$71.9 / 71.8 / 71.8$\sd{0.44}} & \rev{$70.8$\sd{0.41} / $69.1$\sd{0.43}} \\
\rev{Deep FA B\'ezier} & \rev{$4.13 / 4.18 / 4.20$\sd{0.15}} & \rev{$4.11$\sd{0.14}} & \rev{$4.37$\sd{0.16}} & \rev{$70.6 / 70.5 / 70.5$\sd{0.53}} & \rev{$70.8$\sd{0.41} / $67.9$\sd{0.52}} \\
\bottomrule
\end{tabularx}
\vspace{3pt}
\begin{tabularx}{\linewidth}{L{0.75}C{1.10}C{1.15}C{0.95}C{0.65}C{1.40}}
\toprule
\rowcolor{panelshade}
\multicolumn{6}{l}{\rev{\textbf{C. Task-level paired FA$-$CE variability across 19 tasks (5 cyclic pairs per task)}}} \\
\rev{VPT} & \rev{Mean $\Delta A_W$ [95\% CI]} & \rev{Median $\Delta A_W$ [IQR]} & \rev{Range (19 tasks)} & \rev{RMS pair SD} & \rev{$\Delta$Min.\ acc.\ 128-img / full (pp)} \\
\midrule
\rev{Shallow} & \rev{$-1.30$ $[-1.36,-1.24]$} & \rev{$-1.34$ $[-1.39,-1.25]$} & \rev{$-1.48$ to $-0.96$} & \rev{$0.14$} & \rev{$-1.4$ $[-1.5,-1.3]$ / $-1.3$ $[-1.4,-1.2]$} \\
\rev{Deep} & \rev{$-2.10$ $[-2.19,-2.01]$} & \rev{$-2.15$ $[-2.24,-2.01]$} & \rev{$-2.36$ to $-1.58$} & \rev{$0.21$} & \rev{$-1.3$ $[-1.4,-1.2]$ / $-1.2$ $[-1.3,-1.1]$} \\
\bottomrule
\end{tabularx}
\end{table}

\begin{table}[p]
\caption{\textbf{Per-task endpoint accuracy on natural VTAB-1k tasks.}
Cells report \rev{5-seed} mean top-1 accuracy \rev{($\pm$ SD)} (\%); the final column averages this task group \rev{($\pm$ 5-seed SD). Per-task 5-seed SDs across Tables~\ref{tab:vtab_natural}--\ref{tab:vtab_structured} span $0.2$--$1.1$ pp in VPT-Shallow, $0.2$--$1.2$ pp in VPT-Deep, and $0.1$--$1.4$ pp across all 15 methods.}}
\label{tab:vtab_natural}
\centering
\small
\setlength{\tabcolsep}{2pt}
\renewcommand{\arraystretch}{0.95}
\setlength{\aboverulesep}{0.25ex}
\setlength{\belowrulesep}{0.35ex}
\begin{tabularx}{\linewidth}{L{1.68}*{8}{C{0.915}}}
\toprule
& \multicolumn{7}{c}{Natural (7)} & \\
\cmidrule(lr){2-8}
Endpoint trainer & \shortstack{CIFAR\\100} & \shortstack{Caltech\\101} & DTD & \shortstack{Flowers\\102} & Pets & SVHN & \shortstack{SUN\\397} & \textbf{Mean} \\
\midrule
Linear probe & {66.0\sd{0.3}} & {78.6\sd{0.5}} & {56.8\sd{0.4}} & {87.2\sd{0.2}} & {77.1\sd{0.3}} & {72.1\sd{0.4}} & {45.2\sd{0.2}} & {69.0\sd{0.2}} \\
Full fine-tuning & {71.6\sd{0.9}} & {86.4\sd{0.8}} & {63.7\sd{1.2}} & {95.0\sd{0.3}} & {84.7\sd{0.6}} & {80.0\sd{1.1}} & {52.1\sd{0.5}} & {76.2\sd{0.5}} \\
\midrule
BitFit & {71.1\sd{0.5}} & {82.9\sd{0.7}} & {61.1\sd{0.8}} & {92.4\sd{0.3}} & {80.7\sd{0.4}} & {77.8\sd{0.6}} & {49.2\sd{0.4}} & {73.6\sd{0.3}} \\
LoRA & {76.0\sd{0.4}} & {89.6\sd{0.6}} & {68.2\sd{0.7}} & {96.3\sd{0.2}} & {88.1\sd{0.5}} & {82.1\sd{0.7}} & {55.4\sd{0.3}} & {79.4\sd{0.3}} \\
AdaptFormer & {75.8\sd{0.5}} & {90.6\sd{0.4}} & {68.6\sd{0.6}} & {98.6\sd{0.1}} & {87.9\sd{0.3}} & {82.8\sd{0.5}} & {55.1\sd{0.4}} & {79.9\sd{0.2}} \\
SSF & {74.3\sd{0.6}} & {89.1\sd{0.5}} & {66.1\sd{0.8}} & {96.3\sd{0.2}} & {87.6\sd{0.4}} & {81.5\sd{0.6}} & {55.6\sd{0.3}} & {78.6\sd{0.3}} \\
FacT-TT & {76.8\sd{0.4}} & {90.2\sd{0.6}} & {67.3\sd{0.7}} & {96.6\sd{0.2}} & {87.6\sd{0.3}} & {81.9\sd{0.5}} & {55.9\sd{0.4}} & {79.5\sd{0.3}} \\
RepAdapter & {76.0\sd{0.5}} & {89.2\sd{0.7}} & {66.5\sd{0.9}} & {95.9\sd{0.3}} & {86.6\sd{0.5}} & {82.8\sd{0.6}} & {53.4\sd{0.4}} & {78.6\sd{0.4}} \\
NOAH & {77.5\sd{0.7}} & {89.0\sd{0.8}} & {67.1\sd{1.0}} & {96.8\sd{0.3}} & {87.6\sd{0.6}} & {84.1\sd{0.8}} & {56.3\sd{0.5}} & {79.8\sd{0.5}} \\
\midrule
VPT-Shallow & {75.0\sd{0.6}} & {86.6\sd{0.7}} & {66.7\sd{0.9}} & {97.3\sd{0.2}} & {84.7\sd{0.4}} & {82.0\sd{0.5}} & {51.7\sd{0.3}} & {77.7\sd{0.3}} \\
VPT-Deep & {76.2\sd{0.5}} & {88.0\sd{0.8}} & {66.3\sd{0.9}} & {96.2\sd{0.2}} & {86.7\sd{0.5}} & {83.2\sd{0.6}} & {53.4\sd{0.4}} & {78.6\sd{0.4}} \\
$E^2$VPT & {75.5\sd{0.6}} & {89.8\sd{0.5}} & {67.8\sd{0.8}} & {96.9\sd{0.2}} & {87.3\sd{0.6}} & {84.5\sd{0.7}} & {55.6\sd{0.4}} & {79.6\sd{0.4}} \\
CVPT & {77.2\sd{0.4}} & {90.2\sd{0.6}} & {67.4\sd{0.7}} & {97.6\sd{0.2}} & {89.6\sd{0.3}} & {82.4\sd{0.6}} & {55.9\sd{0.3}} & {80.0\sd{0.3}} \\
VAPT & {77.1\sd{0.5}} & {90.2\sd{0.4}} & {69.0\sd{0.6}} & {98.6\sd{0.1}} & {89.9\sd{0.4}} & {83.6\sd{0.5}} & {56.3\sd{0.4}} & {80.7\sd{0.3}} \\
VPT-Deep + PAE & {76.3\sd{0.6}} & {91.7\sd{0.5}} & {67.7\sd{0.9}} & {97.5\sd{0.2}} & {87.5\sd{0.4}} & {82.5\sd{0.8}} & {55.2\sd{0.5}} & {79.8\sd{0.4}} \\
\bottomrule
\end{tabularx}
\end{table}

\begin{table}[p]
\caption{\textbf{Per-task endpoint accuracy on specialized VTAB-1k tasks.}
Cells report \rev{5-seed} mean top-1 accuracy \rev{($\pm$ SD)} (\%); the final column averages this task group \rev{($\pm$ 5-seed SD). Per-task 5-seed SDs across Tables~\ref{tab:vtab_natural}--\ref{tab:vtab_structured} span $0.2$--$1.1$ pp in VPT-Shallow, $0.2$--$1.2$ pp in VPT-Deep, and $0.1$--$1.4$ pp across all 15 methods.}}
\label{tab:vtab_specialized}
\centering
\small
\setlength{\tabcolsep}{3pt}
\renewcommand{\arraystretch}{0.95}
\setlength{\aboverulesep}{0.25ex}
\setlength{\belowrulesep}{0.35ex}
\begin{tabularx}{\linewidth}{L{1.65}*{5}{C{0.87000000}}}
\toprule
& \multicolumn{4}{c}{Specialized (4)} & \\
\cmidrule(lr){2-5}
Endpoint trainer & \shortstack{Patch\\Camelyon} & EuroSAT & RESISC45 & Retinopathy & \textbf{Mean} \\
\midrule
Linear probe & {79.6\sd{0.4}} & {89.7\sd{0.2}} & {82.1\sd{0.3}} & {71.1\sd{0.5}} & {80.6\sd{0.2}} \\
Full fine-tuning & {84.2\sd{1.1}} & {94.7\sd{0.4}} & {86.0\sd{0.6}} & {74.9\sd{0.9}} & {85.0\sd{0.5}} \\
\midrule
BitFit & {82.8\sd{0.7}} & {92.6\sd{0.4}} & {81.9\sd{0.5}} & {73.2\sd{0.8}} & {82.6\sd{0.4}} \\
LoRA & {85.7\sd{0.6}} & {94.0\sd{0.3}} & {85.1\sd{0.4}} & {76.1\sd{0.7}} & {85.2\sd{0.3}} \\
AdaptFormer & {84.1\sd{0.7}} & {94.4\sd{0.3}} & {85.0\sd{0.5}} & {78.1\sd{0.6}} & {85.4\sd{0.3}} \\
SSF & {84.2\sd{0.6}} & {96.5\sd{0.2}} & {87.1\sd{0.4}} & {76.2\sd{0.7}} & {86.0\sd{0.3}} \\
FacT-TT & {86.5\sd{0.7}} & {94.5\sd{0.4}} & {87.4\sd{0.5}} & {76.7\sd{0.8}} & {86.3\sd{0.4}} \\
RepAdapter & {85.9\sd{0.5}} & {96.3\sd{0.3}} & {86.1\sd{0.4}} & {76.5\sd{0.6}} & {86.2\sd{0.3}} \\
NOAH & {86.7\sd{0.8}} & {96.3\sd{0.3}} & {86.9\sd{0.6}} & {77.4\sd{0.9}} & {86.8\sd{0.4}} \\
\midrule
VPT-Shallow & {82.9\sd{0.8}} & {92.2\sd{0.4}} & {82.6\sd{0.5}} & {75.3\sd{0.9}} & {83.3\sd{0.4}} \\
VPT-Deep & {84.5\sd{0.7}} & {94.9\sd{0.3}} & {84.8\sd{0.6}} & {78.4\sd{0.8}} & {85.7\sd{0.4}} \\
$E^2$VPT & {86.6\sd{0.6}} & {95.8\sd{0.3}} & {85.6\sd{0.5}} & {76.7\sd{0.7}} & {86.2\sd{0.3}} \\
CVPT & {85.6\sd{0.7}} & {95.7\sd{0.4}} & {86.7\sd{0.4}} & {76.8\sd{0.8}} & {86.2\sd{0.4}} \\
VAPT & {87.2\sd{0.5}} & {95.2\sd{0.3}} & {86.3\sd{0.4}} & {76.9\sd{0.6}} & {86.4\sd{0.3}} \\
VPT-Deep + PAE & {84.6\sd{0.8}} & {96.8\sd{0.2}} & {86.2\sd{0.5}} & {77.8\sd{0.7}} & {86.4\sd{0.4}} \\
\bottomrule
\end{tabularx}
\end{table}

\begin{table}[p]
\caption{\textbf{Per-task endpoint accuracy on structured VTAB-1k tasks.}
Cells report \rev{5-seed} mean top-1 accuracy \rev{($\pm$ SD)} (\%); the final column averages this task group \rev{($\pm$ 5-seed SD). Per-task 5-seed SDs across Tables~\ref{tab:vtab_natural}--\ref{tab:vtab_structured} span $0.2$--$1.1$ pp in VPT-Shallow, $0.2$--$1.2$ pp in VPT-Deep, and $0.1$--$1.4$ pp across all 15 methods.}}
\label{tab:vtab_structured}
\centering
\small
\setlength{\tabcolsep}{1.8pt}
\renewcommand{\arraystretch}{0.95}
\setlength{\aboverulesep}{0.25ex}
\setlength{\belowrulesep}{0.35ex}
\begin{tabularx}{\linewidth}{L{1.64}*{9}{C{0.92888889}}}
\toprule
& \multicolumn{8}{c}{Structured (8)} & \\
\cmidrule(lr){2-9}
Endpoint trainer & \shortstack{CLEVR\\count} & \shortstack{CLEVR\\distance} & DMLab & \shortstack{KITTI\\distance} & \shortstack{dSprites\\location} & \shortstack{dSprites\\orient.} & \shortstack{sNORB\\azimuth} & \shortstack{sNORB\\elev.} & \textbf{Mean} \\
\midrule
Linear probe & {52.9\sd{0.5}} & {37.3\sd{0.4}} & {22.8\sd{0.3}} & {48.2\sd{0.7}} & {52.2\sd{0.5}} & {26.7\sd{0.3}} & {8.0\sd{0.2}} & {17.8\sd{0.4}} & {33.2\sd{0.3}} \\
Full fine-tuning & {74.5\sd{1.1}} & {56.2\sd{1.3}} & {45.3\sd{0.9}} & {68.8\sd{1.4}} & {71.0\sd{1.2}} & {45.9\sd{0.8}} & {27.1\sd{1.1}} & {38.5\sd{1.0}} & {53.4\sd{0.7}} \\
\midrule
BitFit & {66.9\sd{0.8}} & {49.9\sd{0.9}} & {36.3\sd{0.6}} & {59.6\sd{1.1}} & {62.6\sd{0.9}} & {40.3\sd{0.7}} & {20.2\sd{0.8}} & {29.8\sd{0.6}} & {45.7\sd{0.5}} \\
LoRA & {80.5\sd{0.6}} & {61.9\sd{0.8}} & {47.8\sd{0.7}} & {73.3\sd{1.0}} & {77.0\sd{0.9}} & {52.0\sd{0.5}} & {30.7\sd{0.8}} & {44.2\sd{0.6}} & {58.4\sd{0.5}} \\
AdaptFormer & {80.1\sd{0.7}} & {63.8\sd{0.7}} & {48.7\sd{0.6}} & {72.5\sd{0.9}} & {77.0\sd{0.8}} & {53.2\sd{0.6}} & {31.9\sd{0.7}} & {43.5\sd{0.5}} & {58.8\sd{0.4}} \\
SSF & {79.0\sd{0.8}} & {62.1\sd{0.9}} & {49.1\sd{0.7}} & {71.8\sd{1.1}} & {76.1\sd{0.8}} & {52.8\sd{0.6}} & {33.3\sd{0.9}} & {44.1\sd{0.7}} & {58.5\sd{0.5}} \\
FacT-TT & {82.2\sd{0.6}} & {63.9\sd{0.8}} & {51.4\sd{0.6}} & {74.4\sd{0.9}} & {77.0\sd{0.7}} & {51.7\sd{0.5}} & {33.1\sd{0.8}} & {43.7\sd{0.6}} & {59.7\sd{0.4}} \\
RepAdapter & {80.7\sd{0.7}} & {63.7\sd{0.9}} & {48.1\sd{0.8}} & {74.8\sd{1.0}} & {78.5\sd{0.9}} & {52.2\sd{0.6}} & {33.3\sd{0.7}} & {43.2\sd{0.7}} & {59.3\sd{0.5}} \\
NOAH & {81.8\sd{0.9}} & {63.5\sd{1.0}} & {49.5\sd{0.8}} & {74.2\sd{1.3}} & {78.9\sd{1.1}} & {52.4\sd{0.7}} & {32.2\sd{0.9}} & {45.4\sd{0.8}} & {59.7\sd{0.6}} \\
\midrule
VPT-Shallow & {69.8\sd{0.8}} & {53.0\sd{1.0}} & {40.0\sd{0.7}} & {64.9\sd{1.1}} & {67.9\sd{0.9}} & {41.8\sd{0.6}} & {24.1\sd{0.9}} & {34.4\sd{0.7}} & {49.5\sd{0.6}} \\
VPT-Deep & {77.8\sd{0.9}} & {59.7\sd{1.0}} & {46.8\sd{0.7}} & {70.6\sd{1.2}} & {76.6\sd{0.9}} & {51.0\sd{0.8}} & {32.0\sd{1.1}} & {41.7\sd{0.6}} & {57.0\sd{0.6}} \\
$E^2$VPT & {79.7\sd{0.8}} & {62.2\sd{0.9}} & {49.5\sd{0.6}} & {74.9\sd{1.1}} & {77.8\sd{0.8}} & {53.8\sd{0.7}} & {34.2\sd{0.9}} & {44.5\sd{0.6}} & {59.6\sd{0.5}} \\
CVPT & {82.4\sd{0.6}} & {62.2\sd{0.8}} & {50.2\sd{0.7}} & {74.6\sd{0.9}} & {78.0\sd{0.7}} & {54.3\sd{0.5}} & {34.9\sd{0.8}} & {44.2\sd{0.6}} & {60.1\sd{0.4}} \\
VAPT & {83.4\sd{0.7}} & {63.9\sd{0.7}} & {52.2\sd{0.6}} & {73.8\sd{1.0}} & {79.3\sd{0.8}} & {53.7\sd{0.6}} & {34.3\sd{0.7}} & {44.9\sd{0.5}} & {60.7\sd{0.4}} \\
VPT-Deep + PAE & {81.1\sd{0.8}} & {64.5\sd{0.9}} & {51.1\sd{0.7}} & {75.9\sd{1.1}} & {77.7\sd{0.9}} & {53.9\sd{0.6}} & {34.7\sd{0.8}} & {45.1\sd{0.7}} & {60.5\sd{0.5}} \\
\bottomrule
\end{tabularx}
\end{table}

\clearpage
\section{Additional Qualitative Layouts}
\label{app:qualitative}

\begin{figure}[!ht]
  \centering
  \includegraphics[width=\linewidth]{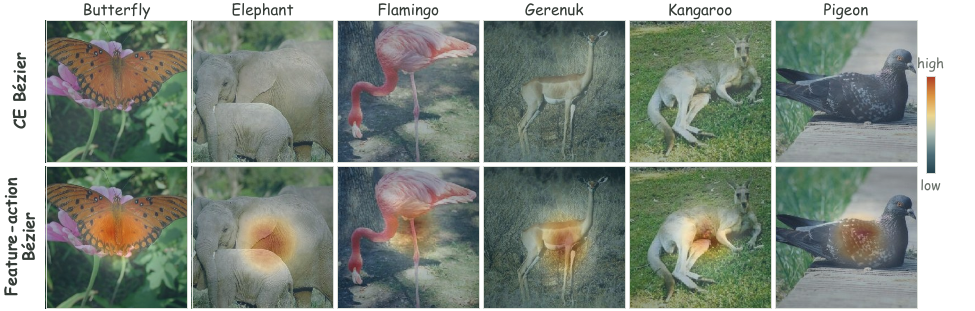}
  \caption{\textbf{Attention on six Caltech-101 animal photographs.}
  CE B\'ezier (top) and Feature-action B\'ezier (bottom).}
  \label{fig:attention_paths}
\end{figure}

Four additional $2\times6$ groups extend the layouts in
Figures~\ref{fig:saliency_paths} and~\ref{fig:attention_paths}.
Each column shares a crop and scale across CE B\'ezier (top) and
Feature-action B\'ezier (bottom).

\begin{figure}[!ht]
  \centering
  \includegraphics[width=\linewidth]{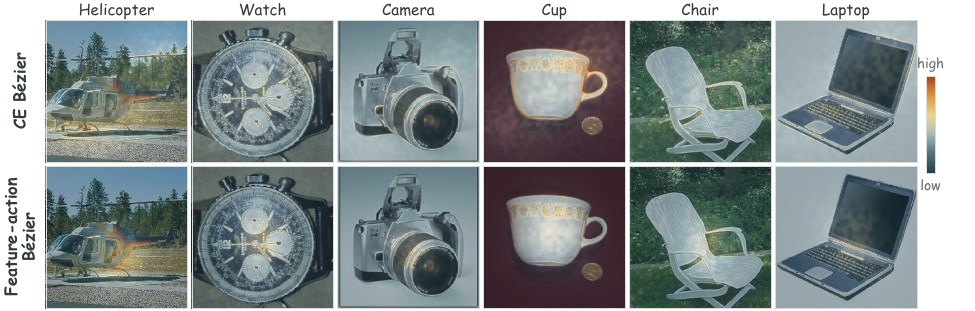}
  \caption{\textbf{Additional saliency: transport and everyday objects.}}
  \label{fig:saliency_objects}
\end{figure}

\begin{figure}[!ht]
  \centering
  \includegraphics[width=\linewidth]{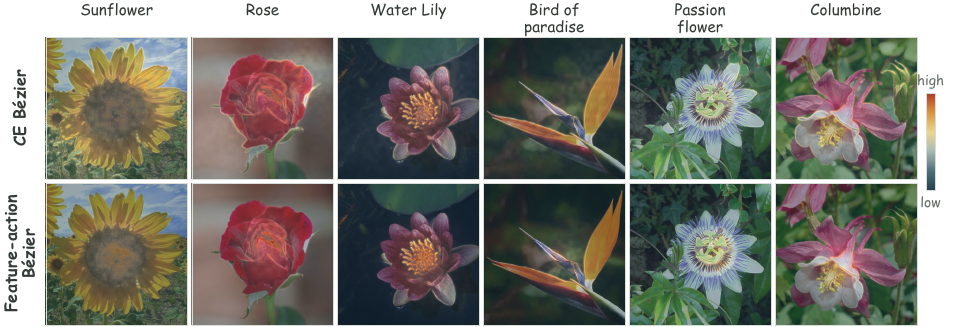}
  \caption{\textbf{Additional saliency: flower structure and texture.}}
  \label{fig:saliency_flowers}
\end{figure}

\clearpage
\subsection{Further attention cases}
\label{app:additional_attention}

\begin{figure}[!ht]
  \centering
  \includegraphics[width=\linewidth]{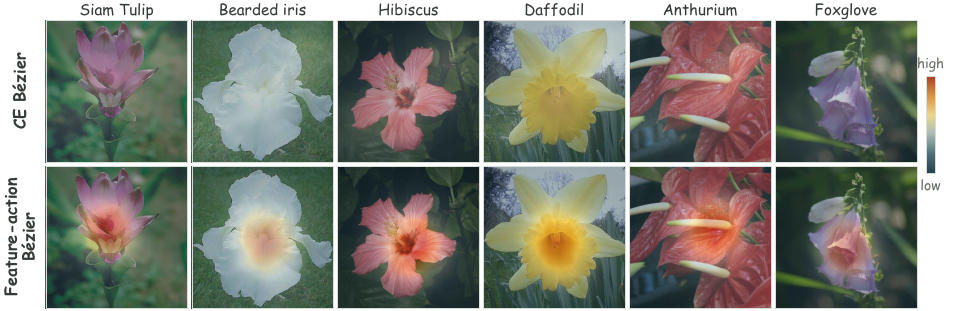}
  \caption{\textbf{Additional attention: six further flower categories.}}
  \label{fig:attention_flowers}
\end{figure}

\begin{figure}[!ht]
  \centering
  \includegraphics[width=\linewidth]{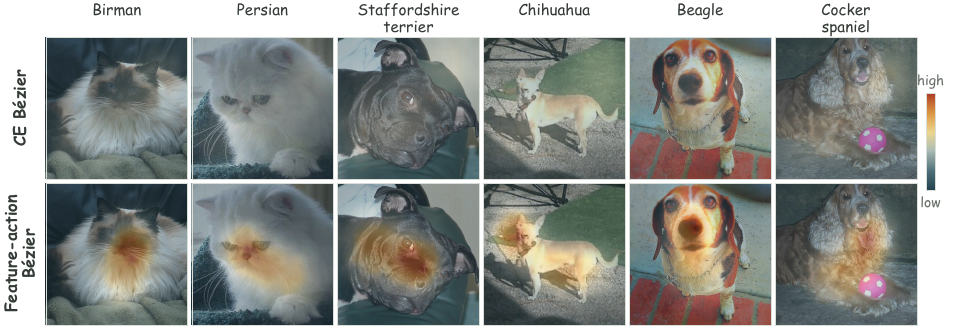}
  \caption{\textbf{Additional attention: pet poses and backgrounds.}}
  \label{fig:attention_pets}
\end{figure}

\clearpage
\section[Feature Width, LayerNorm, and Head Mixing]{Feature width, LayerNorm, and head mixing}
\label{app:architecture_extensions}

These extensions separate three architectural operations: adding output
coordinates, normalizing tokens before attention, and mixing independently
controlled attention responses. All variational endpoints remain output
laws, as in~(\ref{eq:action_problem}).
For the explicit heads below, fix $\beta\ne0$ with
$e^\beta<\min_iZ_i$.

\subsection{Vector outputs preserve the scalar optimum}
\label{app:vector_extension}

Let $d_{\mathrm{out}}\ge1$ be the output dimension and
$\boldsymbol w\in\mathbb R^{d_{\mathrm{out}}}$ a variable shared numerator.
Fix a unit output direction $\boldsymbol a$. The attainable laws and family are
\[
 \nu_{\kappa,\boldsymbol w}
 =\frac1N\sum_{i=1}^N\delta_{\boldsymbol w/(Z_i+\kappa)},\qquad
 \mathcal M^{\mathrm{vec}}
 =\{\nu_{\kappa,\boldsymbol w}:\kappa>0,\;
                   \boldsymbol w\in\mathbb R^{d_{\mathrm{out}}}\}.
\]
Euclidean distance defines vector-valued $W_2$ and its $AC^2$ action.
Write $\mathcal E_*^{\mathrm{vec}}$ for the constrained action infimum within
$\mathcal M^{\mathrm{vec}}$ between the laws
$\nu_{\kappa_b,w_b\boldsymbol a}$. Here $T_\#\nu$ is the pushforward
of $\nu$ by $T$.

\begin{proposition}[Vector outputs preserve the optimum]
\label{prop:vector_extension}
For every $d_{\mathrm{out}}\ge1$ and unit $\boldsymbol a$,
$\mathcal E_*^{\mathrm{vec}}=\mathcal E_*(0)$ and the vector endpoint
distance equals $D$. Thus Theorem~\ref{prop:action} and its strict-gap
threshold remain exact even when paths can leave the fixed-direction
slice $\boldsymbol w=w\boldsymbol a$.
\end{proposition}

\paragraph{Proof of Proposition~\ref{prop:vector_extension}.}
Let $T_{\boldsymbol a}(x)=\boldsymbol a^\top x$ and
$\iota_{\boldsymbol a}(r)=r\boldsymbol a$, where
$\|\boldsymbol a\|_{\mathrm E}=1$.
For any coupling $\gamma$ between vector laws $\nu,\widetilde\nu$,
$(T_{\boldsymbol a},T_{\boldsymbol a})_\#\gamma$ couples their projected
laws. Since
$|T_{\boldsymbol a}(x)-T_{\boldsymbol a}(y)|\le\|x-y\|_{\mathrm E}$,
taking the infimum over couplings gives
\begin{equation}
 W_2\bigl((T_{\boldsymbol a})_\#\nu,
          (T_{\boldsymbol a})_\#\widetilde\nu\bigr)
 \le W_2(\nu,\widetilde\nu).
 \label{eq:extension_projection_contraction}
\end{equation}
Moreover,
$(T_{\boldsymbol a})_\#\nu_{\kappa,\boldsymbol w}
=\mu_{\kappa,\boldsymbol a^\top\boldsymbol w}\in\mathcal M$.
Every admissible vector path therefore projects to a scalar path with
the prescribed endpoints. The contraction preserves the $AC^2$
inequality. At almost every time where both metric derivatives exist,
applying~(\ref{eq:extension_projection_contraction}) to difference
quotients bounds the projected speed by the vector speed. Integrating
their squares and taking the infimum gives
$\mathcal E_*^{\mathrm{vec}}\ge\mathcal E_*(0)$.
This permits atom matching, rotation of $\boldsymbol w$, zero outputs,
and sign changes of its projection.

Conversely, lifting any scalar coupling by
$(\iota_{\boldsymbol a},\iota_{\boldsymbol a})$ preserves its cost.
Together with~(\ref{eq:extension_projection_contraction}), this proves
\begin{equation}
 W_2\bigl((\iota_{\boldsymbol a})_\#\mu,
          (\iota_{\boldsymbol a})_\#\widetilde\mu\bigr)
 =W_2(\mu,\widetilde\mu).
 \label{eq:extension_vector_isometry}
\end{equation}
Since $(\iota_{\boldsymbol a})_\#\mu_{\kappa,w}
=\nu_{\kappa,w\boldsymbol a}$, every admissible scalar path lifts to
an admissible vector path with identical speed and action. Taking
infima proves the reverse inequality, without requiring scalar
attainment. The same isometry proves that the endpoint distance is $D$.
Under Theorem~\ref{prop:action}, lifting its attained optimum therefore
preserves both the exact minimum and the excess action.
\hfill\rule{0.5em}{0.5em}

\paragraph{A frozen vector head realizes the full family.}
Use token dimension $1+d_{\mathrm{out}}$ and write a token as
$p=(p^{(0)},\boldsymbol p)$.
Set $\textcolor{frozenblue}{K}(p)=p^{(0)}$, $\textcolor{frozenblue}{H}(p)=p^{(0)}/\beta$, and
$\textcolor{frozenblue}{V}(p)=\boldsymbol p$. The fixed readout and second content tokens are
$(\beta,\boldsymbol0)$ and
$(\log(Z_i-e^\beta),\boldsymbol0)$, respectively.
Their readout query is one, their content partition is $Z_i$, and
their vector content numerator is zero. For every $m\ge1$, the shared
tokens
\[
 \textcolor{promptorange}{p_j}(\kappa,\boldsymbol w)
 =\textcolor{promptorange}{\left(\log(\kappa/m),\frac{\boldsymbol w}{\kappa}\right)},
 \qquad j=1,\ldots,m,
\]
give aggregate mass $\kappa$ and vector numerator $\boldsymbol w$.
Conversely, arbitrary prompts produce a positive aggregate mass and
a shared vector numerator, so the attainable family is exactly
$\mathcal M^{\mathrm{vec}}$. The displayed lift is smooth for
$\kappa>0$, and the head remains fixed throughout each path.

\paragraph{Extrinsic bending and intrinsic flatness.}
The scalar positive branch has parametrization
$F(\kappa,w)=w u_\kappa$ and tangent plane
$\mathcal T_{\kappa,w}=\operatorname{span}\{u_\kappa,u'_\kappa\}$.
For $d_Z\ge3$, the independence in Appendix~\ref{app:threshold}
yields the nonzero normal component
\[
 \operatorname{Proj}_{\mathcal T_{\kappa,w}^{\perp}}
          \partial_\kappa^2 F
 =w\operatorname{Proj}_{\mathcal T_{\kappa,w}^{\perp}}
          u''_\kappa\ne0\qquad(w>0).
\]
Here $\operatorname{Proj}$ denotes orthogonal projection in the
normalized sample inner product. This nonzero normal component of a
second derivative proves local extrinsic bending. In contrast,
$\partial_w^2F=0$ and $\partial_w\partial_\kappa F=u'_\kappa$ is
tangent. The cone metric $dR^2+R^2d\theta^2$ is intrinsically flat
away from the apex. The fixed-direction vector slice is an isometric
copy of this bent cone; the full vector family has additional
directions, whose availability does not lower the action by
Proposition~\ref{prop:vector_extension}.

\subsection{Exact realization with pre-attention LayerNorm}
\label{app:layernorm_extension}

Let $d_{\mathrm{tok}}$ be the token dimension. For
$x\in\mathbb R^{d_{\mathrm{tok}}}$ and $\varepsilon_{\mathrm{LN}}\ge0$,
identity-affine token LayerNorm \citep{ba2016layernormalization} is
\[
 \textcolor{frozenblue}{\mathrm{LN}_{\varepsilon_{\mathrm{LN}}}}(x)
 =\frac{x-\bar x\boldsymbol1}
 {\sqrt{d_{\mathrm{tok}}^{-1}\|x-\bar x\boldsymbol1\|_{\mathrm E}^2
                +\varepsilon_{\mathrm{LN}}}},\qquad
 \bar x=d_{\mathrm{tok}}^{-1}\boldsymbol1^\top x.
\]
Here $\boldsymbol1$ is the all-ones vector; when
$\varepsilon_{\mathrm{LN}}=0$, raw tokens must be nonconstant.
Pre-attention LayerNorm applies this map before the frozen query, key,
and value maps.

\begin{proposition}[Exact realization with pre-attention LayerNorm]
\label{prop:layernorm_extension}
Fix $m$, $(Z_i)$, $A$, and compact
$\mathcal K\subset(0,\infty)\times\mathbb R$.
For every $d_{\mathrm{tok}}\ge4$ and
$\varepsilon_{\mathrm{LN}}\ge0$, one frozen pre-attention LayerNorm head
has attainable family contained in $\mathcal M_A$ and smoothly realizes
every aggregate pair in $\mathcal K$.
For $A=0$, choosing $\mathcal K$ to contain the optimal path in
Theorem~\ref{prop:action} preserves its exact minimum and gap.
\end{proposition}

\paragraph{Proof of Proposition~\ref{prop:layernorm_extension}.}
Fix $m$, $\mathcal K$, and $\varepsilon_{\mathrm{LN}}$ as in the
proposition.
Since $d_{\mathrm{tok}}\ge4$, the centered subspace
$\boldsymbol1^\perp$ contains three orthonormal vectors
$b_1,b_2,b_3$. The desired key--value pairs lie in
\begin{equation}
 \begin{split}
 \mathcal B={}&
 \left\{\left(\log(\kappa/m),w/\kappa\right):
                (\kappa,w)\in\mathcal K\right\}
 \cup\{(\beta,0)\}\,\cup{}\\[-2pt]
 &\left\{\left(\log(Z_i-e^\beta),
                     \frac{A_i}{Z_i-e^\beta}\right):1\le i\le N\right\}.
 \end{split}
 \label{eq:extension_ln_key_values}
\end{equation}
Compactness follows because $\kappa$ has a positive minimum on
$\mathcal K$ and $Z_i-e^\beta>0$ for every $i$.
If $\varepsilon_{\mathrm{LN}}=0$, set $r=\sqrt{d_{\mathrm{tok}}}$;
otherwise choose $0<r<\sqrt{d_{\mathrm{tok}}}$.
Take $\Lambda>0$ large enough that
$k^2+\xi^2<\Lambda^2r^2$ on a neighborhood of $\mathcal B$,
where $k,\xi$ are local scalar key and value coordinates. Define
\begin{equation}
 Y(k,\xi)=\frac{k}{\Lambda}b_1+\frac{\xi}{\Lambda}b_2
       +\sqrt{r^2-\frac{k^2+\xi^2}{\Lambda^2}}\,b_3.
 \label{eq:extension_ln_sphere}
\end{equation}
This map is smooth on that neighborhood, is centered, and has
constant Euclidean norm $r$. Its raw-token lift is
\begin{equation}
 \Psi(k,\xi)=
 \begin{cases}
 Y(k,\xi),&\varepsilon_{\mathrm{LN}}=0,\\[2pt]
 \displaystyle
 \sqrt{\frac{\varepsilon_{\mathrm{LN}}}{1-r^2/d_{\mathrm{tok}}}}
             \,Y(k,\xi),&\varepsilon_{\mathrm{LN}}>0.
 \end{cases}
 \label{eq:extension_ln_raw_lift}
\end{equation}
For zero epsilon, $\Psi$ has variance one, so its normalized value
is $Y$. For positive epsilon, let $a_{\mathrm{LN}}$ denote the
positive prefactor in the second line. The identity
$a_{\mathrm{LN}}^2r^2/d_{\mathrm{tok}}
 +\varepsilon_{\mathrm{LN}}=a_{\mathrm{LN}}^2$ yields
\begin{equation}
 \textcolor{frozenblue}{\mathrm{LN}_{\varepsilon_{\mathrm{LN}}}}(\Psi(k,\xi))=Y(k,\xi)
 \label{eq:extension_ln_inverse}
\end{equation}
in both cases.

On normalized tokens $y$, choose the frozen maps
$\textcolor{frozenblue}{K}(y)=\Lambda\langle b_1,y\rangle$,
$\textcolor{frozenblue}{V}(y)=\Lambda\langle b_2,y\rangle$, and $\textcolor{frozenblue}{H}(y)=\textcolor{frozenblue}{K}(y)/\beta$.
Equations~(\ref{eq:extension_ln_sphere})--(\ref{eq:extension_ln_inverse})
give key $k$ and value $\xi$ to raw token $\Psi(k,\xi)$.
Choose the fixed raw readout and second content tokens
\[
 \Psi(\beta,0),\qquad
 \Psi\!\left(\log(Z_i-e^\beta),\frac{A_i}{Z_i-e^\beta}\right).
\]
Their normalized readout query is one. Their two key weights sum to
$e^\beta+(Z_i-e^\beta)=Z_i$, and their weighted values sum to $A_i$.
The shared raw prompts
\begin{equation}
 \textcolor{promptorange}{p_j}(\kappa,w)=\textcolor{promptorange}{\Psi\!\left(\log(\kappa/m),w/\kappa\right)},
 \qquad j=1,\ldots,m,
 \label{eq:extension_ln_prompt_lift}
\end{equation}
have total key weight $\kappa$ and weighted value $w$. Thus they
produce $(A_i+w)/(Z_i+\kappa)$, smoothly on a neighborhood of
$\mathcal K$. Conversely, arbitrary valid raw prompts yield finite
normalized keys and values, and hence some $\kappa>0,w\in\mathbb R$.
The fixed content aggregates are unchanged, so every resulting law
belongs to $\mathcal M_A$.

For the action statement, set $A=0$ and take the optimal aggregate
path from Theorem~\ref{prop:action}. Its smooth image is compact in
$\{\kappa>0,w>0\}$; choose $\mathcal K$ to contain it. Every
admissible path of the constructed pre-LN head lies in $\mathcal M$,
so its action is at least $\mathcal E_*(0)$. The optimal scalar path
has the smooth lift~(\ref{eq:extension_ln_prompt_lift}) and attains
that bound. Its endpoint laws and distance are unchanged, proving
exact equality of the minimum and gap. Thus the three-normalizer gap
survives LayerNorm.
\hfill\rule{0.5em}{0.5em}

\paragraph{Attainable aggregates and the compact scope.}
For an arbitrary frozen pre-LN common-query head, the aggregate
formula~(\ref{eq:family}) still holds: normalization only changes
which keys and values shared tokens can produce. In the construction,
the frozen maps may depend on $m$ and $\mathcal K$, but are fixed
throughout the path. This scope is necessary. Normalized token norms
are at most $\sqrt{d_{\mathrm{tok}}}$, so every fixed linear key
map is bounded on them. If its scalar logits have absolute value
at most $C$, then $m e^{-C}\le\kappa\le m e^C$.
A fixed normalized head therefore cannot realize all positive
aggregate masses. Any finite collection of prompt counts can share
one construction by taking the union of their compact key--value
sets in~(\ref{eq:extension_ln_key_values}).

\paragraph{LayerNorm of the measured output can erase the obstruction.}
Let $d_{\mathrm{out}}\ge2$, let $b\in\mathbb R^{d_{\mathrm{out}}}$
be a fixed nonconstant vector, and consider the output features
$b+z_i(\kappa,w)\boldsymbol1$. Applying LayerNorm with the same
definition in output dimension gives, for every
$\varepsilon_{\mathrm{LN}}\ge0$,
\[
 \textcolor{frozenblue}{\mathrm{LN}_{\varepsilon_{\mathrm{LN}}}}
       (b+z_i(\kappa,w)\boldsymbol1)
 =\textcolor{frozenblue}{\mathrm{LN}_{\varepsilon_{\mathrm{LN}}}}(b).
\]
Indeed, subtracting the coordinate mean removes the added scalar
multiple of $\boldsymbol1$, leaving the variance unchanged.
Before normalization, this is an injective affine embedding of the
scalar response, with distances multiplied by
$\sqrt{d_{\mathrm{out}}}$. Afterwards the entire family is constant
and its gap is zero. The example has nonzero variance even at zero
epsilon. It distinguishes preservation by the pre-attention
realization from a universal claim about normalized output features.

\subsection{Concatenated heads and output mixing}
\label{app:multihead_extension}

For $h\in\{1,2\}$, superscript $(h)$ indexes head $h$'s
independently controlled prompt aggregates $\kappa^{(h)}>0$ and
$w^{(h)}\in\mathbb R$; $y_i$ denotes the sum of their outputs.

\begin{proposition}[Exact action for aligned concatenated heads]
\label{prop:concatenated_heads}
Consider $n_{\mathrm h}\ge1$ independently controlled scalar heads
with zero content numerators and positive normalizers
$Z_i^{(h)}$, $h=1,\ldots,n_{\mathrm h}$, each with at least two distinct
values. Suppose every head orders
the samples in the same way, so
$Z_1^{(h)}\ge\cdots\ge Z_N^{(h)}$ for all $h$.
Measure their unmixed concatenated output law
\[
 \nu_{\boldsymbol\kappa,\boldsymbol w}
 =\frac1N\sum_{i=1}^N\delta_{(z_i^{(1)},\ldots,z_i^{(n_{\mathrm h})})},
 \qquad z_i^{(h)}=\frac{w^{(h)}}{Z_i^{(h)}+\kappa^{(h)}}.
\]
For each head, fix endpoint aggregates with $w_0^{(h)},w_1^{(h)}>0$,
$\kappa_0^{(h)}\ne\kappa_1^{(h)}$, and developed angle $\Delta_h<\pi$ as in
Theorem~\ref{prop:action}. Write $\mathcal E_{*,h}$, $D_h$, and
$\mathcal G_h=\mathcal E_{*,h}-D_h^2$ for its scalar quantities.
For the concatenated family, with independent head controls,
\begin{equation}
 \mathcal E_*^{\mathrm{cat}}=\sum_{h=1}^{n_{\mathrm h}}\mathcal E_{*,h},
 \qquad D_{\mathrm{cat}}^2=\sum_{h=1}^{n_{\mathrm h}}D_h^2,
 \qquad \mathcal G_{\mathrm{cat}}=\sum_{h=1}^{n_{\mathrm h}}\mathcal G_h.
 \label{eq:concatenated_heads_decomposition}
\end{equation}
The same identities hold after an isometric embedding of the
concatenated output. In particular, one head with at least three
distinct normalizers gives a strictly positive concatenated gap.
\end{proposition}

\paragraph{Proof of Proposition~\ref{prop:concatenated_heads}.}
For any two vector laws $\nu,\widetilde\nu$, project an optimal
$W_2$ coupling onto each coordinate. Each projection couples the
corresponding scalar marginals, so summing their costs gives
\begin{equation}
 W_2^2(\nu,\widetilde\nu)
 \ge\sum_{h=1}^{n_{\mathrm h}}
 W_2^2((\pi_h)_\#\nu,(\pi_h)_\#\widetilde\nu),
 \label{eq:concatenated_heads_contraction}
\end{equation}
where $\pi_h$ is coordinate projection. Every admissible
concatenated path therefore projects to an admissible path for each
head. Taking metric-derivative limits in
(\ref{eq:concatenated_heads_contraction}) at almost every time and
integrating gives $\mathcal E_*^{\mathrm{cat}}
\ge\sum_h\mathcal E_{*,h}$.

For the reverse bound, run the attained scalar optima from
Theorem~\ref{prop:action} simultaneously. Independent controls
realize this joint path, and all its numerators stay positive.
The common normalizer order makes the same-index matching optimal
in every coordinate between any two times. Its vector cost is the
sum of these scalar optimal costs, so equality holds in
(\ref{eq:concatenated_heads_contraction}) along this path.
Consequently its action is $\sum_h\mathcal E_{*,h}$.
The same matching argument at the positive endpoints gives
$D_{\mathrm{cat}}^2=\sum_hD_h^2$; subtracting yields the gap identity.
An isometric embedding preserves every $W_2$ distance and action.
\hfill\rule{0.5em}{0.5em}

Independent aggregate control is realized by a fixed shared-prompt
block with two token coordinates per head. Choose
$\beta_h\ne0$ with $e^{\beta_h}<\min_iZ_i^{(h)}$ and repeat the
frozen construction below in disjoint coordinate blocks; head $h$ receives
$(\log(\kappa^{(h)}/m),w^{(h)}/\kappa^{(h)})$ in every shared token,
and its frozen maps select that coordinate pair. The decomposition
does not apply to a non-isometric output projection that sums or
otherwise mixes head responses.

\begin{proposition}[Two mixed heads can remove the gap]
\label{prop:two_head_mixing}
Two independently controlled scalar heads with the same $(Z_i)$,
zero content numerators, and summed outputs connect any positive scalar
endpoint laws $\mu_0,\mu_1$ with action $D^2$.
An explicit frozen block with four-dimensional shared tokens realizes
this connection for every $m\ge1$.
\end{proposition}

\paragraph{Proof of Proposition~\ref{prop:two_head_mixing}.}
The summed output of two independently controlled heads is
\[
 y_i=\frac{w^{(1)}}{Z_i+\kappa^{(1)}}
     +\frac{w^{(2)}}{Z_i+\kappa^{(2)}}.
\]
Keep $\kappa^{(1)}=\kappa_0$ and $\kappa^{(2)}=\kappa_1$ fixed,
and set $w^{(1)}(t)=(1-t)w_0$, $w^{(2)}(t)=tw_1$. Then
\begin{equation}
 y_i(t)=(1-t)\frac{w_0}{Z_i+\kappa_0}
             +t\frac{w_1}{Z_i+\kappa_1}.
 \label{eq:extension_two_head_chord}
\end{equation}
The positive scalar endpoint vectors are increasingly ordered under
the prescribed ordering of $(Z_i)$. Their convex combinations remain
ordered. The sorted-atom isometry therefore identifies the empirical
law of~(\ref{eq:extension_two_head_chord}) with the constant-speed
chord between the sorted endpoints. Its action equals $D^2$, which
bounds every path's action below. Thus this two-head family has zero
excess action between the original endpoint laws, including when
their single-head gap is positive.

To realize independent control by shared prompts, take
$p=(p^{(1)},p^{(2)},p^{(3)},p^{(4)})\in\mathbb R^4$ and frozen maps
\[
 \begin{array}{lll}
 \textcolor{frozenblue}{H_1}(p)=p^{(1)}/\beta,&\textcolor{frozenblue}{K_1}(p)=p^{(1)},&\textcolor{frozenblue}{V_1}(p)=p^{(2)},\\
 \textcolor{frozenblue}{H_2}(p)=p^{(3)}/\beta,&\textcolor{frozenblue}{K_2}(p)=p^{(3)},&\textcolor{frozenblue}{V_2}(p)=p^{(4)}.
 \end{array}
\]
For each sample, use readout token $(\beta,0,\beta,0)$ and second
content token
$(\log(Z_i-e^\beta),0,\log(Z_i-e^\beta),0)$.
Both heads have readout query one, content partition $Z_i$, and
zero content numerator. A frozen output projection sums their
outputs. For any positive masses and real numerators, the shared
prompts
\[
 \textcolor{promptorange}{p_j}=\textcolor{promptorange}{\left(
 \log\frac{\kappa^{(1)}}m,\frac{w^{(1)}}{\kappa^{(1)}},
 \log\frac{\kappa^{(2)}}m,\frac{w^{(2)}}{\kappa^{(2)}}
 \right)},\qquad j=1,\ldots,m,
\]
realize the four aggregates independently. Conversely, arbitrary
shared prompts produce such aggregates in their separate coordinate
blocks. Hence this one frozen block realizes the full independent
two-head family for every $m\ge1$. The chord uses the explicit path
\begin{equation}
 \textcolor{promptorange}{p_j}(t)=\textcolor{promptorange}{\left(
 \log\frac{\kappa_0}m,\frac{(1-t)w_0}{\kappa_0},
 \log\frac{\kappa_1}m,\frac{tw_1}{\kappa_1}
 \right)}.
 \label{eq:extension_two_head_prompts}
\end{equation}
This path is smooth and token-linear, with finite parameters and
strictly positive partitions throughout. At each endpoint one
numerator is zero and the other produces the prescribed law.
\hfill\rule{0.5em}{0.5em}

The construction specifies endpoint prompt arrays whose output laws
are $\mu_0,\mu_1$. It does not prescribe an optimal path between
arbitrary preassigned multihead prompt arrays. Its mechanism is the
independent addition of two response families before feature motion
is measured.

\paragraph{A protected concatenated channel retains the obstruction.}
For comparison, suppose the measured feature concatenates the vector
head of Proposition~\ref{prop:vector_extension} with other head
outputs. At both endpoints let the latter equal the same
$r_i\in\mathbb R^{d_{\mathrm{aux}}}$ for sample $i$, where
$d_{\mathrm{aux}}$ is their total dimension. Assume the scalar
optimum can be lifted in the first block while holding these other
outputs fixed. Under Theorem~\ref{prop:action}, the minimum action
of the concatenated family equals $\mathcal E_*(0)$, its endpoint
distance equals $D$, and its gap equals $\mathcal G$.

To prove the lower bound, project onto $\boldsymbol a$ in the first
block and ignore the remaining coordinates. This is a contraction
that sends every admissible law into $\mathcal M$, with the original
scalar endpoints. The action is therefore at least
$\mathcal E_*(0)$. For the attaining path, use features
$(z_i(t)\boldsymbol a,r_i)$, with $z_i(t)$ the positive scalar
optimum. For any two times, matching the same sample indices has
cost $N^{-1}\sum_i|z_i(t)-z_i(s)|^2$, which equals the scalar
Wasserstein cost because the scalar vectors remain ordered.
Projection gives the reverse bound, so these distances, metric
speeds, and actions are exactly the scalar ones. The same argument
at the endpoints proves the distance claim. If coupled prompt
controls prevent this lift, the projection still gives the action
lower bound whenever the stated projected family and endpoints are
retained. Thus concatenation with a protected scalar channel and
summation of independently normalized responses have different
variational effects.

\paragraph{An approximate projected-channel certificate.}
The protected channel need not be exact. Let
$\mathcal V\subset\mathcal P_{2,N}(\mathbb R^{d_{\mathrm{out}}})$ be any
attainable vector-feature family, with endpoints $\nu_0,\nu_1\in\mathcal V$.
Write $\mathcal E_*^{\mathcal V}$ for the infimum of actions of $AC^2$ paths
in $\mathcal V$ between these endpoints. For a unit vector
$\boldsymbol a\in\mathbb R^{d_{\mathrm{out}}}$, let
$\pi_{\boldsymbol a}(x)=\boldsymbol a^\top x$ and
$(\pi_{\boldsymbol a})_\#\nu$ denote the projected law. Write
$\operatorname{dist}_{W_2}(\lambda,\mathcal M)
=\inf_{\mu\in\mathcal M}W_2(\lambda,\mu)$.

\begin{proposition}[Projected-channel gap certificate]
\label{prop:protected_projection}
Suppose $d_Z\ge3$, $w_0,w_1>0$, and $\kappa_0\ne\kappa_1$.
Assume $\varepsilon\ge0$, $(\pi_{\boldsymbol a})_\#\nu_b=\mu_b$
for $b\in\{0,1\}$, and, for every $\nu\in\mathcal V$,
\begin{equation}
 \operatorname{dist}_{W_2}
   ((\pi_{\boldsymbol a})_\#\nu,\mathcal M)\le\varepsilon.
 \label{eq:projected_channel_assumption}
\end{equation}
Set $\tau=W_2^2(\nu_0,\nu_1)-D^2\ge0$. Then
\begin{equation}
 \mathcal E_*^{\mathcal V}-W_2^2(\nu_0,\nu_1)
 \ge 4[\rho-\varepsilon]_+^2-\tau.
 \label{eq:projected_channel_gap}
\end{equation}
In particular, the vector-feature gap is positive when
$2[\rho-\varepsilon]_+>\sqrt\tau$.
\end{proposition}

\paragraph{Proof of Proposition~\ref{prop:protected_projection}.}
Projection by $\pi_{\boldsymbol a}$ is $1$-Lipschitz, so it contracts
$W_2$ and metric action; in particular, it gives $\tau\ge0$.
For any admissible $\nu_t$, put
$\omega_t=Q((\pi_{\boldsymbol a})_\#\nu_t)$.
The projected law path is $AC^2$, and the sorted-vector isometry
makes $\omega_t$ an $AC^2$ path in the normalized sample Hilbert space
with action no larger than that of $\nu_t$.
Cauchy--Schwarz and the parallelogram identity give
\begin{align*}
 \mathcal E[\nu_\cdot]
 &\ge 2\|\omega_{1/2}-\omega_0\|_N^2
       +2\|\omega_1-\omega_{1/2}\|_N^2\\
 &=D^2+4\|\omega_{1/2}-h_0\|_N^2.
\end{align*}
By~(\ref{eq:projected_channel_assumption}) and the definition of $\rho$,
$\|\omega_{1/2}-h_0\|_N\ge[\rho-\varepsilon]_+$.
Taking the infimum and subtracting
$W_2^2(\nu_0,\nu_1)=D^2+\tau$ proves
(\ref{eq:projected_channel_gap}).
\hfill\rule{0.5em}{0.5em}

For sample-aligned endpoints of the form
$x_{i,b}=z_i(\kappa_b,w_b)\boldsymbol a+r_{i,b}$ with
$\boldsymbol a^\top r_{i,b}=0$, the same-index coupling gives the
checkable estimate
$\tau\le N^{-1}\sum_i\|r_{i,1}-r_{i,0}\|_{\mathrm E}^2$.
A uniform bound on projected auxiliary-head responses can likewise
verify~(\ref{eq:projected_channel_assumption}) by same-index coupling.
Small mixing alone does not ensure a family-wide bound on the
independently controlled head responses.

\clearpage
\section{Action bounds for prompt transitions and adaptation}
\label{app:application_bounds}

This section translates feature action into bounds on representation
and prediction changes. Throughout, the $AC^2$ trajectories
$f_i:[0,1]\to\mathbb R^{d_{\mathrm{out}}}$ track the same $N$ images,
as in Section~\ref{sec:extension_notation}. Define their straight
endpoint bridges $\bar f_i(t)=(1-t)f_i(0)+tf_i(1)$ and
\[
 \mathcal A_{\mathrm{pair}}=\frac1N\sum_i\int_0^1
 \|\dot f_i(t)\|_{\mathrm E}^2\,dt,\qquad
 \mathcal B_{\mathrm{pair}}=\frac1N\sum_i
 \|f_i(1)-f_i(0)\|_{\mathrm E}^2,\qquad
 \mathcal R_{\mathrm{pair}}=\mathcal A_{\mathrm{pair}}-\mathcal B_{\mathrm{pair}}.
\]
For $[s,t]\subseteq[0,1]$, let $\mathcal A_{\mathrm{pair}}[s,t]$
restrict the integral to that interval. The bounds below use Euclidean
energy estimates in any output dimension.

\subsection{Prompt transitions with a fixed classifier}
\label{app:transition_bounds}

\paragraph{Continuous empirical action tracks individual motion.}
Define $\nu_t=N^{-1}\sum_i\delta_{f_i(t)}$. The identity coupling gives
$W_2^2(\nu_s,\nu_t)\le N^{-1}\sum_i
\|f_i(t)-f_i(s)\|_{\mathrm E}^2$. In continuous time, the actions agree:
\begin{equation}
 \int_s^t|\dot\nu_r|_{W_2}^2\,dr
 =\mathcal A_{\mathrm{pair}}[s,t].
 \label{eq:empirical_tracked_action}
\end{equation}
Fix a time $r$ and group atoms that coincide at $r$.
If there are distinct groups, let $\mathrm{sep}(r)>0$ be their minimum
separation. For small $h$, every $f_i(r+h)$ stays within
$\mathrm{sep}(r)/2$ of its group's position. An optimal coupling can be
chosen as a permutation \citep{peyre2019computational}.
A permutation crossing groups costs at least $\mathrm{sep}(r)^2/(4N)$,
while the identity cost tends to zero. Thus optimal permutations
stay within groups for small $h$. Since each group has identical
starting positions, every within-group permutation has the identity
cost. If all atoms coincide at $r$, the conclusion holds directly.
Therefore
\[
 W_2^2(\nu_{r+h},\nu_r)
 =\frac1N\sum_i\|f_i(r+h)-f_i(r)\|_{\mathrm E}^2
\]
for sufficiently small $h$. Dividing by $h^2$ and differentiating the
finitely many $AC^2$ trajectories proves the speed identity almost
everywhere. The identity-coupling bound also makes $\nu_\cdot$ an
$AC^2$ path, so integration proves
(\ref{eq:empirical_tracked_action}), including paths with collisions.

\paragraph{Excess action measures departures from the endpoint bridge.}
The main bridge bound is
\begin{equation}
 \frac1N\sum_i\sup_{t\in[0,1]}
 \|f_i(t)-\bar f_i(t)\|_{\mathrm E}^2
 \le\frac14\mathcal R_{\mathrm{pair}}.
 \label{eq:transition_bridge_bound}
\end{equation}
The matched endpoint cost satisfies
$\mathcal B_{\mathrm{pair}}\ge W_2^2(\nu_0,\nu_1)$, because the
identity coupling is one feasible coupling. Expanding the square and
using $\int_0^1\dot f_i(r)\,dr=f_i(1)-f_i(0)$ gives
\begin{equation}
 \mathcal R_{\mathrm{pair}}
 =\frac1N\sum_i\int_0^1
 \bigl\|\dot f_i(r)-[f_i(1)-f_i(0)]\bigr\|_{\mathrm E}^2\,dr
 \ge0.
 \label{eq:bridge_energy_identity}
\end{equation}
For each $i$, absolute continuity further gives
\[
 f_i(t)-\bar f_i(t)
 =\int_0^1\bigl(\mathbf1_{\{r\le t\}}-t\bigr)
       \bigl(\dot f_i(r)-[f_i(1)-f_i(0)]\bigr)\,dr.
\]
The squared integral norm of the scalar kernel is
$t(1-t)\le1/4$. Cauchy--Schwarz therefore yields
\[
 \|f_i(t)-\bar f_i(t)\|_{\mathrm E}^2
 \le\frac14\int_0^1
 \bigl\|\dot f_i(r)-[f_i(1)-f_i(0)]\bigr\|_{\mathrm E}^2\,dr.
\]
Taking a supremum in $t$, averaging, and applying
(\ref{eq:bridge_energy_identity}) proves
(\ref{eq:transition_bridge_bound}). The factor $1/4$ is sharp:
for one scalar trajectory $f_1(t)=\min\{t,1-t\}$, both endpoints
are zero, the action is one, and the maximum squared departure is
$1/4$. The endpoint bridges need not themselves be attainable by
prompts for the bound to hold.

On the positive scalar branch, all samples maintain the order of
their fixed normalizers. Identity matching is then optimal at the
endpoints, so $\mathcal B_{\mathrm{pair}}=D^2$ and
$\mathcal R_{\mathrm{pair}}=\mathcal E[\mu_\cdot]-D^2$.
In particular, the exact minimum-action construction gives
$\mathcal R_{\mathrm{pair}}=\mathcal G$. For arbitrary vector
trajectories, the matched endpoint cost must be retained; endpoint
Wasserstein matching may permute images.

\rev{\paragraph{Spherical readout lower bounds.}
When readout features are normalized to the unit sphere
$\mathbb S^{d_{\mathrm{out}}-1}$, every continuous trajectory remains on
$\mathbb S^{d_{\mathrm{out}}-1}$, so its continuous action is bounded below
by the endpoint optimal-transport cost under squared spherical geodesic
distance, $\mathrm{OT}_{\mathrm{sph}}^2(\nu_0,\nu_1)$. For paths connecting
fixed prompt tensors, identity tracking (\ref{eq:empirical_tracked_action})
further gives the fixed-image spherical lower bound
\[
 \mathcal A_{\mathrm{pair}}
 \ge B_{\mathrm{pair}}^{\mathrm{sph}}
 =\frac1N\sum_{i=1}^N\theta_i^2,
 \qquad
 \theta_i=\arccos\langle f_i(0),f_i(1)\rangle_{\mathrm E}.
\]
Table~\ref{tab:spherical_bounds} evaluates these geometric readout bounds
on the Table~\ref{tab:vit_paths} endpoints across the 95 pairs on 19 VTAB-1k
tasks; this spherical readout constraint is distinct from the Block~1
shared-attention mechanism in Table~\ref{tab:block1_diagnostics}.}

\begin{table}[t]
\color{revblue}
\caption{\rev{\textbf{Endpoint Wasserstein costs and spherical readout lower bounds on ViT-B/16 (mean $\pm$ within-task 5-pair SD).}
Evaluated across 95 endpoint pairs on 19 VTAB-1k tasks ($N=128$ held-out images per task).
Spherical OT gap is $\mathrm{OT}_{\mathrm{sph}}^2 - W_2^2$; fixed-image spherical gap is $B_{\mathrm{pair}}^{\mathrm{sph}} - W_2^2$.}}
\label{tab:spherical_bounds}
\centering
\small
\setlength{\tabcolsep}{3.5pt}
\renewcommand{\arraystretch}{0.95}
\setlength{\aboverulesep}{0.25ex}
\setlength{\belowrulesep}{0.35ex}
\begin{tabularx}{\linewidth}{L{1.10}C{.85}C{1.15}C{.85}C{1.05}C{1.20}}
\toprule
\rev{VPT mode} & \rev{Endpoint $W_2^2$} & \rev{8-segment FA action} & \rev{Sampled excess} & \rev{Spherical OT gap} & \rev{Fixed-image spherical gap} \\
\midrule
\rev{Shallow} & \rev{$1.92$\sd{0.14}} & \rev{$3.00$\sd{0.10}} & \rev{$1.08$\sd{0.08}} & \rev{$0.49$\sd{0.03} ($\mathrm{OT}_{\mathrm{sph}}^2=2.41$)} & \rev{$0.70$\sd{0.05} ($B_{\mathrm{pair}}^{\mathrm{sph}}=2.62$)} \\
\rev{Deep} & \rev{$2.28$\sd{0.17}} & \rev{$3.90$\sd{0.14}} & \rev{$1.62$\sd{0.11}} & \rev{$0.74$\sd{0.05} ($\mathrm{OT}_{\mathrm{sph}}^2=3.02$)} & \rev{$1.01$\sd{0.07} ($B_{\mathrm{pair}}^{\mathrm{sph}}=3.29$)} \\
\bottomrule
\end{tabularx}
\end{table}

\begin{corollary}[Prediction changes along a prompt transition]
\label{prop:transition_margin}
Let $\mathcal C$ be a finite class set and let
$\textcolor{frozenblue}{g_c}(x)=\textcolor{frozenblue}{\omega_c}^\top x+\textcolor{frozenblue}{\zeta_c}$, $c\in\mathcal C$, be fixed affine
logits, with $\textcolor{frozenblue}{\omega_c}\in\mathbb R^{d_{\mathrm{out}}}$ and
$\textcolor{frozenblue}{\zeta_c}\in\mathbb R$. Put
$\Lambda_{\mathrm{cls}}=\max_{c\ne c'}
\|\textcolor{frozenblue}{\omega_c}-\textcolor{frozenblue}{\omega_{c'}}\|_{\mathrm E}$, taking it as zero for a
single class. For $\mathfrak m>0$, let
$\mathcal I_{\mathfrak m}$ contain the images with a common endpoint
class $\chi_i\in\mathcal C$ such that
\[
 \textcolor{frozenblue}{g_{\chi_i}}(f_i(b))-\textcolor{frozenblue}{g_c}(f_i(b))\ge\mathfrak m
 \quad\text{for }b\in\{0,1\},\quad c\ne\chi_i.
\]
Let $\mathcal F_{\mathfrak m}$ consist of those
$i\in\mathcal I_{\mathfrak m}$ for which some $t\in[0,1]$ and
$c\ne\chi_i$ satisfy $\textcolor{frozenblue}{g_c}(f_i(t))\ge \textcolor{frozenblue}{g_{\chi_i}}(f_i(t))$.
Thus ties count as failures. Then
\begin{equation}
 \frac{|\mathcal F_{\mathfrak m}|}{N}
 \le\min\left\{\frac{|\mathcal I_{\mathfrak m}|}{N},\,
 \frac{\Lambda_{\mathrm{cls}}^2}{4\mathfrak m^2}
       \mathcal R_{\mathrm{pair}}\right\}.
 \label{eq:transition_margin_bound}
\end{equation}
\end{corollary}

\paragraph{Proof of Corollary~\ref{prop:transition_margin}.}
For $i\in\mathcal I_{\mathfrak m}$, every competing margin on
$\bar f_i(t)$ is the affine interpolation of its endpoint margins
and is at least $\mathfrak m$. At a failure time, Cauchy--Schwarz
therefore implies
$\mathfrak m\le\Lambda_{\mathrm{cls}}
\|f_i(t)-\bar f_i(t)\|_{\mathrm E}$.
When $\Lambda_{\mathrm{cls}}>0$, bound each failure indicator by
$\Lambda_{\mathrm{cls}}^2\mathfrak m^{-2}
\sup_t\|f_i(t)-\bar f_i(t)\|_{\mathrm E}^2$, sum over
$\mathcal I_{\mathfrak m}$, and apply
(\ref{eq:transition_bridge_bound}). If
$\Lambda_{\mathrm{cls}}=0$, all logit differences are constant,
so no such image can fail. The bound by
$|\mathcal I_{\mathfrak m}|/N$ follows from set inclusion.
\hfill\rule{0.5em}{0.5em}

For correct common endpoint classes, this also bounds classification
failures. Samples whose desired class changes require a task-dependent
criterion. Minimum action alone can pass through zero between opposite
feature endpoints; the margin condition is what gives prediction control.

\rev{\paragraph{Empirical evaluation of Corollary~\ref{prop:transition_margin} on ViT paths.}
At margin $\mathfrak m=0.80$ on the Table~\ref{tab:vit_paths} paths,
eligible images $\mathcal I_{\mathfrak m}$ cover $54.2\%$ (Shallow) and
$58.4\%$ (Deep) of held-out images; $\mathcal B_{\mathrm{pair}}$ is $2.06$
and $2.45$, and the probe Lipschitz constant $\Lambda_{\mathrm{cls}}$ has
task mean $0.88\pm 0.09$ (Shallow) and $0.80\pm 0.08$ (Deep). Using the
128-interval lower approximation
$\widehat{\mathcal R}_{\mathrm{pair}}=A_{\mathrm{pair}}-\mathcal B_{\mathrm{pair}}$,
the plug-in term $\frac{\Lambda_{\mathrm{cls}}^2}{4\mathfrak m^2}\widehat{\mathcal R}_{\mathrm{pair}}$
exceeds eligibility coverage on all 95 pairs for token-linear (sampled
excess $2.92$ and $4.83$) and CE B\'ezier ($2.61$ and $4.11$), and falls
below coverage only as an uncertified grid plug-in for FA B\'ezier ($38.1\%$
and $48.0\%$ mean). Because sampled action does not upper-bound continuous
action between grid nodes, these ViT experiments do not establish a
non-vacuous certified bound from (\ref{eq:transition_margin_bound});
empirically on the 128-interval grid, observed prediction-flip rates across
all held-out images are $2.3\%$ and $2.4\%$ for FA B\'ezier, $1.6\%$ and
$1.5\%$ for CE B\'ezier, and $7.8\%$ and $9.1\%$ for token-linear.}

\subsection{A drift budget for test-time adaptation}
\label{app:adaptation_bounds}

\paragraph{Feature and score drift on any interval.}
For $0\le s<t\le1$, absolute continuity and Cauchy--Schwarz give
\begin{equation}
 \frac1N\sum_i\|f_i(t)-f_i(s)\|_{\mathrm E}^2
 \le(t-s)\mathcal A_{\mathrm{pair}}[s,t].
 \label{eq:interval_feature_drift}
\end{equation}
Indeed, apply
$\|\int_s^t\dot f_i(r)\,dr\|_{\mathrm E}^2
\le(t-s)\int_s^t\|\dot f_i(r)\|_{\mathrm E}^2\,dr$
to each image and average. If $\psi:\mathbb R^{d_{\mathrm{out}}}
\to\mathbb R$ is $L_\psi$-Lipschitz, then
\begin{equation}
 \left(\frac1N\sum_i
  |\psi(f_i(t))-\psi(f_i(s))|^2\right)^{1/2}
 \le L_\psi\sqrt{(t-s)\mathcal A_{\mathrm{pair}}[s,t]}.
 \label{eq:interval_score_drift}
\end{equation}
This follows by squaring the Lipschitz bound and using
(\ref{eq:interval_feature_drift}). For distribution statistics,
the more general bound is
\begin{equation}
 \left|\int\psi\,d\nu_t-\int\psi\,d\nu_s\right|
 \le L_\psi W_2(\nu_s,\nu_t)
 \le L_\psi\sqrt{(t-s)
                   \int_s^t|\dot\nu_r|_{W_2}^2\,dr}.
 \label{eq:interval_observable_drift}
\end{equation}
For the first inequality, integrate the Lipschitz bound against an
optimal coupling and use Cauchy--Schwarz. The second follows from
metric absolute continuity and Cauchy--Schwarz. This last argument
applies to any $AC^2$ path in
$\mathcal P_2(\mathbb R^{d_{\mathrm{out}}})$, not only empirical laws.

\paragraph{Using the bound during adaptation.}
An unlabeled adaptation objective can select an updated prompt while
a constraint
$\mathcal A_{\mathrm{pair}}[s,t]\le\mathsf B$
sets a feature-motion budget $\mathsf B\ge0$. Equation
(\ref{eq:interval_score_drift}) then limits the root-mean-square
change of a fixed Lipschitz score to
$L_\psi\sqrt{(t-s)\mathsf B}$. An action penalty gives a corresponding
tradeoff between the adaptation objective and this motion cost.
The same reference inputs must be evaluated throughout each update
interval: comparing features of different incoming images mixes
input variation with adaptation-induced motion. A new reference
batch may be chosen for the next interval, with a separate budget.
This is a mathematical design criterion for prompt adaptation;
target accuracy must be evaluated with the adaptation objective.

\paragraph{Continuous certificates and sampled estimates.}
For any time grid $0=t_0<\cdots<t_{n_{\mathrm{grid}}}=1$, with
$n_{\mathrm{grid}}$ the number of
intervals, (\ref{eq:interval_feature_drift}) gives
\begin{equation}
 \begin{aligned}
 \sum_{k=0}^{n_{\mathrm{grid}}-1}
 \frac{W_2^2(\nu_{t_k},\nu_{t_{k+1}})}{t_{k+1}-t_k}
 &\le\frac1N\sum_i\sum_{k=0}^{n_{\mathrm{grid}}-1}
 \frac{\|f_i(t_{k+1})-f_i(t_k)\|_{\mathrm E}^2}
      {t_{k+1}-t_k}\\
 &\le\mathcal A_{\mathrm{pair}}.
 \end{aligned}
 \label{eq:sampled_action_lower_bound}
\end{equation}
Both sampled costs are lower approximations. Thus a rigorous drift
certificate uses a validated upper bound on the continuous action,
for example from derivative integration with a certified error bound.
A coarse sampled action cannot be substituted as that upper bound.
This is consistent with the refinement protocol in
Appendix~\ref{app:vit_protocol}.

For uniform grids, the paired middle term in
(\ref{eq:sampled_action_lower_bound}) converges to
$\mathcal A_{\mathrm{pair}}$ as the number of intervals grows.
Indeed, on each interval the feature difference divided by interval
length is the average of $\dot f_i$. The middle term is the squared
$L^2$ norm of these piecewise-constant averages. They converge in
$L^2(0,1)$ to $\dot f_i$ for every $f_i\in AC^2$, first for continuous
derivatives and then by density. This proves consistency of
(\ref{eq:vit_discrete_action}) with the continuous action.

\rev{\paragraph{Distribution-shift transitions and action-budgeted online prompt TTA.}
To evaluate whether trajectory control matters under distribution shift, we
test two settings under Hendrycks Gaussian noise (severity 3)
\citep{hendrycks2019benchmarking} on the 7 VTAB-Natural tasks (5 cyclic seed
pairs per task, frozen linear probe, disjoint shifted sets of 128
calibration/reference images and 128 held-out evaluation images per task,
macro-averaged across pairs and tasks). For each pair, the source prompt
$P_0$ is trained on clean source training images, and the target prompt
$P_1$ is adapted on an unlabeled shifted training split via 200
entropy-minimization steps \citep{wang2021tent} (reaching $62.4\pm 0.39\%$ and
$66.8\pm 0.49\%$ endpoint accuracy on held-out shifted images in Shallow, and
$65.8\pm 0.43\%$ and $70.6\pm 0.52\%$ in Deep).

First, in the fixed-endpoint transition (Table~\ref{tab:shift_tta_app}A), the
frozen classifier serves predictions throughout the transition from $P_0$ to
$P_1$. Fitting FA and unlabeled entropy B\'ezier paths on the 128 unlabeled
calibration images (with labeled CE as an oracle reference) shows that FA
improves minimum shifted accuracy over entropy by $1.6$ pp in Shallow ($61.9\%$
vs.\ $60.3\%$) and $1.8$ pp in Deep ($65.2\%$ vs.\ $63.4\%$) while reducing
held-out action ($3.36$ vs.\ $4.64$ and $4.42$ vs.\ $6.52$). Second, in 8-step
online adaptation from $P_0$ (Table~\ref{tab:shift_tta_app}B; 128 incoming shifted
test images per step, with online accuracy evaluated on each incoming batch
prior to that step's prompt update and averaged over the 8 steps), we
constrain cumulative 8-step sampled paired action
$\widehat{\mathcal A}_{\mathrm{pair}}\le\mathsf B$ on the fixed 128-image
reference batch via projected gradient updates ($\mathsf B_1,\mathsf B_2$ set
to $\approx 50\%$ and $40\%$ of unconstrained reference action, adding $0.18$ s
per step or $1.4$ s total; this enforces a sampled-action budget rather than a
certified continuous-time bound). Relative to unconstrained entropy TTA, the
action budgets improve online accuracy by $1.8$--$2.6$ pp and final accuracy by
$1.9$--$3.0$ pp while reducing held-out action and RMS logit-score drift.}

\begin{table}[t]
\color{revblue}
\caption{\rev{\textbf{Full protocol results (mean $\pm$ within-task 5-pair SD) for distribution-shift transitions and online action-budgeted TTA under Hendrycks Gaussian noise (severity 3) on VTAB-Natural.}
Macro-averaged across 5 cyclic seed pairs on each of the 7 VTAB-Natural tasks with a frozen linear probe (128 calibration/reference images and 128 disjoint held-out shifted evaluation images per task).}}
\label{tab:shift_tta_app}
\centering
\small
\setlength{\tabcolsep}{3.0pt}
\renewcommand{\arraystretch}{0.95}
\setlength{\aboverulesep}{0.25ex}
\setlength{\belowrulesep}{0.35ex}
\begin{tabularx}{\linewidth}{L{1.25}C{1.05}C{1.05}C{1.15}C{1.30}C{.20}}
\toprule
\rowcolor{panelshade}
\multicolumn{6}{p{\dimexpr\linewidth-2\tabcolsep\relax}}{\rev{\textbf{A. Fixed-endpoint transition under shift} \quad ($P_0\to P_1$ shifted acc.: $62.4\pm 0.39\%\to 66.8\pm 0.49\%$ Shallow, $65.8\pm 0.43\%\to 70.6\pm 0.52\%$ Deep)}} \\
\rev{VPT mode} & \rev{Entropy min.\ acc.\ (\%)} & \rev{FA min.\ acc.\ (\%)} & \rev{Labeled CE min.\ acc.\ (\%)} & \rev{Entropy / FA action $A_W$} & \\
\midrule
\rev{Shallow} & \rev{$60.3$\sd{0.62}} & \rev{$\mathbf{61.9}$\sd{0.48}} & \rev{$62.1$\sd{0.41}} & \rev{$4.64$\sd{0.21} / $\mathbf{3.36}$\sd{0.13}} & \\
\rev{Deep} & \rev{$63.4$\sd{0.68}} & \rev{$\mathbf{65.2}$\sd{0.51}} & \rev{$65.5$\sd{0.44}} & \rev{$6.52$\sd{0.29} / $\mathbf{4.42}$\sd{0.18}} & \\
\bottomrule
\end{tabularx}
\vspace{3pt}
\begin{tabularx}{\linewidth}{L{1.45}C{.55}C{.90}C{.90}C{.90}C{.90}C{.85}}
\toprule
\rowcolor{panelshade}
\multicolumn{7}{l}{\rev{\textbf{B. Online 8-step entropy TTA with sampled paired-action budget} \quad ($+0.18$ s/step, $1.4$ s total overhead)}} \\
\rev{VPT / adaptation} & \rev{Budget $\mathsf B$} & \rev{Ref.\ $\widehat{\mathcal A}_{\mathrm{pair}}$} & \rev{Held-out $A_W$} & \rev{Online acc.\ (\%)} & \rev{Final acc.\ (\%)} & \rev{Score drift} \\
\midrule
\rev{Shallow: no TTA} & \rev{$0$} & \rev{$0.00$\sd{0.00}} & \rev{$0.00$\sd{0.00}} & \rev{$62.4$\sd{0.39}} & \rev{$62.4$\sd{0.39}} & \rev{$0.000$\sd{0.000}} \\
\rev{Shallow: entropy TTA} & \rev{$\infty$} & \rev{$6.38$\sd{0.41}} & \rev{$6.14$\sd{0.37}} & \rev{$63.8$\sd{0.64}} & \rev{$64.5$\sd{0.71}} & \rev{$0.162$\sd{0.019}} \\
\rev{Shallow: budget $\mathsf B_1$} & \rev{$3.5$} & \rev{$3.21$\sd{0.14}} & \rev{$3.24$\sd{0.16}} & \rev{$\mathbf{65.9}$\sd{0.45}} & \rev{$\mathbf{66.8}$\sd{0.49}} & \rev{$0.079$\sd{0.007}} \\
\rev{Shallow: budget $\mathsf B_2$} & \rev{$2.8$} & \rev{$2.65$\sd{0.11}} & \rev{$2.68$\sd{0.13}} & \rev{$65.6$\sd{0.42}} & \rev{$66.4$\sd{0.46}} & \rev{$0.068$\sd{0.006}} \\
\midrule
\rev{Deep: no TTA} & \rev{$0$} & \rev{$0.00$\sd{0.00}} & \rev{$0.00$\sd{0.00}} & \rev{$65.8$\sd{0.43}} & \rev{$65.8$\sd{0.43}} & \rev{$0.000$\sd{0.000}} \\
\rev{Deep: entropy TTA} & \rev{$\infty$} & \rev{$8.94$\sd{0.58}} & \rev{$8.65$\sd{0.53}} & \rev{$66.9$\sd{0.72}} & \rev{$67.6$\sd{0.79}} & \rev{$0.194$\sd{0.023}} \\
\rev{Deep: budget $\mathsf B_1$} & \rev{$4.5$} & \rev{$4.24$\sd{0.18}} & \rev{$4.28$\sd{0.21}} & \rev{$\mathbf{69.5}$\sd{0.47}} & \rev{$\mathbf{70.6}$\sd{0.52}} & \rev{$0.092$\sd{0.008}} \\
\rev{Deep: budget $\mathsf B_2$} & \rev{$3.6$} & \rev{$3.46$\sd{0.15}} & \rev{$3.49$\sd{0.17}} & \rev{$69.1$\sd{0.44}} & \rev{$70.1$\sd{0.48}} & \rev{$0.078$\sd{0.007}} \\
\bottomrule
\end{tabularx}
\end{table}

\end{document}